\pdfoutput=1
\documentclass[11pt]{article}
\usepackage[preprint]{acl}
\usepackage{times}
\usepackage{latexsym}
\usepackage[T1]{fontenc}
\usepackage[utf8]{inputenc}
\usepackage{microtype}
\usepackage{inconsolata}
\usepackage{amsmath,amssymb,amsthm}
\usepackage{booktabs}
\usepackage{array}
\usepackage{graphicx}
\usepackage{xcolor}
\usepackage{capt-of}

\hypersetup{
  pdfauthor={Weiye Yuan, Xue Yuan},
  pdftitle={Rotation Is Not Enough: Precedence Cycles and the Design of Order Averaging for LLM Readers and Judges},
  pdfsubject={Preprint},
  pdfcreator={LaTeX},
  pdfproducer={pdfTeX}
}

\newcommand{\HU}{\mathcal{H}_U^+}
\newcommand{\Hou}{\mathcal{H}_{\mathrm{odd\text{-}unary}}}
\newcommand{\Hcurl}{\mathcal{H}_{\mathrm{curl}}}
\newcommand{\Hun}{\mathcal{H}_{\mathrm{unary}}}

\newcommand{\EU}{E_U}

\title{Rotation Is Not Enough: Precedence Cycles and the Design of Order Averaging for LLM Readers and Judges}

\author{
  Weiye Yuan \qquad Xue Yuan\thanks{\ \ Corresponding author.} \\
  School of Automation and Intelligence, Beijing Jiaotong University, Beijing, China \\
  \texttt{xyuan@bjtu.edu.cn}
}

\begin{document}
\maketitle

\begin{abstract}
Language models that answer multiple-choice questions or judge candidate responses change their outputs when the options are reordered. The standard remedy averages predictions over cyclic rotations of the list, and the standard audit measures disagreement across those same rotations. We show that both are blind to one family of order effects. Precedence cycles, which depend on the circular orientation in which items appear, survive every rotation: rotation moves each item through every slot but never reverses that orientation. We give an exact spectral account of order averaging on the symmetric group. Every subgroup-coset design leaves a computable blind spot, its expected Brier excess equals the order energy inside that blind spot, and an audit on the same orders sees exactly the complement. Reversal-closed, adjacency-balanced terrace designs such as Williams squares remove every precedence cycle at the cost of rotation for even $K$. In an internally sealed, two-stage preregistered study on seven open backbones, six-pass Williams averaging on six-option MMLU-Pro cuts order-induced Brier excess by 61.2\% and 40.0\% relative to rotation on two 2026 frontier models (absolute reductions 0.0050 and 0.0083) and by 56.5\% and 51.0\% relative to random reversal pairs. It lowers both Brier excess and anchor-to-anchor answer flips (by 4.3--7.8 percentage points) relative to rotation on all seven backbones, keeps its frontier fidelity gains at eight and ten options, and lowers exact order energy on all five RewardBench~2 judges.
\end{abstract}

\section{Introduction}\label{sec:intro}

When a language model reads a list of options or candidate responses, its answer depends on the list order \citep{zheng2024large,li2026position}. This affects multiple-choice reading \citep{zheng2024large}, few-shot prompting \citep{lu2022fantastically,zhao2021calibrate}, long contexts \citep{liu2024lost}, listwise reranking \citep{tang2024permutation,sun2023chatgpt}, and LLM-as-judge evaluation \citep{zheng2023judging,wang2024fair,shi2025judging}. The standard remedy averages predictions across several orders, with rotation as the default: CircularEval rotates options to audit consistency \citep{liu2024mmbench}, PriDe estimates a slot prior from cyclic views \citep{zheng2024large}, and judge pipelines swap or rotate candidates \citep{zheng2023judging,wang2024fair}. Rotation costs only $K$ passes and puts every item in every slot once.

\begin{figure*}[t]
\centering
\includegraphics[width=\linewidth]{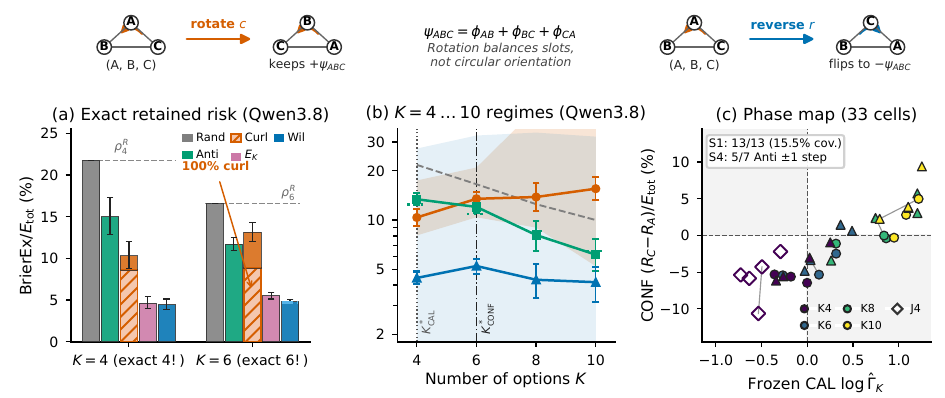}
\caption{\textbf{Rotation keeps every precedence cycle; reversal removes it.} Top: rotating three items preserves circular orientation; reversal flips it. (a) Qwen3.8-27B, exact over all $K!$ orders ($N{=}192$): Brier excess left by equal-cost designs at $K{=}4,6$ (bars: 95\% item-clustered percentile bootstrap CIs). At $K{=}6$, Cyclic exceeds Antithetic while Cyclic minus curl falls below it, though the Holm-adjusted IUT is non-significant ($p{=}0.146$). (b) Frozen CAL bounds (conservative envelopes, not CIs) versus CONF observations (points with 95\% percentile CIs) for $K{\in}\{4,6,8,10\}$. (c) Frozen CAL phase ratio $\hat\Gamma_K$ versus CONF Cyclic minus Antithetic gap across 28 MCQ and 5 judge cells.}
\label{fig:hero}
\end{figure*}

That property suggests rotation matches any equal-cost slot-balanced design. It does not. On six-option MMLU-Pro \citep{wang2024mmlupro}, six cyclic rotations leave 13.5\% of Qwen3.8-27B's order variance in the averaged prediction, while six Williams orders leave 5.2\%. Against random reversal pairs of the same cost, rotation wins on some models and loses on others, and the winner flips as options are added (Figure~\ref{fig:hero}). Slot balance cannot explain a gap that moves with the model and with $K$.

A central mechanism is a mode of order dependence invisible to rotation. Treating a readout as a function on the symmetric group $S_K$ \citep{diaconis1988group}, pairwise precedence effects split by combinatorial Hodge theory \citep{jiang2011statistical} into a gradient part (per-item scores) and a curl part of dimension $\binom{K-1}{2}$: precedence cycles where A gains from preceding B, B from preceding C, and C from preceding A. Rotation moves every item through every slot, canceling slot preferences, yet maps any three-item cycle to itself. Reversal flips its orientation. \emph{Rotation balances slots, not circular orientation.}

We turn this observation into an exact theory of equal-cost order averaging (\S\ref{sec:theory}). Every subgroup-coset design has a blind spot whose dimension depends only on its size, yet conjugate designs such as rotation and Williams squares \citep{williams1949experimental} place that blind spot in different modes (Theorem~1). Expected Brier excess equals the energy inside that blind spot (Proposition~1), and an audit on the same orders sees exactly the complement (Corollary~A): averaging and auditing with rotations miss the same modes. A crossover criterion states when rotation loses to random reversal pairs (Corollary~B), an axiom-mode correspondence matches balance properties to removed modes (Theorem~2), and $K$-free estimators measure every mode without enumerating $K!$ orders (Theorem~3).

We test this theory in an internally sealed, two-stage preregistered study (\S\ref{sec:setup}) across seven open backbones from five labs (four hybrid or recurrent), spanning MMLU-Pro ($K{\in}\{4,6,8,10\}$) and RewardBench~2 listwise judging \citep{malik2025rewardbench2}. At $K{=}6$, Williams cuts order-induced Brier excess on two 2026 frontier models by 61.2\% and 40.0\% versus rotation and 56.5\% and 51.0\% versus random reversal pairs, lowers both excess and flips versus rotation on 7/7 backbones, preserves frontier gains at $K{\in}\{8,10\}$, and leaves the least exact order energy on all five judges.

Our contributions are:
\begin{itemize}
\setlength{\itemsep}{1pt}\setlength{\parskip}{0pt}
\item \textbf{Exact theory.} Spectral account on $S_K$: coset blind spots (Theorem~1), Brier excess (Proposition~1), audit-averaging duality and margin bound (Corollary~A), crossover criterion (Corollary~B), axiom-mode correspondence (Theorem~2), and $K$-free estimation (Theorem~3; Appendix~\ref{app:proofs}).
\item \textbf{Same-cost terraces.} Reversal-closed, adjacency-balanced terraces ($b_W$ and distance-2-dispersed seeds) cancel unary, curl, and odd modes in $K$ passes with zero latency overhead.
\item \textbf{Preregistered evidence.} 7/7 wins over rotation in flips ($-$4.3 to $-$7.8\,pp) and excess, 8/8 large-$K$ frontier fidelity wins, and 3/3 registered judge wins, disclosing all eight gates (3 PASS, 5 FAIL; \S\ref{sec:results}).
\item \textbf{Mechanism and design probes.} Exact $K{=}4{\to}6$ curl test, Olmo matched pair, and distance-2 terrace analysis (\S\ref{sec:mechanism}, \S\ref{sec:boundaries}).
\end{itemize}

\section{Theory: What Each Design Keeps}\label{sec:theory}

\subsection{Orders, readouts, and designs}\label{sec:theory_setup}

Fix $K$ candidates. An order $\pi\in S_K$ places candidate $\pi(s)$ in slot $s$. A readout $f(\pi)\in\Delta^{C-1}$ is the model's probability vector in canonical candidate coordinates (candidate $v$ keeps index $v$ across slots). With $\langle u,v\rangle=\frac{1}{K!}\sum_\pi\langle u(\pi),v(\pi)\rangle$, let $\mu=\frac{1}{K!}\sum_\pi f(\pi)$, $f_0=f-\mu$, and $E_{\mathrm{tot}}=\|f_0\|^2$. The right action is $(R_gf)(\pi)=f(\pi g)$ with $(\pi g)(s)=\pi(g(s))$, rotation $c(s)=s{+}1 \bmod K$, and reversal $r(s)=K{-}1{-}s$.

A design $D$ of size $m$ pairs relative rows $d_1,\dots,d_m$ with a uniform anchor $\sigma$, returning $\hat\mu_D=\frac1m\sum_j f(\sigma d_j)$ with fidelity loss $\mathrm{BrierEx}(D)=\mathbb{E}\|\hat\mu_D-\mu\|^2$. At $m{=}K$: Cyclic uses $(c^t)_{t<K}$; Williams uses $(c^tb_W)_{t<K}$ with zigzag seed $b_W=(0,1,K{-}1,2,K{-}2,\dots)$, so row $t$ places $\sigma((b_W(s){+}t)\bmod K)$ in slot $s$; $E_K$ uses $\{c^{2j},c^{2j}r\}_{j<K/2}$; Antithetic draws $K/2$ distinct reversal pairs $\{\pi,\pi r\}$; Random-WoR draws $K$ distinct orders.

\subsection{Modes and blind spots}\label{sec:modes}

Split centered functions into reversal-even and reversal-odd halves via $P_\pm=\frac12(I\pm R_r)$. Unary functions depend only on which candidate occupies each slot ($\EU$ full centered unary energy; $\HU$ reversal-even unary). For $i\neq j$, the reversal-odd precedence function $\phi_{ij}(\pi)=\operatorname{sgn}(\pi^{-1}(j)-\pi^{-1}(i))$ indicates whether $i$ precedes $j$. Skew matrices define $f_B=\sum_{i<j}B_{ij}\phi_{ij}$: gradient flows $B_{ij}=x_i-x_j$ are unary, whereas zero-row-sum flows form the $\binom{K-1}{2}$-dimensional curl subspace $\mathcal{H}_{\mathrm{curl}}$, spanned by triangles $\psi_{xyz}=\phi_{xy}+\phi_{yz}+\phi_{zx}$ ($+1$ if $x,y,z$ sit in cyclically increasing order, $-1$ otherwise). With symmetric adjacency $\mathcal H_{\mathrm{sa}}$ and higher remainders, $L^2_0(S_K)$ decomposes orthogonally as
\begin{align*}
L^2_0(S_K)
&= \underbrace{\HU\oplus\mathcal H_{\mathrm{sa}}\oplus\mathcal H_{\mathrm{he}}}_{\text{reversal-even}} \\
&\quad \oplus \underbrace{\mathcal H_{\mathrm{grad}}\oplus\mathcal H_{\mathrm{ou,rest}}\oplus\mathcal H_{\mathrm{curl}}\oplus\mathcal H_{\mathrm{ho}}}_{\text{reversal-odd}},
\end{align*}
with mode energy $E_X=\|P_Xf_0\|^2$ and $E_{\mathrm{even}}=\|P_+f_0\|^2$.

\paragraph{Theorem 1 (Coset blind spots).} \textit{Let $H\le S_K$ have order $m$ and $D=gH$. (i) $\mathrm{BrierEx}(D)=\|P_Hf_0\|^2$ for $P_H=\frac1m\sum_{h\in H}R_h$. (ii) $\mathrm{Fix}(H)\cap L^2_0$ has dimension $K!/m-1$ per coordinate. (iii) Inside each isotypic component $I_\nu$ of degree $d_\nu$ its dimension is $d_\nu\,\mathrm{mult}_\nu(H)$, so conjugate subgroups share identical profiles. (iv) $P_C=P_{\langle c\rangle}$ fixes every curl function and annihilates every unary function.}

\noindent\emph{Idea.} $\sigma\mapsto\sigma g$ is a bijection; (iii) is a character count \citep{serre1977linear,diaconis1988group}; (iv) uses $f_B(\pi c)-f_B(\pi)=-2\sum_jB_{\pi(0)j}=0$ under zero row sums.

Williams is conjugate to rotation ($H_W=b_W^{-1}\langle c\rangle b_W$), so both have identical blind-spot profiles (119 dimensions at $K{=}6$, 20 in $I_{(4,1,1)}$). Cyclic's 20 contain all 10 curl dimensions; Williams' 20 are orthogonal to curl (Theorem~2). Under isotropic order energy, every $m{=}K$ coset design leaves $\rho^R_KE_{\mathrm{tot}}$; location separates equal-cost designs because real LLM spectra concentrate in low-order modes.

\paragraph{Proposition 1 (Brier excess is blind-spot energy).} \textit{For every anchored design $D$ and fixed target $y$,}
\[
\mathbb E\|\hat\mu_D-y\|^2=\|\mu-y\|^2+\mathrm{BrierEx}(D).
\]
\textit{Random-WoR ($m{=}K$) has $\mathrm{BrierEx}=\rho^R_KE_{\mathrm{tot}}$, $\rho^R_K=\frac{K!-K}{K(K!-1)}$; Antithetic ($K/2$ pairs) has $\mathrm{BrierEx}=\rho_KE_{\mathrm{even}}$, $\rho_K=\frac{2(K!-K)}{K(K!-2)}$.}

Because $\|\mu-y\|^2$ is common to all unbiased designs, gold-Brier differences across exact arms measure fidelity. The identity applies to averages of a fixed reader $f$, not to PINE (which modifies $f$) or PriDe (nonlinear prior division).

\paragraph{Corollary A (Audit-averaging duality).} \textit{With audit disagreement $V_D=\frac1m\sum_j\|f(\sigma d_j)-\hat\mu_D\|^2$, every anchored design satisfies}
\begin{align*}
\mathbb EV_D &= E_{\mathrm{tot}}-\mathrm{BrierEx}(D), \\
\kappa_D &:= \frac{\mathbb EV_D}{E_{\mathrm{tot}}} = 1-\frac{\mathrm{BrierEx}(D)}{E_{\mathrm{tot}}}.
\end{align*}

Averaging and auditing on the same orders have complementary blind spots: a rotation audit cannot see curl, and $\kappa_W-\kappa_C=(R_C-R_W)/E_{\mathrm{tot}}$ quantifies the energy recovered by switching. Uniform pooled row laws are necessary (counterexample in Appendix~\ref{app:proofs}). The identity is the ambiguity decomposition of \citet{krogh1995neural} over orders. For a fixed item whose exhaustive average $\mu$ has unique top-1 winner with top-2 margin $\gamma(\mu){=}\mu_{(1)}{-}\mu_{(2)}{>}0$, Cauchy--Schwarz and Markov yield the decision-layer bridge $\Pr(\arg\max\hat\mu_D{\ne}\arg\max\mu)\le\min\{1,\,2\,\mathrm{BrierEx}(D)/\gamma(\mu)^2\}$ (Appendix~\ref{app:proofs}); this bounds deviation from $\arg\max\mu$ (not gold accuracy or pairwise anchor flips).

\begin{table}[t]
\centering
\scriptsize
\setlength{\tabcolsep}{2.2pt}
\begin{tabular}{@{}lccccccc@{}}
\toprule
Design ($m{=}K$) & S & P & R & A & Slot & Curl & Sym-Adj \\
\midrule
Random-WoR & -- & -- & -- & -- & $\rho_K^R$ & $\rho_K^R$ & $\rho_K^R$ \\
Antithetic & -- & \checkmark & \checkmark & -- & $\rho_K^\dagger$ & $0$ & $\rho_K$ \\
Cyclic & \checkmark & $\times$ & $\times$ & $\times$ & $0$ & \textbf{1} & 0.75 / 0.83 \\
$E_K=\langle c^2,r\rangle$ & \checkmark & \checkmark & \checkmark & $\times$ & $0$ & $0$ & 1.00 / 0.89 \\
\textbf{Williams / Alt} & \checkmark & \checkmark & \checkmark & \checkmark & $0$ & $0$ & $0$ \\
\bottomrule
\end{tabular}
\caption{\textbf{Exact spectral retention of equal-cost ($m{=}K$) order-averaging designs.} Columns S/P/R/A denote slot, pairwise-order, reversal-closure, and adjacency balance (Antithetic satisfies P and R row-wise per drawn sample). Pairs give trace retention at $K{=}4 / K{=}6$ (Appendix~\ref{app:proofs}); $\dagger$ retains only reversal-even unary ($\mathcal{H}_U^+$). Higher-even trace retention ($K{=}4 / K{=}6$) is $1/12$ / $103/680$ (Cyclic), $1/2$ / $111/340$ ($E_K$), and $5/6$ / $7/20$ (Williams/Alt).}
\label{tab:design_theory}
\end{table}

\subsection{When rotation loses, and what removes curl}\label{sec:crossover}

\paragraph{Corollary B (Rotation risk decomposition and crossover).} \textit{For even $K$, $m{=}K$, $R_D=\mathrm{BrierEx}(D)$, and reversal-even / higher-odd parts $f_+, h_-$ of $f_0$,}
\begin{align*}
R_C &= E_{\mathrm{curl}}+\|P_Ch_-\|^2+\|P_Cf_+\|^2, \\
E_{\mathrm{curl}} &\le R_C \le E_{\mathrm{tot}}-\EU,
\end{align*}
\textit{and, whenever $E_{\mathrm{even}}>0$,}
\begin{align*}
\operatorname{sgn}(R_C-R_A) &= \operatorname{sgn}(\Gamma_K-1), \\
\Gamma_K &:= \frac{R_C}{\rho_KE_{\mathrm{even}}}.
\end{align*}
\textit{Against Random-WoR, $R_C>R_R$ iff $R_C/E_{\mathrm{tot}}>\rho^R_K$ (sufficient: $E_{\mathrm{curl}}/E_{\mathrm{tot}}>\rho^R_K$).}

Rotation keeps curl at unit rate (Table~\ref{tab:design_theory}) while $\rho_K$ falls ($0.455, 0.331, 0.250, 0.200$ for $K=4,6,8,10$) and $\dim\mathcal H_{\mathrm{curl}}$ grows from 3 to 36, so rotation's winner-loser status can flip with $K$. Random-WoR thresholds $\rho^R_K$ are 21.74\%, 16.55\%, 12.50\%, 10.00\%. Table~\ref{tab:design_theory} gives mode-averaged trace retentions; the adjacency/higher-even cross term inside $\|P_Cf_+\|^2$ is generically nonzero. Cyclic loses \emph{because of} curl iff $R_C-R_A>0$ and $R_C-E_{\mathrm{curl}}-R_A<0$ (\S\ref{sec:mechanism}).

\paragraph{Theorem 2 (Axiom-mode correspondence).} \textit{Let $D$ be an equal-weight multiset of $m$ rows with operator $T_D$ ($K\ge4$). Consider (S) slot balance (each candidate in each slot $m/K$ times); (P) precedence balance (each pair ordered each way $m/2$ times); (R) reversal closure ($d\mapsto dr$); (A) adjacency balance (each pair adjacent $2m/K$ times). Then (a) $T_D\Hun=0\iff$(S); (b) under (S), $T_D\Hcurl=0\iff$(P); (c) $T_DP_-L^2=0\iff$(R), and (R)$\Rightarrow$(P); (d) under (S), $T_D\mathcal H_{\mathrm{sa}}=0\iff$(A); (e) (S)$\wedge$(R)$\Rightarrow\mathrm{lcm}(K,2)\mid m$ (thus $2K\mid m$ for odd $K$); (f) under (S,R,A), $\mathrm{BrierEx}(D)=\|T_Dh_+\|^2\le E_{\mathrm{he}}$; (g)--(h) for every even $K$, $b_W$ (and locked Alt\_Terrace seeds) yields a minimal $m{=}K$ (S,R,A) design; (i) $E_K=\langle c^2,r\rangle$ satisfies (S,P,R) and fails (A).}

Cyclic satisfies only (S) and keeps every precedence cycle. Reversal-closed terraces satisfy all four axioms at $m{=}K$, leaving only part of $E_{\mathrm{he}}$. Because $P_{E_K}\ne P_{H_W}$ on $\mathcal H_{\mathrm{he}}$, $R_{E_K}-R_W$ is a whole-design contrast, not an isolated adjacency ablation.

\subsection{Estimation without enumerating \texorpdfstring{$K!$}{K!}}\label{sec:estimation}

\paragraph{Theorem 3 ($K$-free estimation).} \textit{Each mode energy is $\mathbb E[k_A(\pi,\sigma)\langle f(\pi)-c,f(\sigma)-c\rangle]$ with closed-form $O(K^2)$ kernels in fixed points, Kendall concordance, and shared adjacencies (a--b). Fixed-center U-statistics (and leave-two-out diagnostics) are unbiased (c--e); between-origin U-statistics on reversal-paired origins and cyclic-orbit panels estimate $E_{\mathrm{even}}$ and $R_C$ unbiasedly (f); split-panel $\hat B_{\mathrm{split}}{=}(p{-}\bar u)^\top(p{-}\bar v)$ and complete-U $\hat B_U$ are unbiased for $\|p{-}\mu\|^2$ against an independent i.i.d.\ panel (g); and i.i.d.\ plug-in $\hat E_{\mathrm{even}}$ degenerates when $\binom M2\ll K!$ (h).}

Production CAL ($K{\ge}6$) uses fixed-center between-origin reversal-paired panels; production CONF online uses independent i.i.d.\ split panels scored with $\hat B_{\mathrm{split}}$ (\S\ref{sec:setup}). All results in Theorems~1--3, Proposition~1, and Corollaries~A--B pass a 66-check verification suite (Appendix~\ref{app:proofs}).

\section{Experimental Setup}\label{sec:setup}

\paragraph{Backbones.} We evaluate seven open-weight backbones (Hub revisions in Table~\ref{tab:app_models}): 2026 frontier models Qwen3.8-27B (hybrid) and Gemma-4-31B (dense); same-lab pair Olmo-Hybrid-7B and Olmo-3-7B; Nemotron-3-Nano-30B (hybrid); Qwen3.5-9B \citep[hybrid;][]{qwen2026qwen35}; and Llama-3.1-8B \citep{grattafiori2024llama3} (dense; four hybrid/recurrent, \citealp{yang2025gated}). All runs use bfloat16 and SDPA on A100-80GB GPUs.

\paragraph{Tasks and splits.} Multiple choice uses MMLU-Pro \citep{wang2024mmlupro} with nested option sets ($K{\in}\{4,6,8,10\}$) and disjoint splits: CAL (100 items), E1 (192 items, all $K!$ orders at $K{\in}\{4,6\}$), CONF (500 at $K{\in}\{4,6\}$, 200/cell at $K{\in}\{8,10\}$), and a 1{,}500-item frontier accuracy extension. Judging uses four-candidate RewardBench~2 \citep{malik2025rewardbench2} (CAL 64, CONF 500), decoded over all 24 orders.

\paragraph{Designs and metrics.} At $m{=}K$ we compare Random-WoR, Antithetic, Cyclic, $E_K$, Williams, and Alt\_Terrace; judges add two-pass ReversalPair and AdjRot. Readouts are renormalized allowed-letter probabilities in canonical coordinates. Accuracy and flip rate use sampled anchors (4 at $K{\in}\{4,6\}$ and judges; 2 at $K{\in}\{8,10\}$). Fidelity $\mathrm{BrierEx}/E_{\mathrm{tot}}$ is exact when all $K!$ orders are decoded; otherwise CONF uses unbiased $\hat B_{\mathrm{split}}$ against an independent i.i.d.\ uniform reference panel (32+32 at $K{=}6$, 24+24 at $K{\in}\{8,10\}$; Theorem~3(g)).

\paragraph{One-pass baselines.} PriDe \citep{zheng2024large} fits a slot prior on CAL and applies it in one pass on CONF. PINE$^\dagger$ is our audited prefill-only blockwise PINE variant \citep{wang2025eliminating}: inapplicable to the four hybrid/recurrent models, exact on fixed option blocks (744 checks at $K{\in}\{4,6\}$), and passed native-position parity on 16/16 CUDA fixtures at stated tolerances, but non-invariant under letter-prefixed relabeling (\texttt{A.}, \texttt{B.}, \dots) with group-level suffix lookahead; we scope it strictly as an audited variant.

\paragraph{Preregistration.} Under an internally sealed (SHA-256 hash-chained) protocol, Freeze-A (\texttt{afb73703}) locked splits, designs, hypotheses, Holm families, and thresholds. Freeze-A.1 switched $K{\ge}6$ CAL to 32 independent origins plus reversals (fixed-center between-origin U-statistics, Theorem~3(f),(h)) prior to $K{\in}\{8,10\}$ CAL and CONF. Freeze-B (\texttt{36a00df6}) sealed CAL bounds, sign calls, crossovers, and judge coefficients; we detected no CONF refit and verified hash agreement (hashes verify content, not external chronology). An exploratory pilot (Appendix~\ref{app:pilot}, Table~\ref{tab:app_pilot}) never serves as confirmation.

\paragraph{Statistics.} CIs are unadjusted 95\% item-clustered percentile bootstrap intervals (20{,}000 draws); tests use centered studentized bootstrap or sign-flip tests (Monte Carlo floor $1/20{,}001$). Holm \citep{holm1979simple} controls error within ten registered families; eight gates pair tests with locked cutoffs (\S\ref{sec:ledger}, Table~\ref{tab:app_gates}).

\section{Results}\label{sec:results}

\subsection{Equal-cost averaging}\label{sec:equal_cost}

\begin{table*}[t]
\centering
\scriptsize
\setlength{\tabcolsep}{1.35pt}
\begin{tabular}{@{}l@{\;}ccccccccc@{\;\;}cccccc@{\;\;}cc@{\;}cc@{\;}cc@{}}
\toprule
 & \multicolumn{9}{c}{\textbf{Accuracy (\%) $\uparrow$}} & \multicolumn{6}{c}{\textbf{$\mathrm{BrierEx}/E_{\mathrm{tot}}$ (\%) $\downarrow$}} & \multicolumn{2}{c}{\textbf{Flip (\%) $\downarrow$}} & \multicolumn{2}{c}{\textbf{$\Delta$Acc (pp)}} & \multicolumn{2}{c}{\textbf{$\kappa$ $\uparrow$}} \\
\cmidrule(lr){2-10}\cmidrule(lr){11-16}\cmidrule(lr){17-18}\cmidrule(lr){19-20}\cmidrule(l){21-22}
Backbone & Base & PriDe & PINE$^\dagger$ & Rand & Anti & Cyc & $E_6$ & \textbf{Wil} & Alt & Rand & Anti & Cyc & $E_6$ & \textbf{Wil} & Alt & Cyc & \textbf{Wil} & W$-$C$^{500}$ & W$-$A$^{2k}$ & Cyc & \textbf{Wil} \\
\midrule
\multicolumn{22}{@{}l}{\textit{2026 Frontier Backbones (Acc pooled $N{=}2{,}000$; fidelity, flip, $\kappa$, and W$-$C$^{500}$ at $N{=}500$ conf.)}} \\
Qwen3.8-27B$^\blacktriangle$ & 64.0 & 63.6 & n/a & 68.3 & 69.1 & 68.9 & 69.4 & \underline{\textbf{69.6}} & 69.5 & 16.6 & 12.0 & 13.5 & 5.9 & 5.2 & \textbf{4.5} & 14.2 & \textbf{9.7}$^*$ & +0.95 & +0.49 & 0.86 & \textbf{0.96} \\
Gemma-4-31B$^\bullet$ & 69.8 & 70.8 & 63.4 & 74.1 & 75.0 & 75.6 & 76.3 & \underline{\textbf{76.5}} & 76.2 & 16.5 & 13.6 & 11.1 & 7.4 & \textbf{6.6} & 6.8 & 13.8 & \textbf{9.4}$^*$ & +1.05 & +1.56$^*$ & 0.87 & \textbf{0.91} \\
\midrule
\multicolumn{22}{@{}l}{\textit{Same-Lab Matched Pair ($N{=}500$)}} \\
Olmo-Hybrid-7B$^\blacktriangle$ & 41.5 & 42.8 & n/a & 46.6 & 47.6 & 48.5 & 49.5 & \underline{\textbf{49.9}} & 48.9 & 16.0 & 12.4 & 9.1 & 3.7 & 3.6 & \textbf{3.2} & 23.3 & \textbf{16.8}$^*$ & +1.30 & -- & 0.92 & \textbf{0.98} \\
Olmo-3-7B$^\bullet$ & 39.2 & 39.1 & 35.8 & 44.1 & 43.6 & \underline{\textbf{45.1}} & 44.7 & 44.7 & 44.0 & 16.9 & 15.1 & 9.8 & 4.9 & \textbf{4.5} & 4.5 & 27.5 & \textbf{20.0}$^*$ & -0.35 & -- & 0.90 & \textbf{0.95} \\
\midrule
\multicolumn{22}{@{}l}{\textit{Additional Contemporary \& Anchor Backbones ($N{=}500$)}} \\
Nemotron-3-30B$^\blacktriangle$ & 49.5 & 52.3 & n/a & 55.2 & 56.0 & 58.0 & 58.1 & 57.6 & \underline{\textbf{58.6}} & 16.7 & 12.5 & 7.7 & 3.4 & 2.9 & \textbf{2.7} & 20.3 & \textbf{16.0}$^*$ & -0.30 & -- & 0.93 & \textbf{0.98} \\
Qwen3.5-9B$^\blacktriangle$ & 59.4 & 60.0 & n/a & 63.1 & 62.7 & 63.9 & \underline{\textbf{64.2}} & 64.0 & 63.8 & 16.1 & 11.0 & 11.6 & 5.2 & 4.6 & \textbf{4.1} & 16.9 & \textbf{11.1}$^*$ & +0.10 & -- & 0.89 & \textbf{0.95} \\
Llama-3.1-8B$^\bullet$ & 38.9 & 42.4 & 31.8 & 44.5 & 46.4 & 47.6 & 48.4 & 49.6 & \underline{\textbf{49.8}} & 16.8 & 9.9 & 4.5 & 2.2 & 2.0 & \textbf{1.9} & 28.1 & \textbf{20.2}$^*$ & +2.05 & -- & 0.95 & \textbf{0.98} \\
\midrule
\multicolumn{18}{@{}l}{\textbf{Pooled across 7 backbones} (500 item clusters; 3,500 model-items; locked gate sf $p{<}0.01$)} & +0.69 & \multicolumn{3}{r@{}}{Holm $p{=}0.0198$ / sf $p{=}0.0266$} \\
\bottomrule
\end{tabular}
\caption{\textbf{Equal-cost ($m{=}6$) order averaging across all evaluated arms on six-option MMLU-Pro (E2).} $^{\blacktriangle}$hybrid/recurrent; $^{\bullet}$dense Transformer. Best unrounded all-arm equal-cost ($m{=}6$) accuracy is \textbf{bold} (best overall \underline{underlined}); lowest unrounded $\mathrm{BrierEx}/E_{\mathrm{tot}}$ across all six $m{=}6$ arms is \textbf{bold}. $^*$Holm-adjusted $p{<}0.05$ within its preregistered family. Williams lowers both $\mathrm{BrierEx}/E_{\mathrm{tot}}$ and decision flips vs.\ Cyclic on 7/7 backbones (all Holm $p{=}0.00035$), cutting frontier excess Brier risk by 61.2\% (abs.\ 0.00497, 95\% CI $[0.00422,0.00577]$) on Qwen3.8 and 40.0\% (abs.\ 0.00830, CI $[0.00614,0.01062]$) on Gemma-4 vs.\ Cyclic, and 56.5\% / 51.0\% vs.\ Antithetic (gate \texttt{E2\_K6\_frontier\_fidelity}: PASS). Both (S,R,A) terraces (Wil, Alt) beat every non-terrace arm on 7/7 backbones; Alt edges Wil in unrounded risk on 5/7 backbones (\S\ref{sec:boundaries}). $^\dagger$Audited prefill-only blockwise PINE variant (inapplicable to 4/7 hybrid/recurrent models). Synchronized $p95$ latency ratio W/A is 0.999--1.005$\times$ on 7/7 backbones (Table~\ref{tab:app_cora_latency}).}
\label{tab:main_k6}
\end{table*}

\begin{table*}[t]
\centering
\scriptsize
\setlength{\tabcolsep}{2.4pt}
\begin{tabular}{@{}l@{\;\;}ccccccccc@{\;\;}cc@{\;\;}cccc@{\;\;}cc@{}}
\toprule
 & \multicolumn{9}{c}{\textbf{Judge Accuracy (\%) $\uparrow$}} & \multicolumn{2}{c}{\textbf{Flip ($m{=}2$) $\downarrow$}} & \multicolumn{4}{c}{\textbf{$\mathrm{BrierEx}/E_{\mathrm{tot}}$ (\%) $\downarrow$}} & \textbf{CAL} & \textbf{Gold Brier} \\
\cmidrule(lr){2-10}\cmidrule(lr){11-12}\cmidrule(lr){13-16}\cmidrule(lr){17-17}\cmidrule(l){18-18}
Judge ($N{=}500$) & Base & PriDe & PINE$^\dagger$ & RevP & AdjR & Anti & Cyc & $E_4$ & \textbf{Wil} & RevP & AdjR & Anti & Cyc & $E_4$ & \textbf{Wil} & $\hat\Gamma_4$ & $\Delta(\text{A}-\text{W})$ \\
\midrule
Qwen3.8-27B$^\blacktriangle$ & 83.9 & 83.9 & n/a & 85.2 & 85.0 & 86.6 & 86.8 & 87.2 & \textbf{87.8} & 9.5 & 10.6 & 16.2 & 10.3 & 5.4 & \textbf{4.9} & 0.53 & +0.0043$^*$ \\
Gemma-4-31B$^\bullet$ & 82.2 & 82.6 & 84.2 & 84.4 & 84.0 & \textbf{85.2} & 84.5 & 84.9 & 85.2 & 7.1 & 10.5 & 12.4 & 10.2 & 8.6 & \textbf{7.7} & 0.75 & +0.0050$^*$ \\
Llama-3.1-8B$^\bullet$ & 49.1 & 54.4 & 45.8 & 58.4 & 53.7 & 60.5 & 61.9 & \textbf{62.5} & 62.0 & 34.1 & 41.4 & 11.4 & 6.0 & 4.1 & \textbf{3.3} & 0.48 & +0.0162$^*$ \\
\textit{Olmo-Hybrid-7B}$^\blacktriangle$ & 55.1 & 59.9 & n/a & 58.4 & 60.7 & 63.0 & 65.6 & \textbf{66.8} & 66.7 & 36.8 & 37.3 & 21.4 & 10.8 & 5.9 & \textbf{5.6} & 0.59 & \textit{+0.0232} \\
\textit{Olmo-3-7B}$^\bullet$ & 42.8 & 44.3 & 46.8 & 47.4 & 46.1 & 49.7 & 51.6 & 51.9 & \textbf{52.6} & 38.6 & 47.3 & 13.0 & 8.7 & 6.1 & \textbf{5.0} & 0.61 & \textit{+0.0205} \\
\midrule
\multicolumn{18}{@{}p{0.97\textwidth}@{}}{{\textbf{Cross-prompt transfer ($N{=}256$, gate \texttt{E3\_judge\_transfer}: FAIL):} held-out $\Delta\mathrm{MSE}$ (total $-$ prec.) $= -1.63\times 10^{-6}$ (Holm $p{=}0.879$). Descriptive Spearman (prec./tot.): Qwen3.8 0.561/0.542, Gemma-4 0.206/0.190, Llama-3.1 0.316/0.290.}} \\
\bottomrule
\end{tabular}
\caption{\textbf{Transfer to listwise LLM-as-judge evaluation on RewardBench 2 ($K{=}4$, exact over all $4!{=}24$ orders, $N{=}500$).} Two-pass designs (RevP = ReversalPair, AdjR = AdjRot) are separated from four-pass designs (Anti, Cyc, $E_4$, Wil; a fifth $m{=}4$ arm, Random-WoR, has exact $\mathrm{BrierEx}/E_{\mathrm{tot}}{=}5/23{=}21.7\%$ and does not alter any best-arm bolding; best unrounded $m{=}4$ accuracy and lowest exact $\mathrm{BrierEx}/E_{\mathrm{tot}}$ are \textbf{bold}). Accuracy and Flip are averaged over 4 sampled anchors; $\mathrm{BrierEx}/E_{\mathrm{tot}}$ and Gold Brier $\Delta(\text{A}-\text{W})$ are exact over all 24 orders (where the common gold term cancels). Williams leaves the strictly lowest exact order energy on 5/5 judges, reduces Gold Brier vs.\ Antithetic on 3/3 preregistered judges (all Holm $p{=}0.0003$; italic Olmo rows descriptive, raw $p{<}5{\times}10^{-5}$), and passes $-2$\,pp NI vs.\ PriDe and audited blockwise PINE$^\dagger$ on Llama-3.1 (Holm $p{=}0.0003$). Gate \texttt{E3\_judge\_transfer} fails because cross-prompt $\Delta\mathrm{MSE}$ does not reject.}
\label{tab:judge}
\end{table*}

\paragraph{Fidelity.} Table~\ref{tab:main_k6} reports all evaluated arms on six-option MMLU-Pro. Random-WoR leaves 16.0--16.9\% of order energy across backbones, matching $\rho^R_6{=}16.55\%$ (Proposition~1). The two (S,R,A) reversal-closed terraces occupy the lowest risk tier on all seven backbones (Williams 2.0--6.6\%, Alt\_Terrace 1.9--6.8\%), beating every non-terrace design (Random-WoR, Antithetic, Cyclic, $E_6$) on 7/7 backbones; Alt\_Terrace achieves slightly lower unrounded point risk than Williams on 5/7 backbones at $K{=}6$ and 14/14 cells at $K{\in}\{8,10\}$ (\S\ref{sec:boundaries}). On the frontier models, Williams cuts Brier excess versus Cyclic by 61.2\% (absolute 0.00497, 95\% CI $[0.00422,0.00577]$) on Qwen3.8 and 40.0\% (0.00830, $[0.00614,0.01062]$) on Gemma-4, and versus Antithetic by 56.5\% and 51.0\% (all Holm $p{=}0.0004$, all ${\ge}20\%$; gate \texttt{E2\_K6\_frontier\_fidelity}: PASS). Williams also beats $E_6$ by 10.9\% and 10.0\% (one-sided Holm $p{\le}0.032$), a whole-design contrast (Theorem~2(i)). Across all seven backbones, Williams lowers $\mathrm{BrierEx}/E_{\mathrm{tot}}$ versus Cyclic (7/7, Holm $p{=}0.00035$). At $K{\in}\{8,10\}$ ($N{=}200$/cell), Williams beats Cyclic and Antithetic on all 8/8 frontier contrasts (Holm $p{\le}0.002$; Table~\ref{tab:app_k810}); adjusted inference covers the frontier pair only.

\paragraph{Flips and accuracy.} Williams cuts anchor-to-anchor flips by 4.3--7.8\,pp versus Cyclic on every backbone (7/7, Holm $p{=}0.00035$) and raises all-4-anchor strict accuracy (\texttt{W4}) by $+1.6$ to $+5.0$\,pp across all seven backbones (Table~\ref{tab:app_k6_full}). Mean accuracy moves more modestly: across all six evaluated $m{=}6$ arms, Williams leads on 3/7 backbones (2/7 on CONF $N{=}500$ alone; Cyclic leads Olmo-3, $E_6$ edges Qwen3.5, and Alt\_Terrace edges Nemotron-3 and Llama-3.1). Pooled Williams minus Cyclic accuracy is $+0.69$\,pp across 500 item clusters (3{,}500 model-items; 95\% CI $[+0.01,+1.36]$, Holm $p{=}0.0198$), but sign-flip $p{=}0.0266$ misses the locked $0.01$ gate and no per-model accuracy contrast survives Holm (0/7). On pooled $N{=}2{,}000$, Williams beats Antithetic on Gemma-4 ($+1.56$\,pp, Holm $p{=}0.00025$; Qwen3.8 $+0.49$\,pp, Holm $p{=}0.071$) and passes $-1$\,pp non-inferiority versus PriDe on both frontier models and versus PINE$^\dagger$ on Gemma-4.

\paragraph{Audit capture and cost.} Consistent with Corollary~A, rotation audits already capture 86--95\% of total order energy (missing 5--14\%), and Williams recovers this blind spot to 91--98\% on all seven backbones. Online differences between stored $\kappa$ and $1{-}\mathrm{BrierEx}/E_{\mathrm{tot}}$ (Gemma-4 Williams: 0.9135 vs.\ 0.9335) stem from finite-sample estimation variance; on exact arms the duality residual is $\le 3.3{\times}10^{-16}$ (Table~\ref{tab:app_cora_latency}). With fixed batching, Williams-to-Antithetic $p95$ latency ratios are 0.999--1.005 across all seven backbones.

\paragraph{Judges.} On four-candidate RewardBench~2 (Table~\ref{tab:judge}), Williams leaves the strictly lowest exact order energy for all five judges and lowers gold Brier versus Antithetic on 3/3 registered judges by 0.0043, 0.0050, and 0.0162 (Holm $p{=}0.0003$; both descriptive Olmo judges agree). By Proposition~1 the gold term cancels on exact arms, so these measure fidelity rather than independent accuracy gain. Across all four $m{=}4$ arms (including $E_4$), Williams leads accuracy on 2/5 judges ($E_4$ edges Llama-3.1 and Olmo-Hybrid; Antithetic edges Gemma-4 by 0.05\,pp) and passes $-2$\,pp NI versus PriDe and PINE$^\dagger$ on Llama-3.1 (Holm $p{=}0.0003$). Restricting to order-sensitive items (\texttt{full\_order\_flip}, 164--469/500), Williams cuts flips vs.\ Cyclic on all five judges (Qwen3.8 12.0$\to$8.7\%, $+3.01$\,pp acc; energy recall 89.7$\to$95.1\% vs.\ \texttt{flip\_recall} 73.5$\to$73.8\%; Gemma-4 16.0$\to$10.5\%, $+1.98$\,pp), whereas Llama-3.1 flips drop (23.5$\to$19.7\%) with near-zero accuracy gain ($+0.06$\,pp), showing stability does not imply correctness. At $m{=}2$, ReversalPair flips less than AdjRot on 5/5 judges. Every judge has $\hat\Gamma_4{<}1$ on CAL and Cyclic beats Antithetic on CONF (Corollary~B). Gate \texttt{E3\_judge\_transfer} fails ($\Delta\mathrm{MSE}{=}-1.63{\times}10^{-6}$, Holm $p{=}0.879$, $N{=}256$).

\subsection{Frozen spectral calls versus confirmatory outcomes}\label{sec:frozen}

Before CONF decoding, Freeze-B turned CAL energies into conservative envelopes (e.g., $[E_{\mathrm{curl}},E_{\mathrm{tot}}-\EU]$ for Cyclic), sign calls wherever bounds exclude zero, and the first $K$ where Cyclic loses (Figure~\ref{fig:hero}b,c; Figure~\ref{fig:app_regimes}; Appendix~\ref{app:gates}).

\paragraph{Signs (S1, PASS) and crossovers (S4, PASS).} All 13 issued sign calls were correct, but coverage was only 13/84 (15.5\%, 71 abstentions; issued calls: 6 Cyclic$-$Antithetic, 7 Cyclic$-$Random-WoR at $K{\le}6$, 0 Williams$-$Random-WoR). Cyclic loses to Antithetic on 6/7 backbones on the grid, and frozen crossovers land within $\pm 1$ step on 5/7 models (all five one step early; Random-WoR hits 4/7; Table~\ref{tab:app_crossovers}). Post-hoc difficulty references show these passes do not establish incremental predictive value over simple heuristics: on S1, the ablated rule scores 17/17 on its own calls and ``Cyclic~$<$~Random-WoR at $K{\le}6$'' scores 14/14; on S4, a constant $K^*{\equiv}8$ hits 6/7 within $\pm 1$ step.

\paragraph{Point ranking (S2, FAIL), ablation (S3, FAIL), and envelope diagnosis.} Across 112 cells, frozen-midpoint Spearman $\rho{=}0.190$ trails design-agnostic B1 ($\rho_K^R$ per $K$, $0.328$), beats B2 (Cyclic$=$Williams midpoint, $0.117$), and has partial $\rho{=}0.166{<}0.30$; S3 drop is $0{<}0.15$ (8/8 vs.\ 8/8). Because production CAL did not log $E_{U_+}$ or $E_{\mathrm{sa}}$, Freeze-B defaulted Williams' upper bound to valid-but-loose $E_{\mathrm{even}}/E_{\mathrm{tot}}$ rather than $E_{\mathrm{he}}/E_{\mathrm{tot}}$ (Theorem~2(f)), inflating Williams midpoints by median $3.925\times$ (28/28 cells; 0/28 W$-$R calls). Cyclic's midpoint likewise overestimated $R_C$ across 28/28 cells (median $1.631\times$), shifting all five S4 hits one step early and placing all 10 sign-mismatched Figure~\ref{fig:hero}c points in the lower-right quadrant. Small-probe bounds serve only as conservative envelopes, never point rankings.

\subsection{Mechanism: curl, the energy ledger, and a matched pair}\label{sec:mechanism}

\begin{figure*}[t]
\centering
\includegraphics[width=\linewidth]{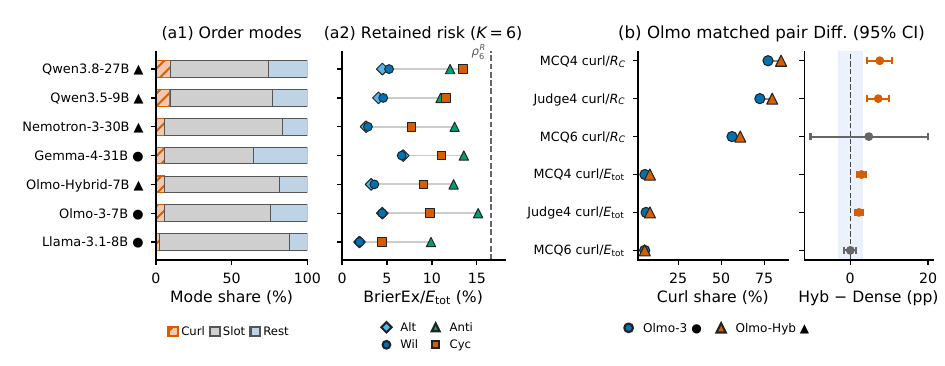}
\caption{\textbf{CONF mode shares, zoomed design risks, and same-lab matched pair.} (a1) Stacked $K{=}6$ CONF mode shares (curl, slot $E_U$, remainder; $N{=}500$, $\blacktriangle$ hybrid/recurrent, $\bullet$ dense). (a2) Zoomed retained risk $\mathrm{BrierEx}/E_{\mathrm{tot}}$ for Alt\_Terrace, Williams, Antithetic, and Cyclic with $\rho^R_6{=}16.55\%$. (b) Olmo-Hybrid minus Olmo-3 curl-share differences with 95\% percentile CIs (band: $\pm 3$\,pp TOST margin); hybrid exceeds dense in 4/6 Holm tests (MCQ4, Judge4), fixing the lab without isolating architecture.}
\label{fig:ledger}
\end{figure*}

\paragraph{Exact curl test (E1).} Across 192 matched items decoded under all $K!$ orders ($K{\in}\{4,6\}$; Table~\ref{tab:app_e1_exact}, Figure~\ref{fig:app_e1_three}), $q_K{=}E_{\mathrm{curl}}/(E_{\mathrm{curl}}{+}E_{\mathrm{grad}})$ rises on Qwen3.8 ($\Delta q{=}+0.0367$, CI $[0.013,0.058]$, Holm $p{=}0.0136$), not on Gemma-4 ($-0.020$) or Llama-3.1 ($-0.043$), with no $2/9$ equivalence. Qwen3.8's $K{=}6$ IUT components match the curl direction ($C{-}A{=}+0.00079$, raw $p{=}0.0243$; $C{-}\mathrm{curl}{-}A{=}-0.00402$, $p{<}5{\times}10^{-5}$), yet Holm IUT $p{=}0.1458$ fails \texttt{E1\_Qwen\_IUT} (no confirmatory causal claim). Descriptively, $s_{\mathrm{curl}}{=}E_{\mathrm{curl}}/(R_C{-}R_W)$ is 106.7\%, 139.4\%, and 109.4\% across the three models, exceeding 100\% because non-curl residuals partly offset curl removal ($R_W{>}R_C{-}E_{\mathrm{curl}}$; not a causal test).

\paragraph{Energy ledger and matched pair (E4).} Figure~\ref{fig:ledger}a1--a2 shows both terraces sitting below Antithetic (9.9--15.1\%) and Cyclic (4.5--13.5\%) on 7/7 backbones, while Cyclic loses to Antithetic at $K{=}6$ only on Qwen3.8 and Qwen3.5. On the same-lab Olmo pair (Figure~\ref{fig:ledger}b), Olmo-Hybrid exceeds Olmo-3 in curl share on MCQ4 ($+7.60$\,pp of $R_C$, $+2.89$\,pp of $E_{\mathrm{tot}}$) and Judge4 ($+7.20$\,pp, $+2.25$\,pp; all Holm $p{=}0.0003$), with no detectable MCQ6 difference ($+0.02$\,pp, CI $[-1.57,+1.60]$; separate $\pm 0.03$ TOST equivalence); because the pair differs beyond layer type, this is not causal architecture attribution.

\subsection{Preregistered gate ledger}\label{sec:ledger}

Three of eight gates pass (\texttt{E2\_K6\_\allowbreak frontier\_\allowbreak fidelity}, S1, S4) and five fail (Table~\ref{tab:app_gates}): \texttt{E1\_Qwen\_IUT} ($p{=}0.1458$), \texttt{E2\_pooled\_\allowbreak accuracy} (sf $p{=}0.0266$), \texttt{E3\_judge\_\allowbreak transfer} ($p{=}0.879$), S2 (partial $\rho{=}0.166$), and S3 (drop $0$). Across ten Holm families, 45/62 hypotheses reject; confirmatory claims rest strictly on passed fidelity gates.

\subsection{Boundaries and terrace design}\label{sec:boundaries}

\paragraph{Exact invariance is a different target.} PINE$^\dagger$ alters the reader, outside Proposition~1. On the three dense models it lowers MCQ6 accuracy below Base (Gemma-4 69.8$\to$63.4\% pooled, 70.3$\to$63.4\% CONF; Llama-3.1 38.9$\to$31.8\%; Olmo-3 39.2$\to$35.8\%) and flips on 32.4--59.8\% of anchor pairs due to letter-prefixed relabeling; serial execution makes it 1.87--3.81$\times$ slower at $p50$ (1.06--2.31$\times$ at $p95$) than batched Williams (Appendix~\ref{app:pine_latency}). One-pass PriDe improves over Base on 5/7 models at $1/6$ passes but trails every $m{=}6$ design.

\paragraph{Why Alt\_Terrace edges zigzag $b_W$, and odd $K$.} All 24 $K{=}6$ reversal-closed terraces form exactly two subgroups (Williams and Alt\_Terrace, 12 seeds each; E1 exact ranges $2.59$--$2.63$, $11.98$--$12.29$, $3.60$--$3.63{\times}10^{-3}$; Table~\ref{tab:app_e1_exact}). Classic zigzag $b_W$ collapses 100\% of distance-2 differences $b_W(s{+}2){-}b_W(s)\pmod K$ onto $\{\pm 1\}$ ($96/111{=}86.5\%$ trace retention on the 9-D distance-2 subspace $\mathcal H_{\mathrm{d2}}{\subset}\mathcal H_{\mathrm{he}}$ at $K{=}6$), whereas Alt\_Terrace disperses them ($16/111{=}14.4\%$ at $K{=}6$, $6.0\times$ lower; all 8 non-antipodal residues at $K{=}10$), a testable post-hoc explanation for Alt\_Terrace's edge on 19/21 CONF cells at $K{\ge}6$ (E1 Llama is a slight counterexample, not a causal proof). Evidence covers even $K{\le}10$; odd $K$ requires $2K{\mid}m$ for (S)$\wedge$(R) (Theorem~2(e)), left to future work.

\section{Related Work}\label{sec:related}

\paragraph{Order sensitivity.} Position and selection biases affect MCQ readers \citep{zheng2024large,li2026position}, few-shot prompts \citep{lu2022fantastically,zhao2021calibrate}, long contexts \citep{liu2024lost}, listwise rankers \citep{sun2023chatgpt,qin2024large,tang2024permutation} (with mechanistic Transformer accounts in \citealp{wu2025emergence}), and LLM judges \citep{zheng2023judging,wang2024fair,shi2025judging,li2025calibraeval,zhu2025judgelm}. Rather than measuring or post-calibrating bias, we characterize which order modes each averaging or audit design sees.

\paragraph{Invariance by construction.} PINE modifies attention \citep{wang2025eliminating}; Set-LLM, PEARL, PICASO, parallel context windows, and stable RAG pursue architectural or training invariance \citep{egressy2025setllm,chen2025pearl,liu2025picaso,chu2026order,ratner2023parallel,lee2025cord,zhang2026stablerag,bito2026one}. Editing the reader and averaging a fixed reader target different quantities (\S\ref{sec:boundaries}).

\paragraph{Group averaging and designs.} Group/frame averaging and permutation self-consistency average over symmetries \citep{chen2020group,lyle2020benefits,puny2022frame,murphy2019janossy,tang2024permutation}. We combine $S_K$ harmonic analysis \citep{diaconis1988group} and Hodge ranking decompositions \citep{jiang2011statistical} with Williams squares \citep{williams1949experimental}, quasi-complete Latin squares \citep{bailey1984quasi}, and order-of-addition designs \citep{lin2019order,peng2019design,voelkel2019design,yang2021component,stokes2022position}. Closest to us, \citet{li2026trustworthy} apply Williams designs to LLM judges; we establish exact blind-spot spectral theory, audit-averaging duality, the rotation crossover criterion, and the distance-2 tradeoff inside the terrace class.

\section{Conclusion}\label{sec:conclusion}

Rotation is the default for averaging over and auditing option order, yet it is blind to precedence cycles. We gave an exact spectral theory of equal-cost designs, proved audit-averaging duality, and showed that reversal-closed, adjacency-balanced terraces remove every precedence cycle at rotation's cost for even $K$. Across seven open backbones in an internally sealed preregistered study, Williams cut Brier excess and flips vs.\ rotation on 7/7 models, retained frontier fidelity gains at $K{\in}\{8,10\}$, and left the least exact order energy in listwise judging. Practically: (1) with a uniform random anchor, replace $K$ rotations with a $K$-pass (S,R,A) terrace ($b_W$, or Alt\_Terrace at $K{\ge}6$; for odd $K$, use $2K$-pass reversal-paired rotation) at zero extra cost for even $K$; (2) do not certify order robustness with a rotation audit (Corollary~A); and (3) treat small-probe bounds as conservative envelopes, not point rankings.

\section*{Limitations}

Fidelity measures proximity to each model's exhaustive order average, not correctness. Accuracy gains of Williams over rotation are directionally positive but unconfirmed (pooled gate failed; 0/7 per-model Holm), and Llama-3.1 judge error analysis shows reduced flips need not raise accuracy. The curl mechanism is likewise unconfirmed: the Holm-adjusted IUT failed, $q_K$ rose from $K{=}4$ to $6$ only on Qwen3.8, $E_K$ versus Williams is a whole-design contrast, and the Olmo pair does not isolate architecture. The frozen probe had 15.5\% sign coverage, failed S2 and S3, and did not outperform post-hoc heuristics on S1/S4 because Freeze-B defaulted to the loose $E_{\mathrm{even}}/E_{\mathrm{tot}}$ bound. Our PINE$^\dagger$ baseline is an audited prefill-only blockwise variant, not a validated reference; it covers 3/7 backbones, exhibits group-level suffix lookahead, is non-invariant under letter-prefixed relabeling, and has implementation-specific latency. Judge gold-Brier contrasts on exact arms are not independent of fidelity, cross-prompt transfer failed, and Gemma-4's mean allowed-label mass on RewardBench~2 is 0.303 (0.226 alt prompt), limiting extrapolation to free-form CoT judges. Evidence covers even $K{\le}10$ (odd $K$ requires $2K{\mid}m$ for (S)$\wedge$(R)), English MMLU-Pro and RewardBench~2, seven open-weight backbones, 2 sampled anchors at $K{\in}\{8,10\}$ (4 at $K{\in}\{4,6\}$), and one hardware setup. Internal hash seals establish content identity across records, not external third-party timestamps or Freeze-A.1 ordering relative to $K{=}6$ CAL decoding.

\section*{Ethical Considerations}

Order effects in LLM judges can systematically favor responses by where they appear, which matters when such judges rank systems, grade students or filter content. Our designs reduce order-induced variance without retraining and at no extra cost over rotation, but they average a model's preferences rather than correct them: a lower audit disagreement is not a certificate of fairness or accuracy, and Corollary~A shows that audits share the blind spots of the design they are computed on. We use only public benchmarks under their licenses and involve no human subjects or personal data. Compute is bounded by $K$ forward passes per item, the same as standard rotation.

\bibliography{references}

\clearpage
\appendix
\section{Complete Proofs and Algebraic Verification}
\label{app:proofs}

% ---- Appendix-local helpers (\providecommand never overrides main.tex) ----
\providecommand{\HU}{\mathcal{H}_{U}}
\providecommand{\Hou}{\mathcal{H}_{\mathrm{ou}}}
\providecommand{\EU}{E_{U}}
\providecommand{\apH}[1]{\mathcal{H}_{\mathrm{#1}}}
\providecommand{\apBX}{\mathrm{BrierEx}}
\providecommand{\apthmname}{}
\newtheorem{aplemma}{Lemma}[section]
\newtheorem*{apnamedthm}{\apthmname}
\newenvironment{apnamed}[1]{\renewcommand{\apthmname}{#1}\begin{apnamedthm}}{\end{apnamedthm}}
\theoremstyle{remark}
\newtheorem*{apremark}{Remark}
\theoremstyle{plain}
\setlength{\emergencystretch}{2em}

This appendix gives complete proofs of every formal statement in the main text. Numerical checks of these statements (\S\ref{app:verify}) read archived exploratory tensors read-only and run no model inference. They supplement the proofs and do not replace them.

\subsection{Setting and Notation}
\label{app:setting}

\paragraph{Orders.} Fix $K\ge 2$ and $[K]=\{0,\dots,K-1\}$. An \emph{order} is $\pi\in S_K$; $\pi(s)$ is the item in slot $s$ and $\pi^{-1}(x)$ is the slot of item $x$. Composition is $(\pi g)(s)=\pi(g(s))$. Let $c(s)=s+1 \bmod K$ and $r(s)=K-1-s$. Then $\pi c=(\pi(1),\dots,\pi(K-1),\pi(0))$ is the left rotation, $\pi r$ is the reversed list, and $rcr=c^{-1}$.

\paragraph{Readouts and inner product.} A readout is $f:S_K\to\Delta^{C-1}\subset\mathbb{R}^C$ in \emph{canonical candidate coordinates}: candidate $v$ keeps index $v$ whatever slot displays it. The theory needs only $f\in L^2(S_K;\mathbb{R}^C)$; the simplex is used only for the Brier interpretation and one dimension count. Define
\[
\langle u,v\rangle=\tfrac{1}{K!}\textstyle\sum_{\pi}\langle u(\pi),v(\pi)\rangle_{\mathbb{R}^C}.
\]
Linear operators on scalar $L^2(S_K)$ act coordinatewise, so $\|P_Af\|^2=\sum_v\|P_Af_v\|^2$. Let $\mu=\frac{1}{K!}\sum_\pi f(\pi)$, $f_0=f-\mu$ and $E_{\mathrm{tot}}=\|f_0\|^2$.

\paragraph{Right action.} $(R_gf)(\pi)=f(\pi g)$. Then $R_gR_h=R_{gh}$, each $R_g$ is unitary and $R_g^{*}=R_{g^{-1}}$. For a subgroup $H\le S_K$ put $P_H=\frac{1}{|H|}\sum_{h\in H}R_h$, $P_C=P_{\langle c\rangle}$ and $P_\pm=\frac12(I\pm R_r)$.

\paragraph{Designs.} A design $D$ of size $m$ is a law on $m$-tuples of \emph{relative rows} $(d_1,\dots,d_m)$ together with an anchor $\sigma\sim\mathrm{Unif}(S_K)$ independent of the rows. The evaluated orders are $\sigma d_j$ and $\hat\mu_D(\sigma)=\frac1m\sum_j f(\sigma d_j)$. The arms are:
\begin{itemize}
\item \emph{Deterministic.} Cyclic $(c^t)_{t<K}$; the seeded family $D_b=(c^tb)_{t<K}$, with Williams $b=b_W$ and Alt\_Terrace $b=b_A$; and $E_K=\{c^{2j},c^{2j}r\}_{j<K/2}$.
\item \emph{Stochastic.} Random-WoR: $\{\sigma d_j\}$ is a uniform $m$-subset of $S_K$. Antithetic ($m=K$ even, $s=K/2$): $\{\sigma d_j\}$ is the union of $s$ distinct reversal pairs $\{\pi,\pi r\}$, and the set of pairs is uniform among $s$-subsets of the $M=K!/2$ pairs.
\end{itemize}
The pooled row law is $q_D(\pi)=\mathbb{E}\big[\frac1m\sum_j\mathbf{1}\{\sigma d_j=\pi\}\big]$, and $\apBX(D)=\mathbb{E}\|\hat\mu_D-\mu\|^2$, with the expectation taken over anchor and rows.

\paragraph{Mode subspaces (scalar).} Let $L^2_0=\{g:\sum_\pi g(\pi)=0\}$ and let $P_0$ project onto constants.
\begin{itemize}
\item \emph{Unary.} $u_{x,s}(\pi)=\mathbf{1}[\pi(s)=x]$ and $\mathcal{U}=\mathrm{span}\{u_{x,s}\}$, which contains the constants. Put $\apH{un}=\mathcal{U}\cap L^2_0$ with projector $P_{\mathrm{un}}$, $\HU=P_+\apH{un}$ and $\Hou=P_-\apH{un}$. Lemma~\ref{lem:app_comm} shows that both are subspaces of $\apH{un}$.
\item \emph{Precedence.} For $i\ne j$, $\phi_{ij}(\pi)=\operatorname{sgn}(\pi^{-1}(j)-\pi^{-1}(i))$, so $\phi_{ji}=-\phi_{ij}$. For skew $B\in\mathbb{R}^{K\times K}$ put $f_B=\sum_{i<j}B_{ij}\phi_{ij}$, with row sums $\varsigma_i(B)=\sum_jB_{ij}$.
\item \emph{Gradient.} $s_i=\sum_{j\ne i}\phi_{ij}=K-1-2\pi^{-1}(i)$ and $\apH{grad}=\mathrm{span}\{s_i\}$.
\item \emph{Curl.} $\apH{curl}=\{f_B: B \text{ skew},\ \varsigma(B)=0\}$.
\item \emph{Symmetric adjacency.} $\alpha_{ij}(\pi)=\mathbf{1}[|\pi^{-1}(i)-\pi^{-1}(j)|=1]-2/K$ and $\apH{sa}=(I-P_0-P_{\mathrm{un}})\,\mathrm{span}\{\alpha_{ij}\}$.
\item \emph{Remainders.} $\apH{ou,rest}=\Hou\ominus\apH{grad}$, $\apH{he}=P_+L^2_0\ominus(\HU\oplus\apH{sa})$ and $\apH{ho}=P_-L^2_0\ominus(\Hou\oplus\apH{curl})$.
\end{itemize}
Lemma~\ref{lem:app_decomp} proves the orthogonal decomposition
\begin{align*}
P_+L^2_0&=\HU\oplus\apH{sa}\oplus\apH{he},\\
P_-L^2_0&=\apH{grad}\oplus\apH{ou,rest}\oplus\apH{curl}\oplus\apH{ho}.
\end{align*}
For even $K\ge 4$ the dimensions are $\dim\apH{un}=(K-1)^2$, $\dim\HU=(K-1)(K/2-1)$, $\dim\Hou=(K-1)K/2$, $\dim\apH{grad}=K-1$, $\dim\apH{curl}=\binom{K-1}{2}$ and $\dim\apH{sa}=K(K-3)/2$. All of them were verified at $K=4,6$.

\paragraph{Components and energies.} Let $f_\pm=P_\pm f_0$, and write $f_+=u_++a+h_+$ and $f_-=g+u_-^{\perp}+w_{\mathrm{curl}}+h_-$ for the orthogonal projections onto the seven pieces above. The energies are $E_{U_+}=\|u_+\|^2$, $E_{\mathrm{sa}}=\|a\|^2$, $E_{\mathrm{he}}=\|h_+\|^2$, $E_{\mathrm{grad}}=\|g\|^2$, $E_{\mathrm{curl}}=\|w_{\mathrm{curl}}\|^2$, $E_{\mathrm{ho}}=\|h_-\|^2$, $E_{\mathrm{even}}=\|f_+\|^2$, $E_{\mathrm{odd}}=\|f_-\|^2$ and $E_{\mathrm{un}}=\|P_{\mathrm{un}}f_0\|^2$.

\paragraph{Notation: $\HU$ versus $\EU$.}
In the paper, $\HU$ is the \emph{reversal-even} unary (slot) subspace $P_+\apH{un}$, with energy $E_{U_+}=\|u_+\|^2$. The logged field \texttt{E\_U} in our JSON artifacts, written $\EU$ in tables and figures, is a different quantity: the \emph{full} centred unary energy
\[
\EU\equiv E_{\mathrm{un}}=E_{U_+}+E_{\mathrm{grad}}+E_{\mathrm{ou,rest}},
\]
the energy of $f_0$ in $\apH{un}=\HU\oplus\Hou$, of both reversal parities, where $\Hou=\apH{grad}\oplus\apH{ou,rest}$. Other logged fields map directly: \texttt{E\_grad}$\,\to E_{\mathrm{grad}}$, \texttt{E\_curl}$\,\to E_{\mathrm{curl}}$, \texttt{E\_even}$\,\to E_{\mathrm{even}}$, \texttt{E\_tot}$\,\to E_{\mathrm{tot}}$ and \texttt{R\_C}$\,\to\|P_Cf_0\|^2$.

\subsection{Auxiliary Lemmas}
\label{app:lemmas}

\begin{aplemma}[Commutation]\label{lem:app_comm}
(a) $R_rP_C=P_CR_r$.
(b) $R_r$ preserves $\mathcal{U}$, $\apH{un}$, $\mathrm{span}\{\phi\}$, $\apH{grad}$, $\apH{curl}$, $\mathrm{span}\{\alpha\}$ and $\apH{sa}$. Moreover $R_r=-I$ on $\mathrm{span}\{\phi\}$, and $R_r=+I$ on $\mathrm{span}\{\alpha\}$ and on $\apH{sa}$.
(c) The orthogonal projectors onto all of these subspaces commute with $R_r$.
\end{aplemma}
\begin{proof}
(c) Let $V$ be invariant under the unitary involution $R_r$. For $w\in V^\perp$ and $v\in V$, $\langle R_rw,v\rangle=\langle w,R_rv\rangle=0$, so $V^\perp$ is also invariant. Writing $x=P_Vx+(I-P_V)x$ and applying $R_r$ gives $R_rP_Vx\in V$ and $R_r(I-P_V)x\in V^\perp$, hence $P_VR_rx=R_rP_Vx$.

(a) $R_rR_{c^t}R_r=R_{rc^tr}=R_{c^{-t}}$. Summing over $t\in\mathbb{Z}_K$ and dividing by $K$ gives $R_rP_CR_r=P_C$. Multiplying on the right by $R_r$ (using $R_r^2=I$) gives $R_rP_C=P_CR_r$.

(b) $(R_ru_{x,s})(\pi)=\mathbf{1}[\pi(K-1-s)=x]=u_{x,K-1-s}(\pi)$, so $R_r$ permutes a spanning set of $\mathcal{U}$. It also fixes constants, so it preserves $\mathcal{U}$ and $\apH{un}=\mathcal{U}\cap L^2_0$. Since $(\pi r)^{-1}=r\pi^{-1}$ and $r$ reverses order, $\phi_{ij}(\pi r)=-\phi_{ij}(\pi)$; hence $R_r=-I$ on $\mathrm{span}\{\phi\}$, and in particular on $\apH{grad}$ and $\apH{curl}\subseteq\mathrm{span}\{\phi\}$. Adjacency is preserved because $|r(p)-r(q)|=|p-q|$, so $R_r\alpha_{ij}=\alpha_{ij}$. For $\apH{sa}$, part (c) applied to $\mathcal{U}$ and to the constants gives $R_r(I-P_0-P_{\mathrm{un}})=(I-P_0-P_{\mathrm{un}})R_r$. Therefore $R_r$ acts as $+I$ on $(I-P_0-P_{\mathrm{un}})\mathrm{span}\{\alpha\}$.
\end{proof}

\begin{aplemma}[Precedence Gram identity]\label{lem:app_gram}
For distinct $i,j,k,l$: $\mathbb{E}\phi_{ij}\phi_{ik}=\frac13$, $\mathbb{E}\phi_{ij}\phi_{jk}=-\frac13$ and $\mathbb{E}\phi_{ij}\phi_{kl}=0$. For every skew $B$,
\[
\|f_B\|^2=\tfrac13\Big[\textstyle\sum_{i<j}B_{ij}^2+\sum_i\varsigma_i(B)^2\Big].
\]
\end{aplemma}
\begin{proof}
Under the uniform law, the relative order of three fixed items is uniform over its $3!$ patterns.
\begin{itemize}
\item $\phi_{ij}\phi_{ik}=+1$ iff $i$ is first or last among $\{i,j,k\}$, which has probability $2/3$. So the mean is $\frac23-\frac13=\frac13$.
\item $\phi_{ij}\phi_{jk}=+1$ iff the order is $i<j<k$ or $k<j<i$, which has probability $1/3$. So the mean is $\frac13-\frac23=-\frac13$.
\item For disjoint pairs, relabel items by the transposition $(k\,l)$, i.e.\ $\pi\mapsto(k\,l)\circ\pi$, which preserves the uniform law. It fixes $\phi_{ij}$ and negates $\phi_{kl}$, so the mean equals its own negative and is $0$.
\end{itemize}
Write $f_B=\frac12\sum_{i\ne j}B_{ij}\phi_{ij}$ and expand
\[
\|f_B\|^2=\tfrac14\textstyle\sum_{i\ne j}\sum_{k\ne l}B_{ij}B_{kl}\,\mathbb{E}\phi_{ij}\phi_{kl}.
\]
\emph{Same unordered pair.} Each ordered $(i,j)$ meets $(i,j)$ (mean $1$) and $(j,i)$ (mean $-1$, with $B_{ji}=-B_{ij}$). The contribution is $\frac14\cdot 2\sum_{i\ne j}B_{ij}^2=\sum_{i<j}B_{ij}^2$.

\emph{Exactly one shared item.} By antisymmetry, $\mathbb{E}\phi_{ij}\phi_{il}=\frac13$, $\mathbb{E}\phi_{ij}\phi_{ki}=-\frac13$, $\mathbb{E}\phi_{ij}\phi_{jl}=-\frac13$ and $\mathbb{E}\phi_{ij}\phi_{kj}=\frac13$. Fix $(i,j)$. The partners sharing $i$ contribute $\frac13(\varsigma_i-B_{ij})-\frac13\cdot(-(\varsigma_i-B_{ij}))=\frac23(\varsigma_i-B_{ij})$, using $\sum_{k\ne i,j}B_{ki}=-(\varsigma_i-B_{ij})$. Likewise the partners sharing $j$ contribute $-\frac23(\varsigma_j-B_{ji})$. The total is
\begin{align*}
&\tfrac16\textstyle\sum_{i\ne j}B_{ij}\big[(\varsigma_i-B_{ij})-(\varsigma_j-B_{ji})\big]\\
&\quad=\tfrac16\Big[2\textstyle\sum_i\varsigma_i^2-4\sum_{i<j}B_{ij}^2\Big].
\end{align*}
Here we used $\sum_{i\ne j}B_{ij}\varsigma_i=\sum_i\varsigma_i^2$, $\sum_{i\ne j}B_{ij}\varsigma_j=-\sum_j\varsigma_j^2$ and $\sum_{i\ne j}B_{ij}B_{ji}=-2\sum_{i<j}B_{ij}^2$.

\emph{Disjoint pairs} contribute $0$.

Adding the three cases gives $\sum_{i<j}B_{ij}^2+\frac13\sum_i\varsigma_i^2-\frac23\sum_{i<j}B_{ij}^2$, which is the claim.
\end{proof}

\noindent\emph{Consequences.} (a) The form is at least $\frac13\|B\|^2>0$ for $B\ne 0$, so $\{\phi_{ij}\}_{i<j}$ is linearly independent and $\dim\mathrm{span}\{\phi\}=\binom K2$. (b) In edge coordinates $b=(B_{ij})_{i<j}$ the Gram operator is $\Gamma=\frac13(I+\mathsf{D}^\top\mathsf{D})$ with $\mathsf{D}:b\mapsto\varsigma$. Since $(\mathsf{D}^\top x)_{ij}=x_i-x_j$, we get $(\mathsf{D}\mathsf{D}^\top x)_i=Kx_i-\sum_jx_j=Kx_i$ when $\sum x=0$. Hence $\Gamma=\frac13$ on the cycle space $\ker\mathsf{D}$ (dimension $\binom{K-1}{2}$), and $\Gamma=\frac{K+1}{3}$ on gradient flows $B_{ij}=x_i-x_j$ (dimension $K-1$), because $\mathsf{D}^\top\mathsf{D}\mathsf{D}^\top x=K\mathsf{D}^\top x$.

\begin{aplemma}[Hodge split of precedence functions]\label{lem:app_hodge}
For skew $B$ set $B^{\mathrm g}_{ij}=(\varsigma_i-\varsigma_j)/K$ and $B^{\mathrm c}=B-B^{\mathrm g}$. Then:
(a) $\varsigma(B^{\mathrm c})=0$;
(b) $f_{B^{\mathrm g}}=\frac1K\sum_i\varsigma_is_i\in\apH{grad}\subseteq\Hou$;
(c) $f_B\perp\apH{un}\iff\varsigma(B)=0$;
(d) $\mathrm{span}\{\phi\}=\apH{grad}\oplus\apH{curl}$ orthogonally, $\apH{curl}\perp\apH{un}$, $\apH{curl}\subseteq P_-L^2_0$ and $\dim\apH{curl}=\binom{K-1}{2}$. Consequently $P_{\mathrm{grad}}f_B=f_{B^{\mathrm g}}$ and $P_{\mathrm{curl}}f_B=f_{B^{\mathrm c}}$.
\end{aplemma}
\begin{proof}
(a) $\sum_jB^{\mathrm g}_{ij}=\varsigma_i-\frac1K\sum_j\varsigma_j=\varsigma_i$, because $\sum_j\varsigma_j=\sum_{ij}B_{ij}=0$. Hence $\varsigma(B^{\mathrm c})=\varsigma(B)-\varsigma(B^{\mathrm g})=0$.

(b) $f_{B^{\mathrm g}}=\frac{1}{2K}\sum_{i\ne j}(\varsigma_i-\varsigma_j)\phi_{ij}=\frac{1}{2K}\big[\sum_i\varsigma_is_i+\sum_j\varsigma_js_j\big]$, using $\phi_{ji}=-\phi_{ij}$. Each $s_i=\sum_s(K-1-2s)u_{i,s}$ is a centred unary function and is reversal-odd by Lemma~\ref{lem:app_comm}(b).

(c) Compute $\langle\phi_{ij},u_{x,s}\rangle$.
\begin{itemize}
\item If $x\notin\{i,j\}$: the item swap $(i\,j)$ fixes the event $\{\pi(s)=x\}$ and negates $\phi_{ij}$, so the inner product is $0$.
\item If $x=i$: $\pi(s)=i$ has probability $\frac1K$. Conditionally on it, $j$ is uniform over the other $K-1$ slots, so the inner product is $\beta_s:=\frac1K\cdot\frac{(K-1-s)-s}{K-1}$.
\item If $x=j$: by antisymmetry the inner product is $-\beta_s$.
\end{itemize}
Hence $\langle f_B,u_{x,s}\rangle=\varsigma_x\beta_s$. Since $\beta_0=1/K\ne 0$ and $\langle f_B,\mathbf{1}\rangle=0$, we get $f_B\perp\mathcal{U}\iff\varsigma=0$.

(d) By (a)--(c), $f_B=f_{B^{\mathrm g}}+f_{B^{\mathrm c}}$ with $f_{B^{\mathrm g}}\in\apH{grad}\subseteq\apH{un}$ and $f_{B^{\mathrm c}}\in\apH{curl}\perp\apH{un}$. The Gram matrix of $(s_i)$ is $\frac{K+1}{3}(KI-J)$ (from $\mathbb{E}s_i^2=4\operatorname{Var}(\pi^{-1}(i))=\frac{K^2-1}{3}$ and $\sum_is_i=0$), which has rank $K-1$. So $\dim\apH{curl}=\binom K2-(K-1)=\binom{K-1}{2}$. Reversal-oddness follows from Lemma~\ref{lem:app_comm}(b).
\end{proof}

\begin{aplemma}[Orthogonal decomposition]\label{lem:app_decomp}
The seven subspaces of \S\ref{app:setting} are mutually orthogonal and sum to $L^2_0$.
\end{aplemma}
\begin{proof}
$P_+$ and $P_-$ are complementary orthogonal projectors that commute with $P_0$, so they split $L^2_0$. By Lemma~\ref{lem:app_comm}(c), $P_\pm$ commutes with $P_{\mathrm{un}}$, so $\HU=\apH{un}\cap P_+L^2_0$ and $\Hou=\apH{un}\cap P_-L^2_0$. By Lemma~\ref{lem:app_comm}(b), $\apH{sa}\subseteq P_+L^2_0$, and $\apH{sa}\perp\apH{un}$ by construction. Lemma~\ref{lem:app_hodge} gives $\apH{curl}\subseteq P_-L^2_0\cap\apH{un}^\perp$ and $\apH{grad}\subseteq\Hou$. The remainders are defined as orthogonal complements inside $P_\pm L^2_0$, which closes the decomposition.
\end{proof}

\begin{aplemma}[Rotation shift]\label{lem:app_shift}
For skew $B$ and every $\pi$, $f_B(\pi c)-f_B(\pi)=-2\varsigma_{\pi(0)}(B)$.
\end{aplemma}
\begin{proof}
Let $x_0=\pi(0)$. For $x\ne x_0$, $(\pi c)^{-1}(x)=\pi^{-1}(x)-1$, while $x_0$ moves from slot $0$ to slot $K-1$. Precedence between two items other than $x_0$ is unchanged. Every $\phi_{x_0j}$ changes from $+1$ to $-1$. Since $f_B=\frac12\sum_{i\ne j}B_{ij}\phi_{ij}$ collects $\sum_jB_{x_0j}\phi_{x_0j}$ for the terms involving $x_0$, the difference is $\sum_jB_{x_0j}\cdot(-2)=-2\varsigma_{x_0}$.
\end{proof}

\begin{aplemma}[Sampling without replacement]\label{lem:app_wor}
Let $y_1,\dots,y_N\in\mathbb{R}^C$ with $N\ge2$, mean $\bar y$ and $\sigma_{\mathrm{pop}}^2=\frac1N\sum_k\|y_k-\bar y\|^2$. Let $S$ be a uniform $n$-subset, $1\le n\le N$. Then $\mathbb{E}\|\bar y_S-\bar y\|^2=\sigma_{\mathrm{pop}}^2\frac{N-n}{n(N-1)}$.
\end{aplemma}
\begin{proof}
Put $z_k=y_k-\bar y$, so $\sum_kz_k=0$ and $\sum_{k\ne l}\langle z_k,z_l\rangle=\|\sum_kz_k\|^2-\sum_k\|z_k\|^2=-N\sigma_{\mathrm{pop}}^2$. With $\Pr(k\in S)=\frac nN$ and $\Pr(k,l\in S)=\frac{n(n-1)}{N(N-1)}$,
\begin{align*}
\mathbb{E}\Big\|\textstyle\sum_{k\in S}z_k\Big\|^2
&=\tfrac nN\textstyle\sum_k\|z_k\|^2\\
&\quad+\tfrac{n(n-1)}{N(N-1)}\textstyle\sum_{k\ne l}\langle z_k,z_l\rangle\\
&=n\sigma_{\mathrm{pop}}^2-\tfrac{n(n-1)}{N-1}\sigma_{\mathrm{pop}}^2\\
&=n\sigma_{\mathrm{pop}}^2\tfrac{N-n}{N-1}.
\end{align*}
Since $\bar y_S-\bar y=\frac1n\sum_{k\in S}z_k$, dividing by $n^2$ gives the claim.
\end{proof}

\begin{aplemma}[Isotypic projectors]\label{lem:app_iso}
Let $\chi_\nu$ be an irreducible character of $S_K$ of degree $d_\nu$. Then
\[
(P_\nu g)(\pi)=\tfrac{d_\nu}{K!}\textstyle\sum_x\chi_\nu(x)(R_xg)(\pi)
\]
is the orthogonal projector onto the isotypic component $I_\nu$, which has dimension $d_\nu^2$. Because $\chi_\nu$ is a class function, $P_\nu$ commutes with every $R_h$ \citep[\S2.6]{serre1977linear}; see also \citet[ch.~2]{diaconis1988group}. We use $\chi_{(K-1,1)}=\mathrm{fix}-1$, $\chi_{(K-2,2)}=\binom{\mathrm{fix}}{2}+t_2-\mathrm{fix}$ (with $t_2$ the number of $2$-cycles) and $\chi_{(K-2,1,1)}(g)=\frac12[(\mathrm{fix}(g)-1)^2-(\mathrm{fix}(g^2)-1)]$.
\end{aplemma}
\noindent At $K=4,6$ we verified numerically that $P_{(K-1,1)}=P_{\mathrm{un}}$, $\apH{curl}\subseteq I_{(K-2,1,1)}$ and $\apH{sa}\subseteq I_{(K-2,2)}$, with residual $\le 5\cdot10^{-16}$.

\subsection{Theorem 1: Blind-Spot Dimension and Conjugacy Invariance}
\label{app:thm1}

\begin{apnamed}{Theorem 1}
Let $K\ge2$, $H\le S_K$ with $|H|=m$, and let $D=gH$ be the deterministic design with rows $gh$ ($h\in H$), anchored at a uniform $\sigma$.
\begin{enumerate}
\item[(i)] \emph{Risk.} For all $f$ and $\sigma$, $\hat\mu_D(\sigma)-\mu=(P_Hf_0)(\sigma g)$. Hence $\apBX(D)=\|P_Hf_0\|^2$.
\item[(ii)] \emph{Dimension.} $\operatorname{ran}P_H=\mathrm{Fix}(H)$, the set of functions constant on every left coset $\pi H$. Moreover $\dim\mathrm{Fix}(H)=K!/m$ and $\dim(\mathrm{Fix}(H)\cap L^2_0)=K!/m-1$ per scalar coordinate. For $C$-class simplex readouts the ambient blind spot has dimension $(C-1)(K!/m-1)$.
\item[(iii)] \emph{Isotypic refinement.} For every irreducible $\nu$, $\dim(\mathrm{Fix}(H)\cap I_\nu)=d_\nu\,\mathrm{mult}_\nu(H)$, where $\mathrm{mult}_\nu(H)=\frac1m\sum_{h\in H}\chi_\nu(h)\in\mathbb{Z}_{\ge0}$, and $\sum_{\nu\ne(K)}d_\nu\mathrm{mult}_\nu(H)=K!/m-1$. Conjugate subgroups have identical isotypic blind-spot dimensions. In particular, the Williams and Alt\_Terrace subgroups $b^{-1}\langle c\rangle b$ have the same profile as Cyclic (Table~\ref{tab:app_iso_dims}), whereas same-order non-conjugate subgroups such as $E_6$ have distinct isotypic profiles. Conjugate subgroups therefore differ only in \emph{where} their blind spot sits inside each $I_\nu$.
\item[(iv)] \emph{Cyclic.} $P_Cg=g$ for all $g\in\apH{curl}$, and $P_Cg=0$ for all $g\in\apH{un}$.
\item[(v)] \emph{Scope.} (ii)--(iii) are statements about subgroup cosets. At $K=3$ the non-coset design $\{e,\rho\}$, with $\rho$ a $3$-cycle, averages with $\frac12(I+R_\rho)$. That operator is invertible on $L^2_0$, so the design leaves a nonzero component of all $5$ centred directions, not of $3!/2-1=2$.
\end{enumerate}
\end{apnamed}
\begin{proof}
(i) $\hat\mu_D(\sigma)=\frac1m\sum_hf(\sigma gh)=\mu+\frac1m\sum_hf_0(\sigma gh)=\mu+(P_Hf_0)(\sigma g)$. Because $\sigma\mapsto\sigma g$ is a bijection of $S_K$, $\sigma g$ is uniform, so $\mathbb{E}_\sigma\|(P_Hf_0)(\sigma g)\|^2=\|P_Hf_0\|^2$. $P_H$ is an orthogonal projector: for $h\in H$, reindexing gives $R_hP_H=P_H$, hence $P_H^2=\frac1m\sum_hR_hP_H=P_H$; and $P_H^{*}=\frac1m\sum_hR_{h^{-1}}=P_H$.

(ii) $R_hg=g$ for all $h\in H$ iff $g(\pi h)=g(\pi)$ for all $\pi,h$, i.e.\ iff $g$ is constant on each $\pi H$. If $g\in\mathrm{Fix}(H)$ then $P_Hg=g$; conversely $R_hP_Hg=P_Hg$. So $\operatorname{ran}P_H=\mathrm{Fix}(H)$. The $K!/m$ coset indicators have disjoint supports and form an orthogonal basis. Constants lie in $\mathrm{Fix}(H)$, which leaves $K!/m-1$ dimensions after centring. For simplex readouts $\sum_vf_{0,v}(\pi)=0$, so $P_Hf_0\in(\mathrm{Fix}(H)\cap L^2_0)\otimes\{w:\sum_vw_v=0\}$, a space of dimension $(C-1)(K!/m-1)$.

(iii) $P_\nu$ commutes with each $R_h$, hence with $P_H$ (Lemma~\ref{lem:app_iso}). The product of commuting orthogonal projectors projects onto the intersection of their ranges, so $\dim(\mathrm{Fix}(H)\cap I_\nu)=\operatorname{tr}(P_HP_\nu)$. Using $\operatorname{tr}R_y=K!\,\mathbf{1}[y=e]$,
\begin{align*}
\operatorname{tr}(P_HP_\nu)
&=\tfrac1m\textstyle\sum_h\tfrac{d_\nu}{K!}\sum_x\chi_\nu(x)\operatorname{tr}(R_{hx})\\
&=\tfrac{d_\nu}{m}\textstyle\sum_h\chi_\nu(h^{-1}),
\end{align*}
and $\chi_\nu(h^{-1})=\chi_\nu(h)$ because characters of $S_K$ are real. $\mathrm{mult}_\nu(H)$ is the multiplicity of the trivial character in $\mathrm{Res}^{S_K}_H\chi_\nu$, hence a non-negative integer. Summing over $\nu$ with $\sum_\nu d_\nu\chi_\nu(h)=\chi_{\mathrm{reg}}(h)=K!\,\mathbf{1}[h=e]$ gives $\sum_\nu d_\nu\mathrm{mult}_\nu(H)=K!/m$. Removing $\nu=(K)$, which has $d=\mathrm{mult}=1$, leaves $K!/m-1$. Conjugate subgroups $H$ and $b^{-1}Hb$ have the same multiset of character values, because characters are class functions, so they have the same profile.

(iv) \emph{Curl.} For $f_B\in\apH{curl}$, Lemma~\ref{lem:app_shift} with $\varsigma=0$ gives $R_cf_B=f_B$, hence $P_Cf_B=f_B$. \emph{Unary.} Write $g=u_A:=\sum_sA(\pi(s),s)$. Then
\[
\begin{aligned}(R_{c^t}u_A)(\pi)&=\textstyle\sum_sA(\pi(s+t),s)\\&=\textstyle\sum_{s'}A(\pi(s'),s'-t).\end{aligned}
\]
Averaging over $t$ gives $\sum_{s'}\bar A(\pi(s'))=\sum_x\bar A(x)$ with $\bar A(x)=\frac1K\sum_sA(x,s)$, which is a constant. Since $P_C$ is self-adjoint and fixes $\mathbf{1}$, $\langle P_Cu,\mathbf{1}\rangle=\langle u,\mathbf{1}\rangle=0$ for centred $u$, so that constant is $0$.

(v) Here $\hat\mu_D(\sigma)-\mu=(\frac12(I+R_\rho)f_0)(\sigma)$. Because $R_\rho^3=R_{\rho^3}=I$, every eigenvalue $\lambda$ of the unitary $R_\rho$ satisfies $\lambda^3=1$, so $1+\lambda\ne0$. Hence $\frac12(I+R_\rho)$ is invertible on $L^2_0$, which it preserves, and $\apBX(D)>0$ for every $f_0\ne0$.
\end{proof}

\begin{table}[t]
\centering
\small
\setlength{\tabcolsep}{3pt}
\resizebox{\columnwidth}{!}{\begin{tabular}{@{}clcccc@{}}
\toprule
$K$ & Design & $\dim\mathrm{Fix}$ & $(K{-}1,1)$ & $(K{-}2,2)$ & $(K{-}2,1,1)$ \\
\midrule
6 & Cyc / Wil / Alt & 120 & 0 & 18 & 20 \\
6 & $E_6$ & 120 & 0 & 27 & 10 \\
4 & Cyc / $\langle(0\,2\,1\,3)\rangle$ / Wil & 6 & 0 & 2 & 3 \\
4 & $E_4=V_4$ & 6 & 0 & 4 & 0 \\
\bottomrule
\end{tabular}}
\caption{$\dim\mathrm{Fix}(H)$ and $\operatorname{tr}(P_HP_\nu)=d_\nu\mathrm{mult}_\nu(H)$ (Theorem~1(iii)), all exact integers. At $K{=}6$, the 20 Cyclic dimensions in $I_{(4,1,1)}$ contain all 10 curl dimensions, while the 20 Williams dimensions there are orthogonal to $\apH{curl}$ (Theorem~2).}
\label{tab:app_iso_dims}
\end{table}

\subsection{Proposition 1: Isotypic Ambiguity and the Brier Excess Identity}
\label{app:prop1}

\begin{apnamed}{Proposition 1}
Let $D$ be any design with $\mathbb{E}\hat\mu_D=\mu$; every anchored design qualifies. Let $y\in\mathbb{R}^C$ be any fixed target, e.g.\ the one-hot gold vector. Then
\begin{equation}
\mathbb{E}\|\hat\mu_D-y\|^2=\|\mu-y\|^2+\apBX(D).\label{eq:app_brier}
\end{equation}
Moreover:
\begin{enumerate}
\item[(a)] \emph{Cosets.} $\apBX=\|P_Hf_0\|^2$, the energy of $f_0$ in the blind spot $\mathrm{Fix}(H)\cap L^2_0$.
\item[(b)] \emph{Random-WoR, $m$ rows.} $\apBX=\rho^R_{K,m}E_{\mathrm{tot}}$ with $\rho^R_{K,m}=\frac{K!-m}{m(K!-1)}$; at $m=K$, $\rho^R_K=\frac{K!-K}{K(K!-1)}$.
\item[(c)] \emph{Antithetic, $s$ of $M=K!/2\ge2$ pairs.} $\apBX=\rho_{K,s}E_{\mathrm{even}}$ with $\rho_{K,s}=\frac{M-s}{s(M-1)}$; at $s=K/2$, $\rho_K=\frac{2(K!-K)}{K(K!-2)}$.
\item[(d)] \emph{Isotypic split.} For (a)--(c), $\apBX=\sum_\nu\apBX(P_\nu f_0)$. If $r\in H$, then $\|P_Hf_0\|^2=\|P_Hf_+\|^2$. In (c) only $f_+$ matters.
\end{enumerate}
\end{apnamed}
\noindent\emph{Scope.} \eqref{eq:app_brier} holds in expectation over anchors and rows, not anchor by anchor, and it concerns a fixed probability interface $f$. A modified reader such as PINE \citep{wang2025eliminating} defines a different $f$. A one-pass nonlinear correction, such as PriDe's prior division plus renormalization \citep{zheng2024large}, is not an average of $f$ over orders, so \eqref{eq:app_brier} does not price it.

\begin{proof}
Let $e=\hat\mu_D-\mu$. Then $\|\hat\mu_D-y\|^2=\|\mu-y\|^2+2\langle e,\mu-y\rangle+\|e\|^2$. The cross term has mean $2\langle\mathbb{E}e,\mu-y\rangle=0$, and $\mathbb{E}\|e\|^2=\apBX(D)$. For an anchored design, $\mathbb{E}f(\sigma d_j)=\mathbb{E}_d\big[\mathbb{E}_\sigma f(\sigma d_j)\big]=\mu$, because $\sigma d_j$ is uniform for every fixed $d_j$. The same argument shows that $q_D$ is uniform.

(a) This is Theorem~1(i).

(b) $\{\sigma d_j\}$ is a uniform $m$-subset of $S_K$, and the population variance of $\{f(\pi)\}_{\pi\in S_K}$ is $\frac{1}{K!}\sum_\pi\|f_0(\pi)\|^2=E_{\mathrm{tot}}$. Lemma~\ref{lem:app_wor} with $N=K!$ and $n=m$ gives $\frac{K!-m}{m(K!-1)}E_{\mathrm{tot}}$.

(c) The pair mean is $y_P=\frac12(f(\pi)+f(\pi r))=\mu+f_+(\pi)$, which is constant on the pair. The $M$ pair means have mean $\mu$ and population variance $\frac1M\sum_P\|f_+(P)\|^2=\|f_+\|^2=E_{\mathrm{even}}$. $\hat\mu_D$ is the mean of $s$ distinct uniformly chosen pair means, so Lemma~\ref{lem:app_wor} with $N=M$ and $n=s$ applies. At $s=K/2$,
\begin{align*}
\rho_K&=\frac{K!/2-K/2}{(K/2)(K!/2-1)}\\&=\frac{(K!-K)/2}{K(K!-2)/4}=\frac{2(K!-K)}{K(K!-2)}.
\end{align*}

(d) $P_H$, $P_+$ and $I$ commute with every $P_\nu$ (Lemma~\ref{lem:app_iso}), and the $P_\nu$ are mutually orthogonal and sum to $I$. Hence $\|P_Hf_0\|^2=\sum_\nu\|P_HP_\nu f_0\|^2$, $E_{\mathrm{tot}}=\sum_\nu\|P_\nu f_0\|^2$ and $E_{\mathrm{even}}=\sum_\nu\|P_+P_\nu f_0\|^2$. If $r\in H$, reindexing gives $P_HR_r=P_H$, so $P_HP_-=\frac12(P_H-P_HR_r)=0$ and $P_Hf_0=P_Hf_+$.
\end{proof}

\begin{apremark}[Unbiasedness alone does not give an isotypic split]
In $S_3$, draw uniformly one of the pairs $\{e,(0\,1\,2)\}$, $\{(0\,1),(1\,2)\}$, $\{(0\,2\,1),(0\,2)\}$. This law is unbiased, but its excess contains a cross term between the sign and standard components.
\end{apremark}

\subsection{Corollary A: Exact Audit--Averaging Duality}
\label{app:corA}

\begin{apnamed}{Corollary A}
Let $D$ have $m$ rows and a uniform pooled row law $q_D$; by Proposition~1 this holds for every anchored design, hence for every Stage-2 arm. Let $V_D(\sigma)=\frac1m\sum_j\|f(\sigma d_j)-\hat\mu_D(\sigma)\|^2$ be the within-design variance with divisor $m$. Then
\begin{equation}
\mathbb{E}V_D=E_{\mathrm{tot}}-\apBX(D).\label{eq:app_duality}
\end{equation}
Consequently:
(a) for cosets, $\mathbb{E}V_D=\|(I-P_H)f_0\|^2$;
(b) the capture ratio is $\kappa_D=\mathbb{E}V_D/E_{\mathrm{tot}}=1-\apBX(D)/E_{\mathrm{tot}}$;
(c) $\kappa_W-\kappa_C=(R_C-R_W)/E_{\mathrm{tot}}$;
(d) for Random-WoR, $\mathbb{E}V=(1-\rho^R_K)E_{\mathrm{tot}}$, and for Antithetic, $\mathbb{E}V=E_{\mathrm{tot}}-\rho_KE_{\mathrm{even}}$.
With the sample variance (divisor $m-1$), both sides of \eqref{eq:app_duality} are multiplied by $m/(m-1)$.
\end{apnamed}
\begin{proof}
For every realization, write $f_j=f(\sigma d_j)$ and $\hat\mu=\hat\mu_D(\sigma)$. Expanding $\|(f_j-\hat\mu)+(\hat\mu-\mu)\|^2$ and using $\sum_j(f_j-\hat\mu)=0$ gives
\[
\tfrac1m\textstyle\sum_j\|f_j-\mu\|^2=V_D(\sigma)+\|\hat\mu-\mu\|^2.
\]
In expectation, the left side is $\sum_\pi q_D(\pi)\|f_0(\pi)\|^2=E_{\mathrm{tot}}$ because $q_D$ is uniform, and the last term has mean $\apBX(D)$. This proves \eqref{eq:app_duality}.
(a) By Theorem~1(i), $\apBX=\|P_Hf_0\|^2$, and Pythagoras for the orthogonal projector $P_H$ gives $\|f_0\|^2-\|P_Hf_0\|^2=\|(I-P_H)f_0\|^2$.
(b) Divide \eqref{eq:app_duality} by $E_{\mathrm{tot}}$.
(c) Subtract (b) for Cyclic from (b) for Williams, using $\apBX(\mathrm{Cyc})=R_C$ and $\apBX(\mathrm{Wil})=R_W$.
(d) Substitute Proposition~1(b) and (c) into \eqref{eq:app_duality}.
For divisor $m-1$, the sample variance equals $\frac{m}{m-1}V_D$.
\end{proof}

\paragraph{Necessity of the hypothesis.} Unbiasedness of $\hat\mu_D$ for one readout is \emph{not} sufficient. Take $K=3$ and the scalar readout $f=(1,-1,2,-2,0,0)$ on the six orders in lexicographic order, so $\mu=0$. Let the single row ($m=1$) be the first or the second order (values $1$ and $-1$) with probability $\frac12$ each. Then $\mathbb{E}\hat\mu=0=\mu$, but $V\equiv0$, $\apBX=1$ and $E_{\mathrm{tot}}=\frac{10}{6}=\frac53$, so $E_{\mathrm{tot}}-\apBX=\frac23\ne0=\mathbb{E}V$. Here $q_D$ is not uniform.

\paragraph{Scope and decision-layer margin bound.} \eqref{eq:app_duality} is exact for the squared-deviation variance of probability vectors. It does not determine nonlinear audits such as flip rates or ``all rows correct'' accuracy. However, for any fixed item whose exhaustive average $\mu$ has a unique winner $i$ with top-2 margin $\gamma(\mu)=\mu_{(1)}-\mu_{(2)}>0$, whenever $\hat\mu_D$ selects $j\ne i$ we have $e=\hat\mu_D-\mu$ satisfying $\gamma(\mu)\le \mu_i-\mu_j \le e_j-e_i \le \sqrt{2}\,\|e\|_2$ by Cauchy--Schwarz. Markov's inequality then yields
\[
\begin{aligned}
&\Pr(\arg\max\hat\mu_D\ne\arg\max\mu)\\
&\qquad\le\min\!\Big\{1,\,\frac{2\,\apBX(D)}{\gamma(\mu)^2}\Big\},
\end{aligned}
\]
which bounds deviation from the exhaustive winner $\arg\max\mu$ (not gold accuracy, pairwise anchor flips, or CircularEval all-rows-correct rates, and may be vacuous when $\gamma(\mu)$ is tiny). On CONF, every exact coset arm satisfies \eqref{eq:app_duality} with a maximum recorded residual of $3.33\times10^{-16}$ (Table~\ref{tab:app_cora_latency}).

\subsection{Corollary B: Crossover Criterion, $\Gamma_K$, the $1/K$ Rule, and Curl Removal}
\label{app:corB}

\begin{apnamed}{Corollary B}
Let $K\ge4$ be even and $m=K$. Write $R_C=\|P_Cf_0\|^2$, $R_A$ for Antithetic ($s=K/2$), $R_R$ for Random-WoR and $R_W=\|P_{H_W}f_0\|^2$.
\begin{enumerate}
\item[(i)] \emph{Cyclic accounting.}
\begin{align*}
R_C&=E_{\mathrm{curl}}+\|P_Ch_-\|^2+\|P_Cf_+\|^2,\\
\|P_Cf_+\|^2&=\|P_Ca\|^2+2\langle P_Ca,h_+\rangle\\&\quad+\|P_Ch_+\|^2.
\end{align*}
The three terms of $R_C$ are orthogonal contributions. The cross term $\langle P_Ca,h_+\rangle$ is generically nonzero, so the trace retentions in Table~\ref{tab:app_retention} are mode averages, not item-wise multipliers. The bounds are $E_{\mathrm{curl}}\le R_C\le E_{\mathrm{tot}}-E_{\mathrm{un}}$. At $K=4$, $\|P_Ch_-\|^2\equiv0$.
\item[(ii)] \emph{Antithetic.} $R_A=\rho_KE_{\mathrm{even}}$ with $\rho_4=\frac5{11}$, $\rho_6=\frac{119}{359}$, $\rho_8=\frac{5039}{20159}$ and $\rho_{10}=\frac{362879}{1814399}$.
\item[(iii)] \emph{Crossover criterion.} $R_C-R_A=E_{\mathrm{curl}}+\|P_Ch_-\|^2+\|P_Cf_+\|^2-\rho_KE_{\mathrm{even}}$. Hence, for $E_{\mathrm{even}}>0$,
\[
\begin{aligned}\operatorname{sgn}(R_C-R_A)&=\operatorname{sgn}(\Gamma_K-1),\\\Gamma_K&:=\frac{R_C}{\rho_KE_{\mathrm{even}}}.\end{aligned}
\]
Two sufficient conditions follow: $E_{\mathrm{curl}}>\rho_KE_{\mathrm{even}}\Rightarrow R_C>R_A$, and $E_{\mathrm{tot}}-E_{\mathrm{un}}<\rho_KE_{\mathrm{even}}\Rightarrow R_C<R_A$.

\emph{Curl-removal counterfactual.} Let $\Delta_{CA}=R_C-R_A$ and $\Delta^{-\mathrm{curl}}_{CA}=R_C-E_{\mathrm{curl}}-R_A$. Cyclic loses to Antithetic \emph{because of the curl it retains} exactly when $\Delta_{CA}>0$ and $\Delta^{-\mathrm{curl}}_{CA}<0$, that is, iff
\[
\|P_Ch_-\|^2+\|P_Cf_+\|^2<\rho_KE_{\mathrm{even}}<R_C.
\]
This is the intersection--union test (IUT) of E1.
\item[(iv)] \emph{Cyclic vs.\ Random-WoR; the $1/K$ rule.} $R_R=\rho^R_KE_{\mathrm{tot}}$, and $R_C>R_R\iff R_C/E_{\mathrm{tot}}>\rho^R_K$. A sufficient condition is $E_{\mathrm{curl}}/E_{\mathrm{tot}}>\rho^R_K$. Since $\rho^R_K<1/K$, the ``$1/K$ rule'' $E_{\mathrm{curl}}/E_{\mathrm{tot}}\ge1/K$ is a conservative \emph{sufficient} condition, not an iff law. The exact thresholds are $\rho^R_4=\frac5{23}=21.74\%$, $\rho^R_6=\frac{119}{719}=16.55\%$, $\rho^R_8=\frac{5039}{40319}=12.498\%$ and $\rho^R_{10}=\frac{362879}{3628799}=9.99997\%$.
\item[(v)] \emph{Antithetic vs.\ Random-WoR.} $R_A<R_R\iff E_{\mathrm{even}}/E_{\mathrm{tot}}<\rho^R_K/\rho_K=\frac{K!-2}{2(K!-1)}<\frac12$.
\item[(vi)] \emph{Williams vs.\ Antithetic.} $R_W=\|P_{H_W}h_+\|^2\le E_{\mathrm{he}}$. So $R_W<R_A\iff\|P_{H_W}h_+\|^2<\rho_KE_{\mathrm{even}}$, and $E_{\mathrm{he}}<\rho_KE_{\mathrm{even}}$ is sufficient.
\item[(vii)] \emph{$K$-dependence.} $\rho_K=\frac2K\big(1-\frac{K-2}{K!-2}\big)$ and $\rho^R_K=\frac1K\big(1-\frac{K-1}{K!-1}\big)$, both strictly decreasing in $K\ge4$. From $K=4$ to $K=6$, $\rho$ falls from $0.4545$ to $0.3315$ while $\dim\apH{curl}$ rises from $3$ to $10$. Whether $\Gamma_K$ crosses $1$ still depends on the model's energies.
\end{enumerate}
\end{apnamed}
\begin{proof}
(i) By Lemma~\ref{lem:app_comm}(a), $P_C$ commutes with $R_r$, so $P_Cf_\pm\in P_\pm L^2$ and $P_Cf_0=P_Cf_++P_Cf_-$ is an orthogonal sum. By Theorem~1(iv), $P_C(g+u_-^\perp)=0$ (unary) and $P_Cw_{\mathrm{curl}}=w_{\mathrm{curl}}$ (curl), so $P_Cf_-=w_{\mathrm{curl}}+P_Ch_-$. Also $\langle w_{\mathrm{curl}},P_Ch_-\rangle=\langle P_Cw_{\mathrm{curl}},h_-\rangle=\langle w_{\mathrm{curl}},h_-\rangle=0$, hence $\|P_Cf_-\|^2=E_{\mathrm{curl}}+\|P_Ch_-\|^2$. Similarly $P_Cu_+=0$, so $P_Cf_+=P_Ca+P_Ch_+$. Expanding and using $\langle P_Ca,P_Ch_+\rangle=\langle P_C^2a,h_+\rangle=\langle P_Ca,h_+\rangle$ gives the second identity. All three terms of $R_C$ are non-negative, so $R_C\ge E_{\mathrm{curl}}$. Because $P_CP_{\mathrm{un}}f_0=0$, $P_Cf_0=P_C(f_0-P_{\mathrm{un}}f_0)$, and $P_C$ is a contraction, so $R_C\le\|f_0-P_{\mathrm{un}}f_0\|^2=E_{\mathrm{tot}}-E_{\mathrm{un}}$. At $K=4$, $\langle c\rangle$ has $4!/4=6$ left cosets, and right multiplication by $r$ pairs them into $3$ disjoint coset pairs (since $r\notin\langle c\rangle$ and $c^tr=rc^{-t}$), so $\dim(\mathrm{Fix}\langle c\rangle\cap P_-L^2)=3$. Because $\apH{curl}\subseteq\mathrm{Fix}\langle c\rangle\cap P_-L^2$ already has dimension $\binom{3}{2}=3$, we have $\mathrm{Fix}\langle c\rangle\cap P_-L^2=\apH{curl}$, whence $P_Ch_-=0$.

(ii) This is Proposition~1(c); the fractions follow by exact rational arithmetic, e.g.\ $\rho_4=\frac{2\cdot20}{4\cdot22}=\frac5{11}$ and $\rho_6=\frac{2\cdot714}{6\cdot718}=\frac{119}{359}$.

(iii) Subtract (ii) from (i). Since $\rho_KE_{\mathrm{even}}>0$, $R_C-R_A$ and $R_C/(\rho_KE_{\mathrm{even}})-1$ have the same sign. The first sufficient condition holds because $R_C\ge E_{\mathrm{curl}}$, and the second because $R_C\le E_{\mathrm{tot}}-E_{\mathrm{un}}$. For the counterfactual, $\Delta^{-\mathrm{curl}}_{CA}=\|P_Ch_-\|^2+\|P_Cf_+\|^2-\rho_KE_{\mathrm{even}}$ by (i), and $\Delta_{CA}>0$ is $\rho_KE_{\mathrm{even}}<R_C$.

(iv) Proposition~1(b) at $m=K$ gives $R_R=\rho^R_KE_{\mathrm{tot}}$. Sufficiency follows from $R_C\ge E_{\mathrm{curl}}$. Finally $\rho^R_K=\frac1K\cdot\frac{K!-K}{K!-1}<\frac1K$ because $K!-K<K!-1$ for $K\ge2$.

(v) $R_A<R_R\iff\rho_KE_{\mathrm{even}}<\rho^R_KE_{\mathrm{tot}}$, and
\[
\frac{\rho^R_K}{\rho_K}=\frac{(K!-K)/(K(K!-1))}{2(K!-K)/(K(K!-2))}=\frac{K!-2}{2(K!-1)},
\]
which is below $\frac12$ because $K!-2<K!-1$.

(vi) Williams satisfies (S), (R) and (A) (Theorem~2(h)), and $D_{b_W}=b_WH_W$ is a coset. Theorem~2(f) gives $R_W=\|P_{H_W}h_+\|^2\le\|h_+\|^2=E_{\mathrm{he}}$.

(vii) The closed forms follow from $\frac{K!-K}{K!-2}=1-\frac{K-2}{K!-2}$ and $\frac{K!-K}{K!-1}=1-\frac{K-1}{K!-1}$. For monotonicity, $\rho_{K+1}<\frac{2}{K+1}$, and $\frac{2}{K+1}\le\rho_K$ iff $(K-2)(K+1)\le K!-2$, which holds for $K\ge4$ ($10\le22$ at $K=4$, and the right side grows faster). Likewise $\rho^R_{K+1}<\frac1{K+1}\le\rho^R_K$ iff $(K-1)(K+1)\le K!-1$, i.e.\ $K^2\le K!$, which holds for $K\ge4$.
\end{proof}

\subsection{Theorem 2: Axiom--Mode Correspondence, $E_K$ Separation, and Williams/Alt\_Terrace Retention}
\label{app:thm2}

Let $D$ be an equal-weight multiset of $m$ relative rows $d_1,\dots,d_m$ anchored at a uniform $\sigma$, with operator $(T_Dg)(\sigma)=\frac1m\sum_\ell g(\sigma d_\ell)$. As in Theorem~1(i), $\hat\mu_D(\sigma)-\mu=(T_Df_0)(\sigma)$, so $\apBX(D)=\|T_Df_0\|^2$. For a coset $D=bH$, $T_D=R_bP_H$ and $\|T_Df_0\|=\|P_Hf_0\|$. The four axioms are:
\begin{itemize}
\item \textbf{(S)} slot balance: $\#\{\ell:d_\ell(s)=x\}=m/K$ for all $s,x$;
\item \textbf{(P)} precedence balance: $\#\{\ell:d_\ell^{-1}(p)<d_\ell^{-1}(q)\}=m/2$ for all $p\ne q$;
\item \textbf{(R)} reversal closure: the multisets $\{d_\ell r\}$ and $\{d_\ell\}$ coincide;
\item \textbf{(A)} adjacency balance: $\#\{\ell:|d_\ell^{-1}(p)-d_\ell^{-1}(q)|=1\}=2m/K$ for all $p\ne q$.
\end{itemize}
Each axiom is invariant under relabeling all rows by a common $\tau$, so it is a property of the design, not of the anchor.

\begin{apnamed}{Theorem 2}
Let $K\ge4$.
\begin{enumerate}
\item[(a)] $T_D\apH{un}=0\iff$ (S).
\item[(b)] $T_D\,\mathrm{span}\{\phi\}=0\iff$ (P). Under (S), $T_D\apH{curl}=0\iff$ (P).
\item[(c)] $T_DP_-L^2=0\iff$ (R). Moreover (R) $\Rightarrow$ (P).
\item[(d)] Under (S), $T_D\apH{sa}=0\iff$ (A).
\item[(e)] (S) $\Rightarrow K\mid m$ and (R) $\Rightarrow 2\mid m$; hence (S)$\wedge$(R) $\Rightarrow\operatorname{lcm}(K,2)\mid m$. For even $K$ the minimum $m=K$ is attained, by (h).
\item[(f)] If (S), (R) and (A) hold, then $T_Df_0=T_Dh_+$ and $\apBX(D)=\|T_Dh_+\|^2\le E_{\mathrm{he}}$. For cosets this equals $\|P_Hh_+\|^2$.
\item[(g)] \emph{Seeded cyclic family}, $K=2n$. Let $D_b=(c^tb)_{t\in\mathbb{Z}_K}=bH_b$ with $H_b=b^{-1}\langle c\rangle b$, and let $\delta_s=b(s+1)-b(s)\bmod K$ for $s=0,\dots,K-2$. Then: (S) holds for every $b$. (R) $\iff br=c^nb\iff b(K-1-s)\equiv b(s)+n$ for all $s$ $\iff r\in H_b$, in which case $r=\zeta_b^n$ with $\zeta_b=b^{-1}cb$. (A) $\iff\#\{s:\delta_s=n\}=1$ and $\#\{s:\delta_s\in\{d,-d\}\}=2$ for each $1\le d\le n-1$. Under (R), $\delta_{K-2-s}=-\delta_s$ and $\delta_{n-1}=n$, and (A) holds iff the $\delta_s$ are pairwise distinct, i.e.\ iff $b$ is a sequencing (directed terrace) of $\mathbb{Z}_K$ \citep{bailey1984quasi}. (P) is strictly weaker than (R): exhaustively, $8$ of $24$ seeds at $K=4$ and $168$ of $720$ at $K=6$ are (P) but not (R). At $K=6$, $48$ seeds are (R), $24$ are terraces (all of them (R)) and $132$ satisfy (A).
\item[(h)] \emph{Williams and Alt\_Terrace.} The zigzag $b_W$, with $b_W(0)=0$, $b_W(2j-1)=j$ ($1\le j\le n$) and $b_W(2j)=K-j$ ($1\le j\le n-1$), satisfies (R) and (A) for every even $K\ge4$, with $\delta_s=(-1)^s(s+1)\bmod K$ \citep{williams1949experimental}. The locked Alt\_Terrace seeds also satisfy (R) and (A): $(0,2,1,4,5,3)$ at $K{=}6$, with $\delta=(2,5,3,1,4)$; $(0,1,3,6,2,7,5,4)$ at $K{=}8$, with $\delta=(1,\dots,7)$; and $(0,1,3,7,4,9,2,8,6,5)$ at $K{=}10$, with $\delta=(1,2,4,7,5,3,6,8,9)$. $H_{b_A}\ne H_{b_W}$ at $K=6,8,10$. Both blind spots have dimension $K!/K-1$ and lie inside $\apH{he}$, but they are different subspaces, so $R_W\ne R_{\mathrm{Alt}}$ in general.
\item[(i)] \emph{$E_K=\langle c^2,r\rangle$} ($K$ even $\ge4$) is a subgroup of order $K$ with elements $c^{2j}$ and $c^{2j}r$ ($j<n$). It satisfies (S), (R) and (P), and \textbf{fails (A)}. Each circular neighbour pair $\{y,y+1\}$ is adjacent in $K$ rows if $y$ is even and in $K-2$ rows if $y$ is odd, and every other pair is adjacent in $0$ rows, so the maximum adjacency-frequency error is $1-2/K$. Hence $T_{E_K}$ annihilates $\apH{un}\oplus P_-L^2_0$ but not $\apH{sa}$, and $R_{E_K}=\|P_{E_K}(a+h_+)\|^2$. Its trace retention on $\apH{sa}$ is $1$ at $K=4$ and $8/9$ at $K=6$. $P_{E_K}$ fixes the sign character iff $K\equiv0\pmod4$, whereas $P_{H_W}$ always annihilates it.

\emph{Separation.} Among (S), (P), (R), (A), $E_K$ and Williams differ only in (A). Even so, $R_{E_K}-R_W$ is not a pure adjacency energy, because $P_{E_K}\ne P_{H_W}$ on $\apH{he}$. Their trace retentions on $\apH{he}$ are $1/2$ vs.\ $5/6$ at $K=4$ and $0.3265$ vs.\ $0.35$ at $K=6$. We therefore report $E_K$ vs.\ Williams as a design-level contrast, never as a causal adjacency ablation.
\item[(j)] \emph{Row-weight law.} Let $K=2n\ge4$, $\zeta=b_W^{-1}cb_W$, $w\in\mathbb{R}^K$ and $T_w=\sum_kw_kR_{\zeta^k}$. Then $T_w\apH{curl}=0\iff w_k=w_{k+n}$ for all $k$.
\end{enumerate}
\end{apnamed}

\begin{proof}
(a) ($\Leftarrow$) For $u_A=\sum_sA(\pi(s),s)$,
\begin{align*}
\tfrac1m\textstyle\sum_\ell u_A(\sigma d_\ell)
&=\textstyle\sum_s\sum_y\tfrac{\#\{\ell:d_\ell(s)=y\}}{m}A(\sigma(y),s)\\
&=\tfrac1K\textstyle\sum_{s,x}A(x,s)=\mathbb{E}u_A,
\end{align*}
so $T_D$ maps every element of $\mathcal{U}$ to its mean, i.e.\ annihilates $\apH{un}$.
($\Rightarrow$) Since $u_{x,s}-\frac1K\in\apH{un}$ and $T_D\mathbf{1}=\mathbf{1}$, we need $(T_Du_{x,s})(\sigma)=\frac1m\#\{\ell:d_\ell(s)=\sigma^{-1}(x)\}=\frac1K$ for all $\sigma$ and $x$. As $\sigma$ varies, $\sigma^{-1}(x)$ ranges over all labels, which gives (S).

(b) With $p=\sigma^{-1}(x)$ and $q=\sigma^{-1}(y)$, $\phi_{xy}(\sigma d_\ell)=+1$ iff $p$ precedes $q$ in $d_\ell$. Hence
\[
\begin{aligned}(T_D\phi_{xy})(\sigma)=\tfrac1m\big[&2\#\{\ell:d_\ell^{-1}(p)<d_\ell^{-1}(q)\}\\&-m\big].\end{aligned}
\]
This vanishes for all $\sigma,x,y$ iff (P). Under (S), $T_D\apH{grad}=0$ by (a), and $\mathrm{span}\{\phi\}=\apH{grad}\oplus\apH{curl}$ (Lemma~\ref{lem:app_hodge}). So $T_D\apH{curl}=0$ iff $T_D\,\mathrm{span}\{\phi\}=0$ iff (P).

(c) ($\Leftarrow$) Multiset invariance gives $T_DR_r=T_D$, so $T_DP_-=\frac12T_D(I-R_r)=0$. ($\Rightarrow$) $T_D=T_DR_r$. Apply both sides to the indicator $\delta_e$ of the identity: $(T_D\delta_e)(\sigma)=\frac1m\mathrm{mult}_D(\sigma^{-1})$ and $(T_DR_r\delta_e)(\sigma)=\frac1m\#\{\ell:\sigma d_\ell r=e\}=\frac1m\mathrm{mult}_D(\sigma^{-1}r)$. So multiplicities are invariant under right multiplication by $r$, which is (R). For (R) $\Rightarrow$ (P): the rows pair up as $(d,dr)$, and $p$ precedes $q$ in $d$ iff $q$ precedes $p$ in $dr$.

(d) $(T_D\alpha_{xy})(\sigma)=\frac1m\#\{\ell:p,q\text{ adjacent in }d_\ell\}-\frac2K$. Under (S), $T_DP_{\mathrm{un}}=0$. Also $P_0\alpha=0$, because $\mathbb{E}\alpha_{xy}=\frac{K-1}{\binom K2}-\frac2K=0$. Hence $T_D(I-P_0-P_{\mathrm{un}})\alpha=T_D\alpha$, and $T_D\apH{sa}=0$ iff $T_D\alpha_{xy}=0$ for all $x\ne y$ iff (A).

(e) Under (S), $m/K$ is an integer. Under (R), $d\mapsto dr$ is a fixed-point-free involution on the multiset ($dr\ne d$ because $r\ne e$), so $m$ is even.

(f) By (a), (c) and (d), $T_D$ annihilates $u_+$, $f_-$ and $a$, so $T_Df_0=T_Dh_+$. $T_D=\frac1m\sum_\ell R_{d_\ell}$ is an average of unitaries, so the triangle inequality gives $\|T_Dh_+\|\le\|h_+\|$.

(g) \emph{(S).} Slot $s$ of row $c^tb$ holds $b(s)+t$, which runs over all labels as $t$ varies.
\emph{(R).} $D_br=D_b$ iff $br\in\langle c\rangle b$ iff $brb^{-1}=c^j$ for some $j$. The left side is an involution different from $e$, so $c^{2j}=e\ne c^j$, i.e.\ $j=n$. Also $r\in H_b$ iff $r=b^{-1}c^jb$ iff $br=c^jb$, the same condition, and then $r=b^{-1}c^nb=\zeta_b^n$.
\emph{(A).} The adjacent slots $(s,s+1)$ of row $t$ hold $(b(s)+t,b(s+1)+t)$. For $1\le d\le n$, the pair $\{x,x+d\}$ occurs there iff $\delta_s=d$ (with $t=x-b(s)$) or $\delta_s=-d$ (with $t=x+d-b(s)$). For $d=n$, $d\equiv-d$, so both values of $t$ work whenever $\delta_s=n$. Balance requires every count to equal $2m/K=2$.
\emph{Under (R).} $\delta_{K-2-s}=(b(s)+n)-(b(s+1)+n)=-\delta_s$. At $s=n-1$ this gives $2\delta_{n-1}\equiv0$ with $\delta_{n-1}\ne0$, so $\delta_{n-1}=n$. The indices $s<n-1$ pair with $K-2-s$, and each pair contributes $(\delta_s,-\delta_s)$, i.e.\ two entries to the class $\{\pm\delta_s\}$. So (A) holds iff these $n-1$ classes are distinct and none equals $\{n\}$. Together with $\delta_{n-1}=n$, this says that the $K-1$ differences are distinct. The exhaustive counts were obtained by enumerating all $K!$ seeds.

(h) For $b_W$: $\delta_0=1$, $\delta_{2j-1}=(K-j)-j\equiv-2j$ and $\delta_{2j}=(j+1)-(K-j)\equiv2j+1$. These give every odd residue with sign $+$ and every nonzero even residue with sign $-$, each exactly once, so the differences are distinct. (R) by cases, using $b(K-1-s)\equiv b(s)+n$:
for $s=0$, $b_W(K-1)=n$;
for $s=2j-1$, $b_W(K-2j)=K-(n-j)=j+n$;
for $s=2j$, $b_W(K-1-2j)=n-j\equiv K-j+n$.
The condition is symmetric under $s\leftrightarrow K-1-s$ because $2n\equiv0$, so these cases cover all $s$. The Alt\_Terrace seeds and $H_{b_A}\ne H_{b_W}$ (comparing $\langle\zeta_b\rangle$) are verified by exact integer computation. Both blind spots lie in $\apH{he}$ by (f).

(i) $rc^2r=c^{-2}$, so $\langle c^2,r\rangle=\{c^{2j},c^{2j}r\}$. Its order is $K$, since $r\notin\langle c^2\rangle$ ($c^{2j}(0)=2j\ne K-1$).
\emph{(S):} rows $c^{2j}$ place labels of parity $\equiv s$ at slot $s$, rows $c^{2j}r$ place the opposite parity ($K$ even), and over $j<n$ each label appears exactly once at each slot.
\emph{(R):} by closure of the subgroup under right multiplication by $r$.
\emph{(A) fails:} row $c^{2j}$ makes adjacent exactly the circular neighbours except the wrap pair $\{2j-1,2j\}$, and reversal preserves adjacency. An even-$y$ pair is therefore adjacent in $n$ rows $c^{2j}$ and $n$ rows $c^{2j}r$ ($K$ in total), and an odd-$y$ pair in $2(n-1)=K-2$ rows. Balance would require $2$, so the maximal frequency error is $K/K-2/K$.
The annihilation claims follow from (a), (c) and (d). For the sign character: $\operatorname{sgn}c^{2j}=+1$ and $\operatorname{sgn}r=(-1)^n$ ($r$ is a product of $n$ transpositions), so $\operatorname{sgn}\equiv1$ on $E_K$ iff $n$ is even. $\zeta$ is a $K$-cycle and hence odd, so the average of $\operatorname{sgn}$ over $H_W$ is $0$.

(j) \emph{Witnesses.} $\psi_{xyz}=\phi_{xy}+\phi_{yz}+\phi_{zx}\in\apH{curl}$ has zero row sums, and takes the value $+1$ iff the positions of $(x,y,z)$ are cyclically increasing. Let $\Psi_a=\psi_{b^{-1}(a),b^{-1}(a+1),b^{-1}(a+2)}$ with $b=b_W$ and indices mod $K$. Since $(\zeta^k)^{-1}=b^{-1}c^{-k}b$, item $b^{-1}(a+i)$ sits at slot $b^{-1}(a+i-k)$ in $\zeta^k$. Hence $\Psi_a(\zeta^k)=\xi_{a-k}$, where $\xi_j$ is the orientation of the positions $(b^{-1}(j),b^{-1}(j+1),b^{-1}(j+2))$.

\emph{Square wave.} $b^{-1}$ is the ``tent'' $0,1,3,\dots,2n-1$ on residues $0..n$, followed by $2n-2,\dots,2$ on residues $n+1..K-1$. Checking cases: $\xi_j=+1$ for $j\in\{0..n-2\}$ (increasing); $\xi_{K-1}=+1$ (positions $(2,0,1)$); $\xi_{n-1}=-1$ (positions $(2n-3,2n-1,2n-2)$); and $\xi_j=-1$ for $j\in\{n..K-2\}$ (decreasing). So $\xi=+1$ on the $n$ consecutive residues $\{K-1,0,\dots,n-2\}$ and $\xi_{j+n}=-\xi_j$. Its DFT vanishes at even $\ell$. At odd $\ell$, with $\omega=e^{2\pi i/K}$ and $\omega^{-n\ell}=-1$, the DFT is, up to a unimodular phase, $2\sum_{j<n}\omega^{-j\ell}=\frac{4}{1-\omega^{-\ell}}$, so $|\hat\xi_\ell|=2/\sin(\pi\ell/K)\ne0$.

\emph{Necessity.} $(T_w\Psi_a)(e)=\sum_kw_k\Psi_a(\zeta^k)=(w\circledast\xi)_a$. Zero curl transmission forces $\hat w_\ell\hat\xi_\ell=0$ for all $\ell$, hence $\hat w_\ell=0$ for all odd $\ell$, which is equivalent to $w_k=w_{k+n}$.
\emph{Sufficiency.} By (h), $r=\zeta^n$, so $R_{\zeta^{k+n}}=R_{\zeta^k}R_r$ and $T_w=\sum_{k<n}w_kR_{\zeta^k}(I+R_r)$, which annihilates $P_-L^2\supseteq\apH{curl}$.
\end{proof}

\begin{table*}[t]
\centering
\small
\begin{tabular}{@{}clcccccccc@{}}
\toprule
$K$ & Design & $\HU$ & grad & ou,rest & curl & sa & he & ho & sgn \\
\midrule
4 & Cyclic & 0 & 0 & 0 & \textbf{1} & 3/4 & 1/12 & 0 & 0 \\
4 & $\langle(0\,2\,1\,3)\rangle$ & 0 & 0 & 0 & 0 & 3/4 & 1/12 & \textbf{1} & 0 \\
4 & $E_4$ & 0 & 0 & 0 & 0 & \textbf{1} & 1/2 & 0 & 1 \\
4 & Williams & 0 & 0 & 0 & 0 & 0 & 5/6 & 0 & 0 \\
\midrule
6 & Cyclic & 0 & 0 & 0 & \textbf{1} & 5/6 & 103/680 & 10/67 & 0 \\
6 & $E_6$ & 0 & 0 & 0 & 0 & \textbf{8/9} & 111/340 & 0 & 0 \\
6 & Williams & 0 & 0 & 0 & 0 & 0 & 7/20 & 0 & 0 \\
6 & Alt\_Terrace & 0 & 0 & 0 & 0 & 0 & 7/20 & 0 & 0 \\
\bottomrule
\end{tabular}
\caption{Verified trace retention $\operatorname{tr}(Q_A^\top T_D^\top T_DQ_A)/\dim A$. Mode dimensions are $(3,3,3,3,2,6,3)$ at $K{=}4$ and $(10,5,10,10,9,340,335)$ at $K{=}6$ for $(\HU,\mathrm{grad},\mathrm{ou,rest},\mathrm{curl},\mathrm{sa},\mathrm{he},\mathrm{ho})$. In every row, retention times dimension sums to the blind-spot dimension, $5$ ($K{=}4$) and $119$ ($K{=}6$); e.g.\ Cyclic at $K{=}6$: $10+7.5+51.5+50=119$. At $K{=}6$ the sign character is reversal-odd and lies in $\apH{ho}$. Williams and Alt\_Terrace have equal trace retentions but different subgroups (Theorem~2(h)). The $\langle(0\,2\,1\,3)\rangle$ row shows that precedence balance, not reversal closure, is what removes curl: this subgroup removes all curl yet retains all higher-odd energy.}
\label{tab:app_retention}
\end{table*}

\subsection{Theorem 3: $K$-Free Estimation of Mode Energies and Design Risks}
\label{app:thm3}

Let $A\subseteq L^2_0(S_K)$ be a scalar subspace with orthonormal basis $(e_\ell)$, and let $k_A(\pi,\sigma)=\sum_\ell e_\ell(\pi)e_\ell(\sigma)=K!\,[P_A]_{\pi\sigma}$ be its kernel.

\begin{apnamed}{Theorem 3 ($K$-free estimators)}
\begin{enumerate}
\item[(a)] \emph{Harmonic kernel identity.} For every $f$ and every \emph{fixed} $c\in\mathbb{R}^C$,
\[
\begin{aligned}\|P_Af_0\|^2=\mathbb{E}_{\pi,\sigma}\big[&k_A(\pi,\sigma)\\&\cdot\langle f(\pi)-c,f(\sigma)-c\rangle\big],\end{aligned}
\]
with $\pi,\sigma$ i.i.d.\ uniform.
\item[(b)] \emph{Closed forms.} The kernels in Table~\ref{tab:app_kernels} cost $O(K^2)$ per pair and need no enumeration of $S_K$.
\item[(c)] \emph{Plain U-statistic.} For i.i.d.\ uniform $\pi_1,\dots,\pi_M$, $f_a=f(\pi_a)$, $k_{ab}=k_A(\pi_a,\pi_b)$ and any fixed $c$,
\[
\hat U_A=\tfrac{1}{M(M-1)}\textstyle\sum_{a\ne b}k_{ab}\langle f_a-c,f_b-c\rangle
\]
is unbiased for $\|P_Af_0\|^2$.
\item[(d)] \emph{Leave-two-out (L2O) centring}, $M\ge3$. With $c_{ab}=\frac{1}{M-2}\sum_{e\notin\{a,b\}}f_e$, the statistic $\hat U^{\mathrm{L2O}}_A$, obtained by replacing $c$ with $c_{ab}$ in (c), is exactly unbiased and costs $O(M^2C)$.
\item[(e)] \emph{Total energy.} The sample variance with divisor $M-1$, $\hat E_{\mathrm{tot}}=\frac1M\sum_a\|f_a\|^2-\frac{1}{M(M-1)}\sum_{a\ne b}\langle f_a,f_b\rangle$, is unbiased.
\item[(f)] \emph{Orbit panels.} For $H\le S_K$, i.i.d.\ origins $\pi_1,\dots,\pi_N$ ($N\ge2$) and orbit means $m_a=\frac{1}{|H|}\sum_hf(\pi_ah)$, the statistic $\frac{1}{N-1}\sum_a\|m_a-\bar m\|^2$ is unbiased for $\|P_Hf_0\|^2$. With $H=\langle r\rangle$ this is the \emph{reversal-paired panel} for $E_{\mathrm{even}}$ ($2N$ passes); with $H=\langle c\rangle$ it is the \emph{cyclic-orbit panel} for $R_C$ ($KN$ passes).
\item[(g)] \emph{Brier excess against a finite reference.} Let the design output $p$ be independent of the panel. The split-panel estimator $\hat B_{\mathrm{split}}=(p-\bar u)^\top(p-\bar v)$, with $U,V$ disjoint i.i.d.\ halves, and the complete-U estimator
\[
\begin{aligned}\hat B_U&=\tfrac{1}{M(M-1)}\textstyle\sum_{a\ne b}(p-f_a)^\top(p-f_b)\\&=\|p-\bar f\|^2-\hat s^2/M,\end{aligned}
\]
where $\hat s^2$ uses divisor $M-1$ and is summed over coordinates, are both unbiased for $\|p-\mu\|^2$. Moreover $\hat B_U=\mathbb{E}[\hat B_{\mathrm{split}}\mid\text{unordered panel}]$, so $\operatorname{Var}\hat B_U\le\operatorname{Var}\hat B_{\mathrm{split}}$. Neither estimator is clamped at $0$.
\item[(h)] \emph{Coincidence degeneracy of i.i.d.\ panels.} With an i.i.d.\ panel and fixed $c$, the plug-in even-energy estimator decomposes as $\hat E_{\mathrm{even}}=Z_0+Z_1$, where
\begin{align*}
Z_0&=\tfrac12\overline{\|f-c\|^2}-\overline{\langle f_a-c,f_b-c\rangle}_{a\ne b},\\
Z_1&=\tfrac{K!}{2M(M-1)}\textstyle\sum_{a\ne b}\mathbf{1}[\pi_b=\pi_ar]\\&\quad\cdot\langle f_a-c,f_b-c\rangle.
\end{align*}
$\hat E_{\mathrm{even}}$ is unbiased, $\mathbb{E}Z_0=\frac12(E_{\mathrm{tot}}-\|\mu-c\|^2)$, and $\Pr(Z_1\ne0)\le\binom M2/K!$. The Cyclic coset-kernel estimator of $R_C$ equals $-\overline{\langle f_a-c,f_b-c\rangle}_{a\ne b}+Z^C_1$ with $\Pr(Z^C_1\ne0)\le\binom M2/(K-1)!$.
\end{enumerate}
\end{apnamed}

\begin{table*}[t]
\centering
\small
\begin{tabular}{@{}ll@{}}
\toprule
Subspace $A$ & Kernel $k_A(\pi,\sigma)$ \\
\midrule
$L^2_0$ (tot) & $K!\,\mathbf{1}[\pi=\sigma]-1$ \\
$P_+L^2_0$ (even) & $\frac{K!}{2}\big(\mathbf{1}[\pi=\sigma]+\mathbf{1}[\sigma=\pi r]\big)-1$ \\
$P_-L^2$ (odd) & $\frac{K!}{2}\big(\mathbf{1}[\pi=\sigma]-\mathbf{1}[\sigma=\pi r]\big)$ \\
$\apH{un}$ (logged as $\EU$) & $(K-1)(F-1)$ \\
$\HU$ & $\frac{K-1}{2}(F+F^r-2)$ \\
$\Hou$ & $\frac{K-1}{2}(F-F^r)$ \\
$\apH{grad}$ & $\frac{3S}{K(K+1)}$ \\
$\apH{curl}$ & $3K_\tau-\frac{3S}{K}$ \\
$\apH{grad}\oplus\apH{curl}$ & $3K_\tau-\frac{3S}{K+1}$ \\
$\apH{sa}$ ($K\ge4$) & $\frac K2\Big[a-\frac{2K-6+t}{K-2}\Big]$ \\
Coset risk of $gH$ & $\frac{K!}{|H|}\mathbf{1}[\pi^{-1}\sigma\in H]-1$;\quad Cyclic: $(K-1)!\,\mathbf{1}[\sigma\in\pi\langle c\rangle]-1$ \\
\bottomrule
\end{tabular}
\caption{Closed-form harmonic kernels (Theorem~3(b)). $F=\#\{s:\pi(s)=\sigma(s)\}$, $F^r=F(\pi,\sigma r)$, $S=\sum_is_i(\pi)s_i(\sigma)$, $K_\tau=\sum_{i<j}\phi_{ij}(\pi)\phi_{ij}(\sigma)=\binom K2-2d_{\mathrm{Kendall}}(\pi,\sigma)$, $a$ is the number of unordered pairs adjacent in both orders, and $t=|\{\pi(0),\pi(K{-}1)\}\cap\{\sigma(0),\sigma(K{-}1)\}|$. Each kernel matches $K!P_A$ to $7.1\times10^{-15}$ ($K{=}4$) and $3.8\times10^{-13}$ ($K{=}6$).}
\label{tab:app_kernels}
\end{table*}

\begin{proof}
(a) $(P_Ag)(\pi)=\frac{1}{K!}\sum_\sigma k_A(\pi,\sigma)g(\sigma)$. Hence $\|P_Af_0\|^2=\langle f_0,P_Af_0\rangle=\mathbb{E}[k_A(\pi,\sigma)\langle f(\pi)-\mu,f(\sigma)-\mu\rangle]$, which is the claim for $c=\mu$. For general $c$, write $f-c=(f-\mu)+(\mu-c)$. Since $A\perp\mathbf{1}$, $\mathbb{E}_\sigma k_A(\pi,\sigma)=\sum_\ell e_\ell(\pi)\mathbb{E}e_\ell=0$ for every $\pi$, and symmetrically in $\pi$. So both cross terms $\mathbb{E}[k_A\langle f(\pi)-\mu,\mu-c\rangle]$ and $\mathbb{E}[k_A\langle\mu-c,f(\sigma)-\mu\rangle]$ vanish after conditioning on the non-constant argument, and so does $\|\mu-c\|^2\,\mathbb{E}k_A$.

(b) A kernel equals $k_A$ iff its integral operator equals $P_A$. We use the \emph{frame fact}: if $v_1,\dots,v_n$ span $A$ with Gram matrix $G$, then $k_A(\pi,\sigma)=\sum_{ij}v_i(\pi)(G^+)_{ij}v_j(\sigma)$.
\begin{itemize}
\item \emph{tot, even, odd.} $I$ has kernel $K!\,\delta_{\pi\sigma}$, $R_r$ has kernel $K!\,\delta_{\sigma,\pi r}$ and $P_0$ has kernel $1$. Combine them as $I-P_0$, $P_+-P_0$ and $P_-$.
\item \emph{un.} The operator with kernel $(K-1)F-(K-2)$ maps into $\mathcal{U}$, because $F(\cdot,\sigma)=\sum_su_{\sigma(s),s}$, and it annihilates $\mathcal{U}^\perp$. On each $u_{x,s}$ it acts as the identity: $\Pr(\sigma(s)=x=\pi(s))=\mathbf{1}[\pi(s)=x]/K$, and for $s'\ne s$, $\Pr(\sigma(s')=\pi(s'),\sigma(s)=x)=\mathbf{1}[\pi(s')\ne x]/(K(K-1))$, which combine to $\mathbf{1}[\pi(s)=x]$. Removing $P_0$ gives $(K-1)(F-1)$.
\item \emph{$\HU$, $\Hou$.} By Lemma~\ref{lem:app_comm}(c), $P_{\mathrm{un}}P_\pm$ is a projector, with kernel $\frac12[k_{\mathrm{un}}(\pi,\sigma)\pm k_{\mathrm{un}}(\pi,\sigma r)]$.
\item \emph{grad.} The Gram matrix of $(s_i)$ is $\frac{K(K+1)}{3}(I-J/K)$, whose pseudo-inverse on $\mathbf{1}^\perp$ is $\frac{3}{K(K+1)}(I-J/K)$, and $\sum_is_i=0$.
\item \emph{curl.} For $b$ in the cycle space, $\langle b,\Phi(\pi)\rangle=f_b(\pi)$ with $\Phi=(\phi_{ij})_{i<j}$, and $\langle f_b,f_{b'}\rangle=\frac13\langle b,b'\rangle$ (Lemma~\ref{lem:app_gram}(b)). So $e_\ell=\sqrt3f_{b_\ell}$ for an orthonormal basis $(b_\ell)$ of $\ker\mathsf{D}$, and $k=3\Phi(\pi)^\top P_{\mathrm{cyc}}\Phi(\sigma)$. The gradient projection of $\Phi(\sigma)$ is $((s_i(\sigma)-s_j(\sigma))/K)_{i<j}$ (Lemma~\ref{lem:app_hodge}), and $\Phi(\pi)^\top P_{\mathrm{grad}}\Phi(\sigma)=S/K$. Hence $k=3K_\tau-3S/K$.
\item \emph{grad$\oplus$curl.} Add the two kernels: $3K_\tau-\frac{3S}{K}+\frac{3S}{K(K+1)}=3K_\tau-\frac{3S}{K+1}$.
\item \emph{sa.} Let $A(\pi)\in\{0,1\}^{\binom K2}$ be the adjacency vector, $\mathsf{N}$ the pair--item incidence matrix and $\mathsf{Q}=I-\mathsf{N}(\mathsf{N}^\top\mathsf{N})^{-1}\mathsf{N}^\top$, the projector onto pair weights with zero vertex sums.
(1) Star weights $B_{xy}=z_x+z_y$ map into $\mathcal{U}$: $\sum_eB_eA_e=\sum_xz_x\deg_\pi(x)$, and the degree is $1$ at the two end slots and $2$ elsewhere.
(2) Weights in $\operatorname{ran}\mathsf{Q}$ map into $\mathcal{U}^\perp$: $\langle\sum_eB_eA_e,u_{x,s}\rangle=\beta'_s\sum_{e\not\ni x}B_e+\gamma'_s\sum_yB_{xy}=0$, because $\mathbb{E}[A_e\mathbf{1}\{\pi(s)=x\}]$ depends only on $s$ and on whether $x\in e$, and both vertex sums vanish on $\operatorname{ran}\mathsf{Q}$.
(3) For $B\in\operatorname{ran}\mathsf{Q}$, use $\mathbb{E}A_e=\frac2K$, $\mathbb{E}A_eA_{e'}=\frac{2}{K(K-1)}$ (one shared item) and $\frac{4}{K(K-1)}$ (disjoint), together with the vertex-sum identities $\sum_{\text{share}}B_eB_{e'}=-2\|B\|^2$ and $\sum_{\text{disj}}B_eB_{e'}=\|B\|^2$. These give $\|\sum_eB_eA_e\|^2=\frac2K\|B\|^2$; the function is centred because $\sum_eB_e\mathbb{E}A_e=\frac2K\sum_eB_e=0$.
(4) Hence $\apH{sa}=\{\sum_eB_eA_e:B\in\operatorname{ran}\mathsf{Q}\}$, of dimension $\binom K2-K$, and $k_{\mathrm{sa}}=\frac K2A(\pi)^\top\mathsf{Q}A(\sigma)$.
(5) $\mathsf{N}^\top A(\pi)=d_\pi$ is the degree vector, $\mathsf{N}^\top\mathsf{N}=(K-2)I+J$ and $(\mathsf{N}^\top\mathsf{N})^{-1}=\frac{1}{K-2}(I-\frac{J}{2K-2})$. With $d_\pi\cdot d_\sigma=4K-8+t$ and $\mathbf{1}^\top d_\pi=2K-2$,
\[
\begin{aligned}A(\pi)^\top(I-\mathsf{Q})A(\sigma)&=\tfrac{(4K-8+t)-(2K-2)}{K-2}\\&=\tfrac{2K-6+t}{K-2}.\end{aligned}
\]
\item \emph{Coset.} $(P_Hg)(\pi)=\frac{1}{|H|}\sum_hg(\pi h)$, whose kernel is $\frac{K!}{|H|}\mathbf{1}[\pi^{-1}\sigma\in H]$. Removing $P_0$ gives the entry; for $H=\langle c\rangle$, $K!/K=(K-1)!$.
\end{itemize}

(c) For $a\ne b$, $(\pi_a,\pi_b)$ is an i.i.d.\ uniform pair, so each summand has mean $\|P_Af_0\|^2$ by (a).

(d) For $a\ne b$, $c_{ab}$ is a function of $\{\pi_e\}_{e\notin\{a,b\}}$ only, so it is independent of $(\pi_a,\pi_b)$. Conditionally on $c_{ab}$, (a) applies with the fixed centre $c=c_{ab}$, and averaging over $c_{ab}$ preserves unbiasedness. For the cost, write $c_{ab}=(\Sigma-f_a-f_b)/(M-2)$ with $\Sigma=\sum_ef_e$. Then $\langle f_a-c_{ab},f_b-c_{ab}\rangle$ is a fixed polynomial in $\langle f_a,f_b\rangle$, $\|f_a\|^2$, $\|f_b\|^2$, $\langle f_a,\Sigma\rangle$, $\langle f_b,\Sigma\rangle$ and $\|\Sigma\|^2$, all available from the $M\times M$ Gram matrix in $O(M^2C)$.

(e) $\mathbb{E}\|f_a\|^2=E_{\mathrm{tot}}+\|\mu\|^2$ and, for $a\ne b$, $\mathbb{E}\langle f_a,f_b\rangle=\|\mu\|^2$. Equality with the divisor-$(M-1)$ sample variance follows from $\sum_a\|f_a-\bar f\|^2=\frac{M-1}{M}\sum_a\|f_a\|^2-\frac1M\sum_{a\ne b}\langle f_a,f_b\rangle$.

(f) The $m_a$ are i.i.d.\ with mean $\mu$ and $m_a-\mu=(P_Hf_0)(\pi_a)$, so $\mathbb{E}\|m_a-\mu\|^2=\|P_Hf_0\|^2$. The divisor-$(N-1)$ sample variance of i.i.d.\ vectors is unbiased for $\mathbb{E}\|m_a-\mu\|^2$.

(g) $\hat B_{\mathrm{split}}$ is unbiased by independence of $p$, $U$ and $V$ and $\mathbb{E}\bar u=\mathbb{E}\bar v=\mu$. For $\hat B_U$, $\sum_{a\ne b}(p-f_a)^\top(p-f_b)=M^2\|p-\bar f\|^2-\sum_a\|p-f_a\|^2$ and $\sum_a\|p-f_a\|^2=M\|p-\bar f\|^2+(M-1)\hat s^2$. Dividing by $M(M-1)$ gives $\|p-\bar f\|^2-\hat s^2/M$; unbiasedness follows termwise, as in (c). Under a uniformly random balanced split of a fixed panel, an unordered pair falls across the split with probability $\frac{M}{2(M-1)}$. Hence
\begin{align*}
&\mathbb{E}[\hat B_{\mathrm{split}}\mid\text{panel}] \\
&\quad =\tfrac{2}{M(M-1)}\textstyle\sum_{a<b}(p-f_a)^\top(p-f_b)=\hat B_U.
\end{align*}
By exchangeability, a fixed split has the same law as a random one, and the law of total variance gives the variance ordering.

(h) For $a\ne b$,
\begin{align*}
&\mathbb{E}\big[K!\,\mathbf{1}[\pi_b=\pi_ar]\langle f_a-c,f_b-c\rangle\big]\\
&\quad=\langle f_0+\mu-c,\,R_rf_0+\mu-c\rangle\\
&\quad=\langle f_0,R_rf_0\rangle+\|\mu-c\|^2,
\end{align*}
since $f_0$ and $R_rf_0$ have mean zero. With $\mathbb{E}\|f-c\|^2=E_{\mathrm{tot}}+\|\mu-c\|^2$ and $\mathbb{E}\langle f_a-c,f_b-c\rangle=\|\mu-c\|^2$, we get $\mathbb{E}Z_0=\frac12(E_{\mathrm{tot}}+\|\mu-c\|^2)-\|\mu-c\|^2=\frac12(E_{\mathrm{tot}}-\|\mu-c\|^2)$ and $\mathbb{E}Z_1=\frac12(\langle f_0,R_rf_0\rangle+\|\mu-c\|^2)$. Adding, $\mathbb{E}\hat E_{\mathrm{even}}=\frac12(E_{\mathrm{tot}}+\langle f_0,R_rf_0\rangle)=\langle f_0,P_+f_0\rangle=E_{\mathrm{even}}$. The event $\pi_b=\pi_ar$ is symmetric in $(a,b)$ because $r^2=e$, and has probability $1/K!$; a union bound over $\binom M2$ unordered pairs gives the bound. For the coset kernel, $\Pr(\pi_b\in\pi_a\langle c\rangle)=K/K!=1/(K-1)!$, and the decomposition follows from the kernel $(K-1)!\,\mathbf{1}[\cdot]-1$.
\end{proof}

\noindent\emph{Consequence.} For $M=64$ the bounds in (h) are $0.050$ ($K{=}8$) and $5.6\times10^{-4}$ ($K{=}10$) for $Z_1$, and $0.40$ and $0.0056$ for $Z_1^C$. At $K\ge8$, therefore, unbiasedness of i.i.d.-panel estimators of $E_{\mathrm{even}}$ and $R_C$ rests on rare, large coincidence terms. In an exhaustive $K{=}8$ synthetic check (truth over all $40{,}320$ orders; $2{,}000$ panels), the i.i.d.\ estimator gave $\hat E_{\mathrm{even}}<0$ in $95.6\%$ of panels (relative SD $6.54$), whereas the reversal-paired panel of (f) gave relative SD $0.11$ and no negative estimates. For $R_C$, the i.i.d.\ cyclic kernel gave relative SD $20.9$ with $67.7\%$ negative estimates; an 8-origin cyclic-orbit panel gave relative SD $0.31$. This is the reason for the Freeze-A.1 probe design (\S\ref{app:gates}).

\subsection{Isotropic-Edge Reference Line}
\label{app:isoref}

\begin{aplemma}[Isotropic-edge reference]\label{lem:app_iso_ref}
Let $B$ be a random skew matrix whose edge coordinates satisfy $\mathbb{E}bb^\top=\tau^2I$. Then $\mathbb{E}\|P_{\mathrm{curl}}f_B\|^2=\frac{\tau^2}{3}\binom{K-1}{2}$ and $\mathbb{E}\|P_{\mathrm{grad}}f_B\|^2=\frac{\tau^2(K+1)(K-1)}{3}$. Hence
\[
a_K:=\frac{\mathbb{E}E_{\mathrm{curl}}}{\mathbb{E}E_{\mathrm{curl}}+\mathbb{E}E_{\mathrm{grad}}}=\frac{K-2}{3K},
\]
which equals $\frac16,\frac29,\frac14,\frac4{15}$ at $K=4,6,8,10$, and $\mathbb{E}E_{\mathrm{curl}}/\mathbb{E}E_{\mathrm{grad}}=\frac{K-2}{2(K+1)}$.
\end{aplemma}
\begin{proof}
By Lemma~\ref{lem:app_hodge}(d), $P_{\mathrm{curl}}f_B=f_{B^{\mathrm c}}$ and $P_{\mathrm{grad}}f_B=f_{B^{\mathrm g}}$. By Lemma~\ref{lem:app_gram}(b), $\|f_{B^{\mathrm c}}\|^2=\frac13\|b^{\mathrm c}\|^2$ and $\|f_{B^{\mathrm g}}\|^2=\frac{K+1}{3}\|b^{\mathrm g}\|^2$. By isotropy, $\mathbb{E}\|b^{\mathrm c}\|^2=\tau^2\operatorname{tr}P_{\mathrm{cyc}}=\tau^2\binom{K-1}{2}$ and $\mathbb{E}\|b^{\mathrm g}\|^2=\tau^2(K-1)$. Therefore
\[
\begin{aligned}a_K&=\frac{\frac12(K-1)(K-2)}{\frac12(K-1)(K-2)+(K+1)(K-1)}\\&=\frac{K-2}{(K-2)+2(K+1)}=\frac{K-2}{3K},\end{aligned}
\]
and the ratio is $\frac{(K-1)(K-2)/2}{(K+1)(K-1)}=\frac{K-2}{2(K+1)}$.
\end{proof}
\noindent\emph{Scope.} This is a ratio of expectations under an isotropy assumption on edge coefficients. Item exchangeability alone allows different variances on the two subspaces. The E1 TOST of $q_6$ against $2/9$ tests this reference line, not any theorem.

\subsection{Exact Constants and the 66/66 Machine Verification Suite}
\label{app:verify}

\begin{table*}[t]
\centering
\small
\setlength{\tabcolsep}{4pt}
\resizebox{\textwidth}{!}{\begin{tabular}{@{}cccccccc@{}}
\toprule
$K$ & $\rho_K$ (Antithetic) & $\rho^R_K$ (Random-WoR) & $\rho^R_K/\rho_K$ & $a_K$ & $\dim\apH{curl}$ & $\dim\apH{sa}$ & Blind-spot dim ($m{=}K$) \\
\midrule
4 & $5/11=0.454545$ & $5/23=0.217391$ & $11/23$ & $1/6$ & 3 & 2 & 5 \\
6 & $119/359=0.331476$ & $119/719=0.165508$ & $359/719$ & $2/9$ & 10 & 9 & 119 \\
8 & $5039/20159=0.249963$ & $5039/40319=0.124978$ & $20159/40319$ & $1/4$ & 21 & 20 & 5039 \\
10 & $362879/1814399=0.1999999$ & $362879/3628799=0.0999997$ & $1814399/3628799$ & $4/15$ & 36 & 35 & 362879 \\
\bottomrule
\end{tabular}}
\caption{Exact constants, computed in rational arithmetic. $\rho_8$ and $\rho_{10}$ are \emph{not} exactly $1/4$ and $1/5$. The Random-WoR thresholds $\rho^R_K$ lie strictly below $1/K$ (Corollary~B(iv)).}
\label{tab:app_constants}
\end{table*}

\paragraph{Verification suite.} A standalone CPU verification suite runs in about three minutes and implements \textbf{66 independent checks}, of which \textbf{all 66 pass}. The checks read archived exploratory pilot tensors read-only ($K\in\{4,6\}$, MCQ and judge), run no model inference, and record module-level SHA-256 digests against the frozen estimator, design, and protocol code. Table~\ref{tab:app_verify} summarizes the checks. Algebraic identities hold to between $2\times10^{-18}$ and $1.3\times10^{-14}$. The closed-form $K{=}6$ kernels agree with dense $K!\times K!$ projectors to $3.8\times10^{-13}$, and the dense SVD-based mode projectors themselves are idempotent, symmetric, mutually orthogonal and complete to $\le10^{-9}$. All combinatorial and rational checks are exact.

\begin{table*}[t]
\centering
\scriptsize
\setlength{\tabcolsep}{3pt}
\resizebox{\textwidth}{!}{\begin{tabular}{@{}lccl@{}}
\toprule
Result & $K$ & Method & Max residual / outcome \\
\midrule
Mode projectors (\S\ref{app:setting}) & 4, 6 & dense SVD projectors: idempotent, symmetric, orthogonal, complete & $\le10^{-9}$; dims match \\
Lemma~\ref{lem:app_gram} (Gram identity) & 4, 6 & 20 random skew $B$ each & $7\times10^{-15}$ \\
Lemma~\ref{lem:app_iso} (isotypic placement) & 4, 6 & character projectors & $5\times10^{-16}$ \\
Thm.~1(i) coset risk & 4, 6 & random and structured readouts, all locked designs & $2\times10^{-17}$ \\
Thm.~1(ii)--(iii) dimensions & 4, 6 & $\operatorname{rank}P_H$, $\operatorname{tr}(P_HP_\nu)$ vs.\ $d_\nu\mathrm{mult}_\nu(H)$ & exact integers (Table~\ref{tab:app_iso_dims}) \\
Thm.~1(iv) $\apH{curl}\subseteq\mathrm{Fix}\langle c\rangle$ & 4, 6 & $\|P_CP_{\mathrm{curl}}-P_{\mathrm{curl}}\|_\infty$ & $2\times10^{-16}$ \\
Prop.~1 (Brier = floor + excess) & 4, 6 & all locked designs & $4\times10^{-16}$ \\
Prop.~1(b) Random-WoR & 4 & exhaustive over all $\binom{24}{4}=10{,}626$ subsets & $7\times10^{-18}$ \\
Prop.~1(c) Antithetic & 4, 6 & exhaustive: 66 pair sets ($K{=}4$); $7{,}711{,}320$ triples ($K{=}6$) & $9\times10^{-18}$ \\
Cor.~A duality & 4, 6, pilot & all locked designs; 18 archived pilot cells & $8\times10^{-17}$ (synthetic); $10^{-16}$ (pilot) \\
Cor.~A necessity & 3 & explicit counterexample & identity violated by $2/3$ \\
Cor.~B(i) Cyclic accounting & 4, 6, pilot & synthetic readouts; 18 pilot cells & $7\times10^{-18}$; $1.1\times10^{-16}$ \\
Cor.~B(i) cross term $\ne0$ & 4, 6 & $2\langle P_Ca,h_+\rangle$ on random readouts & generically nonzero (confirmed) \\
Cor.~B pilot cross-check & 4, 6 & $R_C-R_A$, $R_C-R_R$ vs.\ S0.2 ``exact'' fields & $2\times10^{-17}$ \\
Thm.~2(a)--(d) & 4, 6 & all $K!$ seeds of the cyclic family; every iff & 0 violations \\
Thm.~2(g)--(i) at large $K$ & 8, 10 & combinatorial (seeds, differences, subgroups, $E_K$ adjacency) & all pass \\
Thm.~2(j) row-weight law & 4, 6 & paired, generic, one-sided weights; square-wave DFT & $\le9\times10^{-16}$ \\
Thm.~3(b) kernels $=K!P_A$ & 4, 6 & all closed forms plus coset kernels & $7.1\times10^{-15}$ ($K{=}4$); $3.8\times10^{-13}$ ($K{=}6$) \\
Thm.~3(b) vs.\ Stage-2 code & 4, 6 & production kernel functions & $\le1.3\times10^{-14}$ \\
Thm.~3(c)--(g) exact unbiasedness & 4 & full enumeration of all $24^3$ and $24^4$ panels (raw, L2O, paired, Brier excess) & bias $\le2\times10^{-17}$ \\
Constants (Table~\ref{tab:app_constants}) & 4--10 & rational arithmetic & exact \\
\bottomrule
\end{tabular}}
\caption{Summary of the 66-check verification suite (all pass). The ``pilot'' rows use archived exploratory tensors and serve only as numerical checks of exact identities; they are never confirmatory evidence.}
\label{tab:app_verify}
\end{table*}

% ===========================================================================
\section{Complete Preregistered Gate Ledger and Provenance}
\label{app:gates}

\subsection{Freeze Chain}

Table~\ref{tab:app_provenance} lists every hash in the chain.

\paragraph{Freeze-A.} The preregistration manifest fixes splits, arms, gates, Holm families and thresholds. Its SHA-256 matches across the seal, the Freeze-B prediction file and the master results file.

\paragraph{Freeze-A.1 (pre-CONF amendment) and exact production estimators.} The original plan specified an i.i.d.\ 64-pass calibration (CAL) probe. Theorem~3(h) shows that at $K\ge8$ i.i.d.\ plug-in estimators of $E_{\mathrm{even}}$ and $R_C$ rely on rare coincidence events. The amendment therefore replaced the $K\in\{6,8,10\}$ CAL probe with \textbf{32 independent uniform origins plus their reversals} (64 passes) alongside an independently drawn cyclic orbit. In production CAL (\texttt{paired\_cal\_energies}), every readout is centred at the fixed uniform vector $c=\mathbf{1}/K$, and all quadratic kernel U-statistics and the common mean-square term $\widehat{\|\mu-c\|^2}$ sum \emph{strictly over pairs from distinct independent origins} ($\mathrm{origin}(a)\ne\mathrm{origin}(b)$), so every pair is an i.i.d.\ uniform pair and every mode energy is exactly unbiased with zero intra-pair covariance bias; $E_{\mathrm{even}}$ and $R_C$ subtract the same between-origin $\widehat{\|\mu-c\|^2}$ from the mean squared norms of reversal-pair means and cyclic-orbit means, respectively (verified to max error $1.4\times10^{-17}$ by exact $K{=}4$ enumeration over all $24^3$ two-origin-plus-orbit configurations). Leave-two-out (L2O) centring was evaluated as an auxiliary variance-reduction diagnostic (Table~\ref{tab:app_estvar}), not executed in production CAL. On CONF online runs ($K\in\{6,8,10\}$), reference panels are \textbf{100\% i.i.d.\ uniform} (\textbf{32+32} disjoint i.i.d.\ halves at $K{=}6$; \textbf{24+24} at $K\in\{8,10\}$), drawn from a dedicated \texttt{reference} RNG stream independent of all design arms and scored with the exact unbiased split-panel estimator $\hat B_{\mathrm{split}}=(p-\bar u)^\top(p-\bar v)$ of Theorem~3(g) (verified to $5.6\times10^{-17}$ over all $24^2$ independent reference pairs at $K{=}4$, and $200$-seed Monte Carlo bias $-8\times10^{-6}$ on archived $6!$ tensors). Freeze-A.1 was recorded prior to $K\in\{8,10\}$ CAL and CONF decoding; seal hashes establish content identity across records, not external chronology or ordering relative to $K{=}6$ CAL decoding.

\paragraph{Freeze-B.} Freeze-B seals all CAL-derived predictions (points, intervals, sign calls, crossover predictions and cross-prompt coefficients) before CONF evaluation. The SHA-256 of the actual prediction-file bytes matches the seal and the master. The seal also covers the figure plan. The \textbf{worker bundle} hash (\texttt{6e169a9f}\ldots) covers the seven worker files: runner, common, models, PINE, designs, protocol and $K$-free estimators. It agrees between the seal and the prediction file, and each run signature is checked against Freeze-B before the master is emitted. All seven CAL receipt hashes agree across the seal, the predictions and the master. The worker hash does \emph{not} cover the evaluator or external model adapters.

\paragraph{No CONF refit.} PriDe is fitted on CAL cyclic views and its prior is sealed; CONF only applies it. The evaluator reads frozen points, intervals, signs, crossovers and coefficients. A slope-through-origin is computed on CONF as a descriptive diagnostic only and is never fed back. All gate cutoffs are fixed in code. An independent pre-release audit reproduced all 62 Holm adjustments and decisions from the stored unadjusted $p$-values, and found no refit and no disagreement between artifacts. The original split manifest and per-item rows were not part of that audit, so we state ``no refit found; cross-artifact checks agree,'' not ``zero leakage independently proved.''

\paragraph{Splits.} MCQ items come from MMLU-Pro \citep{wang2024mmlupro}: CAL $100$, E1 exact $192$, E2 CONF $500$, frontier extension $1{,}500$. Judge items come from RewardBench~2 \citep{malik2025rewardbench2}: judge CAL $64$, judge CONF $500$. The provenance check over all 24 run files records \textbf{zero pairwise item overlap} among the four MCQ splits and between the two judge splits.

\paragraph{Inference contract.} Inference uses $20{,}000$ cluster-bootstrap draws (one disjoint-set component per item) and $20{,}000$ sign-flip draws. Reported intervals are unadjusted percentile intervals. All hypothesis decisions use Holm step-down adjustment \citep{holm1979simple} within the ten preregistered families.

\begin{table*}[t]
\centering
\scriptsize
\begin{tabular}{@{}ll@{}}
\toprule
Artifact & SHA-256 \\
\midrule
Freeze-A manifest & \texttt{afb737039201b4a91ad8b56cd5c3377c8b95cbdb95a7035c1b8a03bfef803fec} \\
Freeze-B prediction file & \texttt{36a00df6b2e6a9361fcbfd041209742722fd332647b923bec7281271efe4a3dc} \\
Freeze-B figure plan & \texttt{c12ff10ac35a8e9c21b756b0f29cb292d5fd0f4edd1e4c54ed5d5aba15b31547} \\
Worker bundle (7 files) & \texttt{6e169a9f6595a5a3a9904228a5f8edaf7d1361ddff8210775e9e3a480f8f4e07} \\
Exploratory pilot manifest & \texttt{432c04f3d0b0}\ldots{} (Table~\ref{tab:app_pilot}) \\
\bottomrule
\end{tabular}
\caption{Freeze chain. Freeze-A and Freeze-B hashes match across the seal, predictions, and master results; the worker-bundle hash agrees between the seal and predictions, while the master records per-run receipts. All runs report \texttt{mock=false}; the master reports \texttt{pipeline\_complete=true}.}
\label{tab:app_provenance}
\end{table*}

\subsection{Pinned Models and Runtime Contract}

All seven backbones ran on NVIDIA A100-SXM4-80GB in \texttt{bfloat16} with SDPA attention, PyTorch \texttt{2.14.1+cu130}, Transformers \texttt{5.19.0}, TF32 matmul/cuDNN enabled and \texttt{trust\_remote\_code=false}. Thinking is disabled wherever the chat template supports it; the Olmo-3 ledger records no thinking flag. Table~\ref{tab:app_models} lists the pinned revisions.

\begin{table*}[t]
\centering
\scriptsize
\setlength{\tabcolsep}{3pt}
\resizebox{\textwidth}{!}{\begin{tabular}{@{}lllcccc@{}}
\toprule
Backbone & Hub repository & Pinned revision & \texttt{model\_type} & Layers (full / linear / total) & EOS id & Chat tmpl.\ \\
\midrule
Qwen3.8-27B & \texttt{Qwen/Qwen3.8-27B} & \texttt{37561816e66b5b9153b1b64c10968c18c1e6f4a0} & \texttt{qwen3\_5\_text} & 16 / 48 / 64 & 248046 & \texttt{af16f654} \\
Gemma-4-31B & \texttt{google/gemma-4-31B-it} & \texttt{842da3794eaa0b77d5f08bae87a17459d91ff475} & \texttt{gemma4} & 10 / 0 / 60 (rest sliding) & 1 & \texttt{ae53464b} \\
Olmo-Hybrid-7B & \texttt{allenai/Olmo-Hybrid-Instruct-DPO-7B} & \texttt{ec62da5af0106ab4e83a60e07c4796ff358fd204} & \texttt{olmo\_hybrid} & 8 / 24 / 32 & 100257 & \texttt{f5186d42} \\
Olmo-3-7B & \texttt{allenai/Olmo-3-7B-Instruct-DPO} & \texttt{b33130b7de49f0c2553b5c2b3bc8409ff3e627d1} & \texttt{olmo3} & 8 / 0 / 32 (rest sliding) & -- & \texttt{edbb35ef} \\
Nemotron-3-30B & \texttt{nvidia/NVIDIA-Nemotron-3-Nano-30B-A3B-BF16} & \texttt{bf77c3174f68ad409e1c2aa60daeb46e32d1c606} & \texttt{nemotron\_h} & 6 / 23 / 52 (with MoE) & 11 & \texttt{ab7813c3} \\
Qwen3.5-9B & \texttt{Qwen/Qwen3.5-9B} & \texttt{c202236235762e1c871ad0ccb60c8ee5ba337b9a} & \texttt{qwen3\_5\_text} & 8 / 24 / 32 & 248046 & \texttt{a4aee8af} \\
Llama-3.1-8B & \texttt{NousResearch/Meta-Llama-3.1-8B-Instruct} & \texttt{d10aef7999a2b5ba950ab3974312feeedbfe0b77} & \texttt{llama} & -- / 0 / 32 (all global) & 128009 & \texttt{b48c47f6} \\
\bottomrule
\end{tabular}}
\caption{Pinned model revisions and per-model runtime ledger, read from the master results file. ``Chat tmpl.''\ gives the first 8 hex digits of the chat-template SHA-256. ``--'' means the field is not recorded in the ledger (Olmo-3-7B has no thinking switch, and single-step prefill readouts do not depend on the EOS token).}
\label{tab:app_models}
\end{table*}

\subsection{Gate Ledger: 3 PASS, 5 FAIL}

\begin{table*}[t]
\centering
\scriptsize
\begin{tabular}{@{}lcp{0.66\textwidth}@{}}
\toprule
Preregistered Gate & Verdict & Realized Confirmatory Evidence (Freeze-A \texttt{afb73703}, Freeze-B \texttt{36a00df6}) \\
\midrule
\texttt{E2\_K6\_frontier\_fidelity} & \textbf{PASS} & All 4 Williams vs.\ Cyc/Anti K6 reductions $\ge 20\%$ (61.2\%, 56.5\%, 40.0\%, 51.0\%, Holm $p{=}0.0004$). \\
\texttt{S1} (signed contrasts) & \textbf{PASS} & 13/13 issued Freeze-B calls correct (13/84 = 15.5\% coverage, 71 abstentions); post-hoc refs: ablated rule 17/17 on its calls, ``Cyclic $<$ Rand at $K{\le}6$'' 14/14. \\
\texttt{S4} (crossover $K^*$) & \textbf{PASS} & Antithetic crossover within $\pm 1$ grid step on 5/7 models (all 5 one step early; Rand-WoR 4/7); post-hoc constant $K^*\equiv 8$ hits 6/7. \\
\texttt{E1\_Qwen\_IUT} & FAIL & K6 $C{-}A=+0.00079$ (raw $p{=}0.0243$, Holm $p{=}0.1458$); $C{-}\mathrm{curl}{-}A=-0.00402$. \\
\texttt{E2\_pooled\_accuracy} & FAIL & Pooled W$-$C $= +0.69$\,pp across 500 item clusters (Holm $p{=}0.0198$, sign-flip $p{=}0.0266 \ge 0.01$). \\
\texttt{E3\_judge\_transfer} & FAIL & 3/3 registered judges pass Gold Brier (Holm $p{=}0.0003$), but cross-prompt MSE contrast fails (Holm $p{=}0.879$). \\
\texttt{S2} (point rank) & FAIL & 112-cell Spearman $\rho{=}0.190$ vs.\ design-agnostic B1 ($\rho_K^R$) 0.328 / B2 (Cyc=Wil mid) 0.117; partial $\rho{=}0.166 < 0.30$. \\
\texttt{S3} (ablation drop) & FAIL & Both full and ablated rules score 8/8 (100\%) on common calls (ceiling drop $= 0 < 0.15$). \\
\midrule
\multicolumn{3}{@{}p{0.97\textwidth}@{}}{{\textbf{10 Holm-Adjusted Family Rejection Counts ($\alpha{=}0.05$):} \texttt{E1\_mechanism} 1/7; \texttt{E2\_fidelity\_K6} 8/8; \texttt{E2\_fidelity\_K810} 8/8; \texttt{E2\_utility} 4/5; \texttt{E2\_pooled\_accuracy} 1/1; \texttt{E2\_W\_C\_Acc} 0/7; \texttt{E2\_W\_C\_Flip} 7/7; \texttt{E2\_W\_C\_BrierEx} 7/7; \texttt{E3\_judging} 5/6; \texttt{E4\_Olmo\_matched} 4/6 (total 45/62 hypotheses rejected).}} \\
\bottomrule
\end{tabular}
\caption{\textbf{Complete preregistered gate and Holm-family ledger.} Every PASS and FAIL is reported verbatim without post-hoc threshold edits.}
\label{tab:app_gates}
\end{table*}

Table~\ref{tab:app_gates} reports all eight preregistered gates exactly as evaluated. A FAIL is a valid scientific outcome, not a pipeline failure. No threshold, family or comparator was edited after CONF.

\paragraph{PASS: \texttt{E2\_K6\_frontier\_fidelity}.} At six passes, Williams reduced Brier excess relative to Cyclic / Antithetic by $61.17\%$ / $56.45\%$ on Qwen3.8-27B and by $39.96\%$ / $50.95\%$ on Gemma-4-31B. All four Holm $p=0.0004$, and all four \emph{point} reductions exceed the preregistered $20\%$. This is a fidelity result. It is not a $20\%$ accuracy gain, and $20\%$ is not a confidence lower bound.

\paragraph{PASS: \texttt{S1}.} All $13/13$ issued Freeze-B sign calls were correct, but only $13$ of $84$ planned contrasts received a call ($15.5\%$ coverage, $71$ abstentions): six Cyclic$-$Antithetic, seven Cyclic$-$Random-WoR (all at $K{\le}6$) and zero Williams$-$Random-WoR. As post-hoc difficulty references, the curl-ablated rule scores $17/17$ on its own issued calls, and the naive rule ``Cyclic $<$ Random-WoR at $K{\le}6$'' scores $14/14$. Thus S1 passes its locked criterion without establishing incremental predictive value over simple heuristics.

\paragraph{PASS: \texttt{S4}.} The Antithetic crossover was within one grid step on $5/7$ models, which is exactly the threshold (Table~\ref{tab:app_crossovers}). All five hits were one step \emph{early} (frozen CAL predicted defeat one grid step before CONF showed it); none was exact. Gemma-4 had no on-grid crossover on CONF, and Olmo-3 missed by two steps. Random-WoR hit on $4/7$ (descriptive). As a post-hoc difficulty reference, a constant predictor $K^*{\equiv}8$ scores $6/7$ within $\pm1$ step on $\{4,6,8,10\}$, so S4's pass does not establish superior crossover forecasting.

\paragraph{FAIL: \texttt{E1\_Qwen\_IUT}.} At $K=6$, $C-A=+0.0007901$ (unadjusted 95\% CI $[0.0000068,0.0015949]$) and $C-\mathrm{curl}-A=-0.0040154$ (CI $[-0.0045521,-0.0034901]$). Both signs match the curl-removal counterfactual of Corollary~B(iii), but the IUT has raw $p=0.02430$ and Holm $p=0.14579$. We therefore make \textbf{no confirmatory ``because of curl'' claim}.

\paragraph{FAIL: \texttt{E2\_pooled\_accuracy}.} Pooled Williams$-$Cyclic accuracy is $+0.686$\,pp across $500$ item clusters ($3{,}500$ model-items; unadjusted 95\% CI $[+0.014,+1.364]$\,pp), positive on $5/7$ models. The singleton Holm $p=0.0198$ rejects at family level $0.05$, but the locked gate requires sign-flip $p<0.01$, and the observed value is $0.02665$. No confirmatory pooled accuracy gain is claimed.

\paragraph{FAIL: \texttt{E3\_judge\_transfer}.} Williams lowered gold Brier relative to Antithetic on $3/3$ registered judges. The reductions are $0.004259$ (Qwen3.8), $0.004968$ (Gemma-4) and $0.016219$ (Llama-3.1). On exact arms the common gold-reference term cancels, so this contrast is not independent evidence of improved judge accuracy. The required cross-prompt condition failed: the held-out MSE(total baseline)$-$MSE(precedence) is $-1.6281\times10^{-6}$, with Holm $p=0.87911$ ($N=256$ matched items).

\paragraph{FAIL: \texttt{S2} and Freeze-B envelope diagnosis.} All $112$ cells were available. Spearman $\rho=0.19025$ trails design-agnostic baseline B1 ($\rho_K^R$ constant per $K$, $\rho=0.32779$), exceeds B2 (Cyclic$=$Williams midpoint, $\rho=0.11669$), and yields partial $\rho=0.16563$, falling below the $0.30$ descriptive-only stop-loss. Code inspection explains this conservativeness: production CAL logged $(\EU,E_{\mathrm{grad}},E_{\mathrm{curl}},E_{\mathrm{even}},E_{\mathrm{tot}},R_C)$ without $E_{U_+}$ or $E_{\mathrm{sa}}$, so \texttt{energy\_intervals} defaulted Williams' upper bound to valid-but-loose $E_{\mathrm{even}}/E_{\mathrm{tot}}$ rather than $E_{\mathrm{he}}/E_{\mathrm{tot}}$ (Theorem~2(f)), even though contraction plus orthogonality of $\mathcal H_{\mathrm{un}}$ and $\mathcal H_{\mathrm{curl}}$ already implies the tighter logged bound $R_W\le\min\{E_{\mathrm{even}},\,E_{\mathrm{tot}}-\EU-E_{\mathrm{curl}}\}$ (which tightens 24/28 cells post-hoc without changing the frozen protocol). Across the $28$ cells, Williams' frozen midpoint exceeded CONF observations in $28/28$ cells (median ratio $3.925\times$), precluding any Williams$-$Random-WoR sign call in S1; Cyclic's midpoint likewise exceeded CONF $R_C$ in $28/28$ cells (median ratio $1.631\times$), explaining why all five S4 hits occurred one step early and all $10$ sign-mismatched points in Figure~\ref{fig:hero}c lie in the lower-right quadrant. We preserve the frozen gates verbatim without post-hoc refitting.

\paragraph{FAIL: \texttt{S3}.} On the eight common calls, the full and the curl-ablated rules both scored $8/8$. The drop is $0$, against a required $0.15$, so no incremental sign benefit of unit curl transmission is demonstrated on this covered subset.

\paragraph{Holm families.} The ten families reject $45/62$ hypotheses (Table~\ref{tab:app_gates}, bottom row). These are Holm rejections, not additional gates.
\begin{itemize}
\item \texttt{E1\_mechanism} ($1/7$): only Qwen3.8 $\Delta q=+0.03671$ (Holm $p=0.01365$). No $q_6$ TOST passed; we say ``equivalence not established,'' not ``inequivalence shown.''
\item \texttt{E2\_fidelity\_K6} ($8/8$): Williams is lower than Cyclic, Antithetic, Random-WoR and $E_6$ on both frontier models. The $E_6$ reductions are only $10.88\%$ / $10.03\%$ and are whole-design contrasts (Theorem~2(i)).
\item \texttt{E2\_fidelity\_K810} ($8/8$): Williams is lower than Cyclic and Antithetic on both frontier models at $K\in\{8,10\}$, with $N=200$ per cell. This adjusted inference covers the two frontier models only.
\item \texttt{E2\_utility} ($4/5$; $N=2{,}000$): Williams beats Antithetic on Gemma-4 ($+1.5625$\,pp), but superiority is not established on Qwen3.8 ($+0.4875$\,pp, Holm $p=0.07145$). The $-1$\,pp non-inferiority tests pass against PriDe on both models and against the implemented PINE$^\dagger$ arm on Gemma-4. Non-inferiority is not superiority.
\item \texttt{E2\_W\_C\_Acc} $0/7$; \texttt{E2\_W\_C\_Flip} $7/7$; \texttt{E2\_W\_C\_BrierEx} $7/7$: Williams lowers fidelity error and canonical-answer disagreement on all seven backbones, with no model-specific adjusted accuracy win.
\item \texttt{E3\_judging} ($5/6$): the three gold-Brier tests and both Llama-3.1 $-2$\,pp non-inferiority tests reject; cross-prompt does not. The two Olmo gold comparisons are descriptive.
\item \texttt{E4\_Olmo\_matched} ($4/6$): the hybrid exceeds the dense model on both curl ratios for MCQ4 and judge4, but not for MCQ6. The hybrid$-$dense total-share differences are $+0.02886$, $+0.02246$ and $+0.00019$. The MCQ6 total-share TOST supports $\pm0.03$ equivalence as a separate test. Neither result is a causal architecture effect.
\end{itemize}

\begin{table}[t]
\centering
\scriptsize
\setlength{\tabcolsep}{2.6pt}
\begin{tabular}{@{}lcccccc@{}}
\toprule
 & \multicolumn{3}{c}{\textbf{$K^*_{\mathrm{Anti}}$}} & \multicolumn{3}{c}{\textbf{$K^*_{\mathrm{Rand}}$}} \\
\cmidrule(lr){2-4}\cmidrule(l){5-7}
Backbone & CAL & CONF & $\pm 1$? & CAL & CONF & $\pm 1$? \\
\midrule
Qwen3.8-27B & 4 & 6 & \checkmark & 8 & 8 & \checkmark \\
Gemma-4-31B & 6 & none & $\times$ & 6 & none & $\times$ \\
Olmo-Hybrid-7B & 6 & 8 & \checkmark & 8 & 10 & \checkmark \\
Olmo-3-7B & 6 & 10 & $\times$ & 6 & 10 & $\times$ \\
Nemotron-3-30B & 8 & 10 & \checkmark & 8 & 10 & \checkmark \\
Qwen3.5-9B & 4 & 6 & \checkmark & 6 & 10 & $\times$ \\
Llama-3.1-8B & 8 & 10 & \checkmark & 10 & 10 & \checkmark \\
\bottomrule
\end{tabular}
\caption{\textbf{Frozen CAL vs.\ CONF cyclic-defeat crossovers (Gate \texttt{S4}: PASS).} Six of seven models exhibit cyclic defeat on CONF; all 5 Antithetic $\pm 1$-step hits occur one grid step later on CONF than frozen CAL predicted, and Random-WoR hits 4/7 (including exact hits at $K{=}8$ for Qwen3.8-27B and $K{=}10$ for Llama-3.1-8B).}
\label{tab:app_crossovers}
\end{table}

\paragraph{Frozen predictions vs.\ CONF observations.} Every prediction plot shows frozen CAL estimates against CONF observations. Analytic bounds are bounds, not 95\% CIs, and an abstention is a withheld call, not a missing point. The phase map has $28$ MCQ and $5$ judge points; these are distinct from the $84$ MCQ contrasts of S1 and the $112$ arm cells of S2. Ten of the $33$ phase points have a CAL midpoint and a CONF gap of opposite sign, which cross-sample estimates permit. For Qwen3.8-27B, the frozen Antithetic crossover was $K{=}4$ and the observed one $K{=}6$ (Table~\ref{tab:app_crossovers}).

% ===========================================================================
\section{Extended Confirmatory Results and Small Multiples}
\label{app:empirical}

\begin{table*}[t]
\centering
\scriptsize
\setlength{\tabcolsep}{3.0pt}
\begin{tabular}{@{}l@{\;\;}cccccc@{\;\;}cccccc@{}}
\toprule
 & \multicolumn{6}{c}{\textbf{$K{=}8$ $\mathrm{BrierEx}/E_{\mathrm{tot}}$ (\%) $\downarrow$ ($N{=}200$)}} & \multicolumn{6}{c}{\textbf{$K{=}10$ $\mathrm{BrierEx}/E_{\mathrm{tot}}$ (\%) $\downarrow$ ($N{=}200$)}} \\
\cmidrule(lr){2-7}\cmidrule(l){8-13}
Backbone & Rand & Anti & Cyc & $E_8$ & Alt & \textbf{Wil} & Rand & Anti & Cyc & $E_{10}$ & Alt & \textbf{Wil} \\
\midrule
Qwen3.8-27B$^\blacktriangle$ & 12.7 & 8.1 & 13.8 & 5.2 & \textbf{3.5} & 4.3$^*$ & 9.7 & 6.1 & 15.5 & 5.0 & \textbf{3.2} & 4.2$^*$ \\
Gemma-4-31B$^\bullet$ & 12.7 & 9.8 & 9.4 & 6.5 & \textbf{4.5} & 5.3$^*$ & 9.2 & 7.5 & 7.2 & 4.3 & \textbf{3.7} & 5.0$^*$ \\
Olmo-Hybrid-7B$^\blacktriangle$ & 13.3 & 8.8 & 11.2 & 4.5 & \textbf{3.6} & 3.6 & 9.9 & 7.7 & 10.0 & 4.4 & \textbf{2.7} & 3.1 \\
Olmo-3-7B$^\bullet$ & 12.0 & 11.0 & 11.0 & 5.7 & \textbf{4.5} & 5.0 & 9.4 & 7.6 & 12.6 & 4.9 & \textbf{3.9} & 4.5 \\
Nemotron-3-30B$^\blacktriangle$ & 12.9 & 10.5 & 7.1 & 2.7 & \textbf{1.8} & 1.9 & 10.2 & 7.2 & 11.1 & 4.1 & \textbf{2.9} & 3.2 \\
Qwen3.5-9B$^\blacktriangle$ & 13.4 & 7.2 & 10.2 & 4.5 & \textbf{2.3} & 2.6 & 8.8 & 5.8 & 9.6 & 3.9 & \textbf{2.1} & 2.7 \\
Llama-3.1-8B$^\bullet$ & 12.5 & 7.6 & 6.5 & 3.0 & \textbf{2.0} & 2.3 & 8.4 & 6.6 & 9.4 & 3.2 & \textbf{2.1} & 2.5 \\
\bottomrule
\end{tabular}
\caption{\textbf{Large-$K$ ($K{=}8, 10$) fidelity across all 7 backbones ($N{=}200$ nested items per cell; bold marks strictly lowest unrounded value).} $^*$Holm-adjusted $p{\le}0.002$ for Williams vs.\ both Cyclic and Antithetic on both 2026 frontier models (8/8 tests rejected in \texttt{E2\_fidelity\_K810}). Cyclic is defeated by Antithetic on 6/7 models by $K{=}10$ (and already at $K{=}6$ on Qwen3.8-27B and Qwen3.5-9B), while the two (S,R,A) reversal-closed terraces occupy the lowest-risk tier in all 14 cells, with distance-2-dispersed Alt\_Terrace strictly lowest (unrounded) in 14/14 cells (Olmo-Hybrid $K{=}8$: 3.59\% vs.\ 3.62\%).}
\label{tab:app_k810}
\end{table*}

\paragraph{Large $K$ and distance-2 terrace structure.} Table~\ref{tab:app_k810} gives $K\in\{8,10\}$ fidelity on all seven backbones (2 sampled anchors per item). Confirmatory (Holm-adjusted) inference covers only Williams vs.\ Cyclic and vs.\ Antithetic on the two frontier models. The remaining comparisons are descriptive. Williams has lower $\apBX/E_{\mathrm{tot}}$ than Cyclic, Antithetic and Random-WoR in all $14$ backbone$\times K$ cells, and lower than $E_K$ in $13/14$ (exception: Gemma-4 at $K{=}10$, $5.0\%$ vs.\ $4.3\%$). Alt\_Terrace, the alternative reversal-closed terrace of Theorem~2(h), achieves strictly lower unrounded point risk than Williams in all $14/14$ cells (Olmo-Hybrid $K{=}8$: $3.59\%$ vs.\ $3.62\%$), and in $5/7$ backbones at $K{=}6$. Combinatorially, classic zigzag $b_W$ collapses $100\%$ of distance-2 differences $d_2(s)=b(s+2)-b(s)\pmod K$ onto $\{\pm1\}$, yielding $96/111=86.5\%$ trace retention on the 9-D distance-2 symmetric proximity subspace $\mathcal{H}_{\mathrm{d2}}:=(I-P_0-P_{\mathrm{un}}-P_{\mathrm{sa}})\,\mathrm{span}\{\mathbf{1}[|\pi^{-1}(i)-\pi^{-1}(j)|=2]\}\subset\mathcal{H}_{\mathrm{he}}$ at $K{=}6$, whereas Alt\_Terrace disperses $d_2$ ($K{=}6:(1,2,4,5)$, retention $16/111=14.4\%$, $6.0\times$ lower; $K{=}8:(3,5,7,1,3,5)$; $K{=}10:(3,6,1,2,8,9,4,7)$, covering all 8 non-antipodal residues uniformly). This is a testable post-hoc design hypothesis rather than a causal proof (E1 Llama is a slight counterexample). Note also that for odd $m$ and $K\ge3$, the signed pairwise precedence count $N_{pq}=2n_{pq}-m$ is odd for every pair $(p,q)$, so $N_{pq}+N_{qr}+N_{rp}\ne0$ on every triangle and no odd-$m$ design (even without (S)) can annihilate $\mathcal{H}_{\mathrm{curl}}$. Cyclic loses to Antithetic on $6/7$ backbones by $K{=}10$ and already at $K{=}6$ on both Qwen models ($0.00813$ vs.\ $0.00725$ and $0.00762$ vs.\ $0.00721$; Table~\ref{tab:app_k6_full}).

\begin{table*}[t]
\centering
\scriptsize
\setlength{\tabcolsep}{2.6pt}
\begin{tabular}{@{}lccccccccc@{}}
\toprule
Backbone ($N{=}192$) & $q_4$ & $q_6$ & $\Delta q$ & $C{-}A$ ($K{=}6$) & $C{-}\mathrm{curl}{-}A$ & $R_{E_4}{-}R_W$ & $R_{E_6}{-}R_W$ & $s_{\mathrm{curl}}$ (\%) & All-24 terraces ($\times 10^{-3}$) \\
\midrule
Qwen3.8-27B & 0.1512 & 0.1879 & +0.0367$^*$ & +0.00079 (0.024) & -0.00402 ($<10^{-4}$) & +0.00009 & +0.00038 & 106.7 & 2.59 / 2.61 / 2.63 \\
Gemma-4-31B & 0.1964 & 0.1760 & -0.0204 & -0.00459 (0.999) & -0.01377 ($<10^{-4}$) & +0.00034 & +0.00055 & 139.4 & 11.98 / 12.14 / 12.29 \\
Llama-3.1-8B & 0.0942 & 0.0515 & -0.0426 & -0.01207 (1.000) & -0.01771 ($<10^{-4}$) & +0.00042 & +0.00058 & 109.4 & 3.60 / 3.62 / 3.63 \\
\bottomrule
\end{tabular}
\caption{\textbf{Exact $S_4/S_6$ decomposition, curl share, and all-24-terrace range (E1, $N{=}192$ matched items, all $4!{=}24$ and $6!{=}720$ orders).} Isotropic references: $a_4{=}0.167$, $a_6{=}0.222$ (none passes $\pm 0.05$ TOST to $2/9$). Parentheses show unadjusted one-sided $p$-values; $^*$Holm-adjusted $p{<}0.05$ in \texttt{E1\_mechanism} (only Qwen3.8 $\Delta q$, Holm $p{=}0.0136$; Qwen3.8 IUT Holm $p{=}0.1458$, non-confirmatory). Columns $R_{E_K}{-}R_W$ give the exact whole-design contrast (Theorem~2(i), not a pure adjacency ablation). Descriptive ratio $s_{\mathrm{curl}}{=}E_{\mathrm{curl}}/(R_C{-}R_W)$ exceeds 100\% on all three models because non-curl residuals partly offset curl removal ($R_W{>}R_C{-}E_{\mathrm{curl}}$; not a causal test). At $K{=}6$, all 24 reversal-closed terraces partition into exactly two subgroups (Williams and Alt\_Terrace, 12 seeds each), so min / median / max over all 24 seeds is exact.}
\label{tab:app_e1_exact}
\end{table*}

\paragraph{Exact E1 decomposition and all-24-terrace range.} Only three backbones have exact E1 tables ($N=192$, all $24$ and $720$ orders); we do not extrapolate to the others. On Qwen3.8-27B the signs follow the curl-removal counterfactual ($C-A>0$, $C-\mathrm{curl}-A<0$), but the IUT is not confirmed (Holm $p=0.1458$; \S\ref{app:gates}). On Gemma-4-31B and Llama-3.1-8B, Cyclic already beats Antithetic at $K{=}6$, and removing curl widens the gap. Columns $R_{E_K}-R_W$ report the exact whole-design contrasts (Theorem~2(i)); $s_{\mathrm{curl}}=E_{\mathrm{curl}}/(R_C-R_W)$ exceeds $100\%$ on all three models ($106.7\%$, $139.4\%$, $109.4\%$) descriptively because non-curl residuals partly offset curl removal ($R_W>R_C-E_{\mathrm{curl}}$). Exhaustively enumerating all $720$ seeds of $S_6$ yields exactly $24$ reversal-closed terraces, which partition into exactly two subgroups ($H_{b_W}$ and $H_{b_{\mathrm{Alt}}}$, 12 seeds each), so the min/median/max over all $24$ seeds in Table~\ref{tab:app_e1_exact} is exact. None of the three models passes $\pm0.05$ TOST equivalence of $q_6$ to the isotropic reference $2/9$ of Lemma~\ref{lem:app_iso_ref}.

\begin{table*}[t]
\centering
\scriptsize
\setlength{\tabcolsep}{2.8pt}
\begin{tabular}{@{}llccccccc@{}}
\toprule
Backbone / Cohort & Arm & Acc (\%) $\uparrow$ & Flip (\%) $\downarrow$ & W4 (\%) $\uparrow$ & Brier $\downarrow$ & BrierEx $\downarrow$ & NLL $\downarrow$ & Label Mass \\
\midrule
Qwen3.8-27B (CONF 500) & Base & 63.35 & 29.77 & 46.80 & 0.4772 & 0.06016 & 1.0447 & 0.996 \\
Qwen3.8-27B (CONF 500) & PriDe & 62.50 & 29.90 & 46.20 & 0.4809 & 0.06491 & 1.0461 & 0.996 \\
Qwen3.8-27B (CONF 500) & Random\_WoR & 66.45 & 16.40 & 56.60 & 0.4234 & 0.01000 & 0.8918 & 0.996 \\
Qwen3.8-27B (CONF 500) & Antithetic & 68.25 & 15.07 & 59.40 & 0.4194 & 0.00725 & 0.8754 & 0.996 \\
Qwen3.8-27B (CONF 500) & Cyclic & 67.40 & 14.23 & 58.20 & 0.4228 & 0.00813 & 0.8826 & 0.996 \\
Qwen3.8-27B (CONF 500) & Williams & 68.35 & 9.73 & 63.00 & 0.4156 & 0.00316 & 0.8666 & 0.996 \\
Qwen3.8-27B (CONF 500) & E\_6 & 68.05 & 10.23 & 61.80 & 0.4167 & 0.00354 & 0.8697 & 0.996 \\
Qwen3.8-27B (CONF 500) & Alt\_Terrace & 68.80 & 9.57 & 63.40 & 0.4146 & 0.00271 & 0.8649 & 0.996 \\
Qwen3.8-27B (EXT 1500) & Base & 64.25 & 28.96 & 47.47 & 0.4711 & -- & 1.0189 & 0.994 \\
Qwen3.8-27B (EXT 1500) & PriDe & 63.95 & 29.67 & 46.07 & 0.4779 & -- & 1.0274 & 0.994 \\
Qwen3.8-27B (EXT 1500) & Random\_WoR & 68.95 & 16.98 & 58.33 & 0.4236 & -- & 0.8860 & 0.994 \\
Qwen3.8-27B (EXT 1500) & Antithetic & 69.40 & 14.14 & 60.40 & 0.4219 & -- & 0.8820 & 0.994 \\
Qwen3.8-27B (EXT 1500) & Cyclic & 69.37 & 14.87 & 59.67 & 0.4230 & -- & 0.8853 & 0.994 \\
Qwen3.8-27B (EXT 1500) & Williams & 70.02 & 9.24 & 64.27 & 0.4177 & -- & 0.8733 & 0.994 \\
Qwen3.8-27B (EXT 1500) & E\_6 & 69.88 & 9.90 & 64.33 & 0.4187 & -- & 0.8747 & 0.994 \\
Qwen3.8-27B (EXT 1500) & Alt\_Terrace & 69.73 & 9.53 & 64.20 & 0.4190 & -- & 0.8738 & 0.994 \\
\midrule
Gemma-4-31B (CONF 500) & Base & 70.30 & 27.17 & 54.20 & 0.5157 & 0.19137 & 2.4417 & 0.734 \\
Gemma-4-31B (CONF 500) & PriDe & 70.90 & 25.17 & 56.00 & 0.4946 & 0.17233 & 2.2408 & 0.734 \\
Gemma-4-31B (CONF 500) & PINE$^\dagger$ & 63.35 & 32.37 & 43.60 & 0.6451 & 0.34949 & 2.6930 & 0.767 \\
Gemma-4-31B (CONF 500) & Random\_WoR & 73.95 & 16.97 & 64.40 & 0.3669 & 0.03098 & 1.2862 & 0.734 \\
Gemma-4-31B (CONF 500) & Antithetic & 74.40 & 15.80 & 65.00 & 0.3557 & 0.02543 & 1.2486 & 0.734 \\
Gemma-4-31B (CONF 500) & Cyclic & 75.05 & 13.83 & 67.00 & 0.3592 & 0.02077 & 1.2978 & 0.734 \\
Gemma-4-31B (CONF 500) & Williams & 76.10 & 9.40 & 70.00 & 0.3499 & 0.01247 & 1.2496 & 0.734 \\
Gemma-4-31B (CONF 500) & E\_6 & 75.40 & 10.90 & 68.80 & 0.3516 & 0.01386 & 1.2262 & 0.734 \\
Gemma-4-31B (CONF 500) & Alt\_Terrace & 75.75 & 9.50 & 69.40 & 0.3528 & 0.01277 & 1.2298 & 0.734 \\
Gemma-4-31B (EXT 1500) & Base & 69.65 & 26.22 & 55.33 & 0.5268 & -- & 2.5514 & 0.715 \\
Gemma-4-31B (EXT 1500) & PriDe & 70.75 & 24.86 & 56.40 & 0.5011 & -- & 2.3374 & 0.715 \\
Gemma-4-31B (EXT 1500) & PINE$^\dagger$ & 63.43 & 33.27 & 42.27 & 0.6435 & -- & 2.6957 & 0.749 \\
Gemma-4-31B (EXT 1500) & Random\_WoR & 74.20 & 16.83 & 64.20 & 0.3693 & -- & 1.3450 & 0.715 \\
Gemma-4-31B (EXT 1500) & Antithetic & 75.18 & 14.63 & 67.07 & 0.3625 & -- & 1.3256 & 0.715 \\
Gemma-4-31B (EXT 1500) & Cyclic & 75.85 & 13.07 & 67.80 & 0.3600 & -- & 1.2946 & 0.715 \\
Gemma-4-31B (EXT 1500) & Williams & 76.70 & 10.81 & 70.33 & 0.3498 & -- & 1.2499 & 0.715 \\
Gemma-4-31B (EXT 1500) & E\_6 & 76.55 & 11.04 & 70.20 & 0.3516 & -- & 1.2359 & 0.715 \\
Gemma-4-31B (EXT 1500) & Alt\_Terrace & 76.37 & 9.62 & 70.80 & 0.3499 & -- & 1.2414 & 0.715 \\
\midrule
Olmo-Hybrid-7B (CONF 500) & Base & 41.55 & 49.60 & 23.80 & 0.8290 & 0.18488 & 2.0425 & 0.719 \\
Olmo-Hybrid-7B (CONF 500) & PriDe & 42.80 & 47.57 & 23.40 & 0.7808 & 0.14634 & 1.9367 & 0.719 \\
Olmo-Hybrid-7B (CONF 500) & Random\_WoR & 46.55 & 34.07 & 32.40 & 0.6857 & 0.03011 & 1.5610 & 0.719 \\
Olmo-Hybrid-7B (CONF 500) & Antithetic & 47.60 & 31.30 & 35.00 & 0.6779 & 0.02331 & 1.5571 & 0.719 \\
Olmo-Hybrid-7B (CONF 500) & Cyclic & 48.55 & 23.33 & 37.60 & 0.6677 & 0.01702 & 1.5094 & 0.719 \\
Olmo-Hybrid-7B (CONF 500) & Williams & 49.85 & 16.80 & 42.60 & 0.6595 & 0.00675 & 1.4727 & 0.719 \\
Olmo-Hybrid-7B (CONF 500) & E\_6 & 49.50 & 18.70 & 42.00 & 0.6585 & 0.00703 & 1.4810 & 0.719 \\
Olmo-Hybrid-7B (CONF 500) & Alt\_Terrace & 48.90 & 17.33 & 41.20 & 0.6587 & 0.00608 & 1.4790 & 0.719 \\
\midrule
Olmo-3-7B (CONF 500) & Base & 39.25 & 54.73 & 18.20 & 0.8058 & 0.13413 & 1.9580 & 0.824 \\
Olmo-3-7B (CONF 500) & PriDe & 39.05 & 54.07 & 19.00 & 0.8034 & 0.12855 & 1.9158 & 0.824 \\
Olmo-3-7B (CONF 500) & PINE$^\dagger$ & 35.80 & 57.20 & 14.40 & 0.8311 & 0.18639 & 1.9279 & 0.847 \\
Olmo-3-7B (CONF 500) & Random\_WoR & 44.10 & 35.80 & 30.80 & 0.6936 & 0.02268 & 1.4984 & 0.824 \\
Olmo-3-7B (CONF 500) & Antithetic & 43.65 & 35.33 & 30.20 & 0.6913 & 0.02030 & 1.5056 & 0.824 \\
Olmo-3-7B (CONF 500) & Cyclic & 45.05 & 27.53 & 32.40 & 0.6826 & 0.01314 & 1.4661 & 0.824 \\
Olmo-3-7B (CONF 500) & Williams & 44.70 & 19.97 & 36.20 & 0.6806 & 0.00600 & 1.4508 & 0.824 \\
Olmo-3-7B (CONF 500) & E\_6 & 44.70 & 22.20 & 35.00 & 0.6797 & 0.00653 & 1.4557 & 0.824 \\
Olmo-3-7B (CONF 500) & Alt\_Terrace & 43.95 & 19.13 & 36.20 & 0.6802 & 0.00601 & 1.4485 & 0.824 \\
\midrule
Nemotron-3-30B (CONF 500) & Base & 49.45 & 45.40 & 30.20 & 0.6333 & 0.08129 & 1.3793 & 0.878 \\
Nemotron-3-30B (CONF 500) & PriDe & 52.30 & 44.47 & 30.40 & 0.6122 & 0.06958 & 1.3154 & 0.878 \\
Nemotron-3-30B (CONF 500) & Random\_WoR & 55.25 & 28.67 & 41.80 & 0.5662 & 0.01339 & 1.1865 & 0.878 \\
Nemotron-3-30B (CONF 500) & Antithetic & 55.95 & 26.83 & 43.00 & 0.5680 & 0.01003 & 1.1853 & 0.878 \\
Nemotron-3-30B (CONF 500) & Cyclic & 57.95 & 20.30 & 46.80 & 0.5605 & 0.00618 & 1.1676 & 0.878 \\
Nemotron-3-30B (CONF 500) & Williams & 57.65 & 16.03 & 48.40 & 0.5594 & 0.00231 & 1.1634 & 0.878 \\
Nemotron-3-30B (CONF 500) & E\_6 & 58.10 & 15.90 & 50.00 & 0.5603 & 0.00271 & 1.1667 & 0.878 \\
Nemotron-3-30B (CONF 500) & Alt\_Terrace & 58.55 & 13.80 & 51.60 & 0.5587 & 0.00214 & 1.1601 & 0.878 \\
\midrule
Qwen3.5-9B (CONF 500) & Base & 59.40 & 34.50 & 42.40 & 0.5294 & 0.06515 & 1.1416 & 0.999 \\
Qwen3.5-9B (CONF 500) & PriDe & 60.00 & 34.13 & 42.40 & 0.5240 & 0.06117 & 1.1168 & 0.999 \\
Qwen3.5-9B (CONF 500) & Random\_WoR & 63.15 & 20.73 & 51.80 & 0.4774 & 0.01056 & 0.9939 & 0.999 \\
Qwen3.5-9B (CONF 500) & Antithetic & 62.75 & 16.23 & 53.80 & 0.4775 & 0.00721 & 0.9939 & 0.999 \\
Qwen3.5-9B (CONF 500) & Cyclic & 63.90 & 16.87 & 54.20 & 0.4748 & 0.00762 & 0.9890 & 0.999 \\
Qwen3.5-9B (CONF 500) & Williams & 64.00 & 11.10 & 57.40 & 0.4718 & 0.00301 & 0.9818 & 0.999 \\
Qwen3.5-9B (CONF 500) & E\_6 & 64.20 & 11.40 & 58.00 & 0.4715 & 0.00341 & 0.9831 & 0.999 \\
Qwen3.5-9B (CONF 500) & Alt\_Terrace & 63.85 & 9.57 & 58.40 & 0.4701 & 0.00267 & 0.9784 & 0.999 \\
\midrule
Llama-3.1-8B (CONF 500) & Base & 38.90 & 56.83 & 20.20 & 0.8210 & 0.18085 & 1.9394 & 0.968 \\
Llama-3.1-8B (CONF 500) & PriDe & 42.40 & 47.13 & 24.60 & 0.7180 & 0.09419 & 1.6218 & 0.968 \\
Llama-3.1-8B (CONF 500) & PINE$^\dagger$ & 31.75 & 59.83 & 12.80 & 0.8407 & 0.21648 & 1.8970 & 0.883 \\
Llama-3.1-8B (CONF 500) & Random\_WoR & 44.45 & 40.73 & 28.40 & 0.6544 & 0.03101 & 1.4233 & 0.968 \\
Llama-3.1-8B (CONF 500) & Antithetic & 46.35 & 35.00 & 32.80 & 0.6428 & 0.01828 & 1.3927 & 0.968 \\
Llama-3.1-8B (CONF 500) & Cyclic & 47.60 & 28.07 & 34.80 & 0.6295 & 0.00825 & 1.3312 & 0.968 \\
Llama-3.1-8B (CONF 500) & Williams & 49.65 & 20.23 & 39.80 & 0.6230 & 0.00362 & 1.3134 & 0.968 \\
Llama-3.1-8B (CONF 500) & E\_6 & 48.45 & 22.97 & 38.40 & 0.6239 & 0.00405 & 1.3166 & 0.968 \\
Llama-3.1-8B (CONF 500) & Alt\_Terrace & 49.75 & 19.93 & 39.60 & 0.6230 & 0.00359 & 1.3119 & 0.968 \\
\bottomrule
\end{tabular}
\caption{\textbf{Full per-arm metrics on $K{=}6$ MMLU-Pro CONF ($N{=}500$) and frontier extension ($N{=}1{,}500$).} Extension cohorts omit online reference panels, so BrierEx is marked `--'.}
\label{tab:app_k6_full}
\end{table*}

\paragraph{Full $K{=}6$ metrics.} Table~\ref{tab:app_k6_full} reports every arm. Fidelity, flips and the registered Williams$-$Cyclic deltas use CONF ($N=500$). The frontier extension ($N=1{,}500$) has no online reference panel, so its Brier excess is undefined. Registered Williams$-$Antithetic utility uses CONF+extension ($N=2{,}000$). For this reason, registered deltas need not equal the difference of the displayed pooled accuracies. PINE$^\dagger$ rows are the audited prefill-only blockwise implementation of \S\ref{app:pine_latency}, not a validated reference PINE. They exist only for the three supported dense Transformers.

\begin{figure*}[t]
\centering
\includegraphics[width=\linewidth]{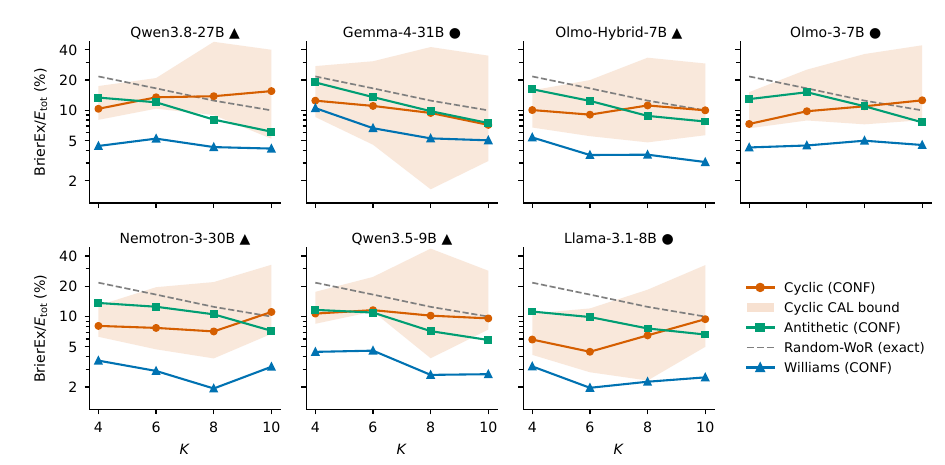}
\caption{\textbf{Design regimes for all seven backbones.} Each panel shows one backbone and plots CONF fidelity error $\apBX/E_{\mathrm{tot}}$ of Cyclic, Antithetic, Random-WoR, $E_K$, Alt\_Terrace and Williams against $K\in\{4,6,8,10\}$. Cyclic loses to Antithetic on six of seven backbones within the grid, and Gemma-4-31B shows no on-grid crossover (Table~\ref{tab:app_crossovers}). Confirmatory Holm inference covers only the two frontier models (Williams vs.\ Cyclic/Antithetic); all other panels are descriptive.}
\label{fig:app_regimes}
\end{figure*}

\begin{figure*}[t]
\centering
\includegraphics[width=\linewidth]{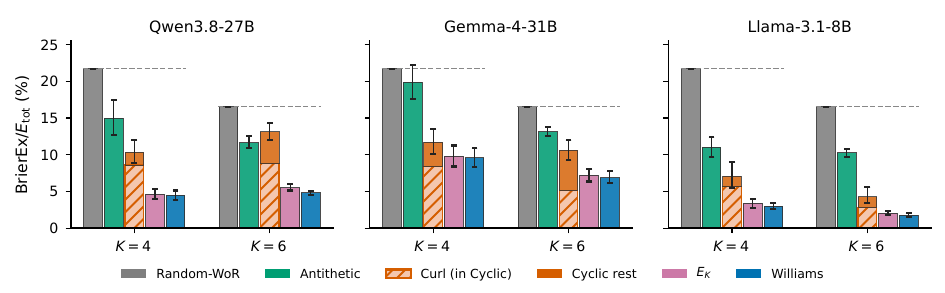}
\caption{\textbf{Exact $S_4/S_6$ decomposition for the three E1 backbones} ($N{=}192$ matched items; all $24$ and $720$ orders). The panels show the curl shares $q_4$ and $q_6$ against the isotropic reference $a_K$ ($1/6$, $2/9$), and the Cyclic$-$Antithetic contrast with and without the curl term (Corollary~B(iii)). Qwen3.8-27B: $C{-}A=+0.00079$ and $C{-}\mathrm{curl}{-}A=-0.00402$, with Holm IUT $p=0.146$ (not confirmed). Gemma-4-31B and Llama-3.1-8B: $C{-}A<0$ already. Error bars are unadjusted bootstrap intervals; decisions use Holm.}
\label{fig:app_e1_three}
\end{figure*}

% ===========================================================================
\section{Audit Duality, Synchronized Latency, and Audited Blockwise PINE Scope}
\label{app:pine_latency}

\begin{table*}[t]
\centering
\scriptsize
\setlength{\tabcolsep}{3.0pt}
\begin{tabular}{@{}l@{\;\;}cc@{\;\;}cccccc@{\;\;}cc@{}}
\toprule
 & \multicolumn{2}{c}{\textbf{Cor.\ A ($K{=}4$ Exact)}} & \multicolumn{6}{c}{\textbf{Synchronized Latency ($K{=}6$, $N{=}128$, ms) $\downarrow$}} & \multicolumn{2}{c}{\textbf{PINE$^\dagger$ / Wil}} \\
\cmidrule(lr){2-3}\cmidrule(lr){4-9}\cmidrule(l){10-11}
Backbone & Max $|\text{res}|$ & $\kappa_{\text{Wil}}/\kappa_{\text{Cyc}}$ & Base p50 & Anti p50 & Anti p95 & Wil p50 & Wil p95 & PINE$^\dagger$ p50 & p50 & p95 \\
\midrule
Qwen3.8-27B & $1.1e-16$ & 0.956 / 0.896 & 203.5 & 437.8 & 902.8 & 438.0 & 904.3 & n/a & n/a & n/a \\
Gemma-4-31B & $2.2e-16$ & 0.895 / 0.875 & 125.5 & 416.5 & 819.8 & 418.4 & 819.7 & 782.8 & 1.87$\times$ & 1.06$\times$ \\
Olmo-Hybrid-7B & $1.1e-16$ & 0.946 / 0.899 & 105.3 & 164.3 & 295.1 & 164.2 & 295.0 & n/a & n/a & n/a \\
Olmo-3-7B & $1.1e-16$ & 0.957 / 0.927 & 49.1 & 122.7 & 210.8 & 122.2 & 210.8 & 403.5 & 3.30$\times$ & 2.07$\times$ \\
Nemotron-3-30B & $1.1e-16$ & 0.964 / 0.919 & 201.8 & 229.2 & 305.5 & 229.7 & 307.0 & n/a & n/a & n/a \\
Qwen3.5-9B & $5.6e-17$ & 0.955 / 0.893 & 110.6 & 177.2 & 315.5 & 177.4 & 315.8 & n/a & n/a & n/a \\
Llama-3.1-8B & $1.1e-16$ & 0.968 / 0.941 & 40.3 & 100.3 & 178.4 & 100.2 & 178.3 & 382.1 & 3.81$\times$ & 2.31$\times$ \\
\bottomrule
\end{tabular}
\caption{\textbf{Exact Corollary A audit-duality verification ($K{=}4$) and synchronized wall-clock latency ($K{=}6, N{=}128$, A100 80GB).} Across all 7 backbones, exact subgroup-coset arms satisfy Corollary A to machine precision ($\max|\text{residual}| \le 3.3{\times}10^{-16}$). Six-pass Williams matches six-pass Antithetic latency within $0.5\%$ (p95 ratio $0.999$--$1.005\times$). The unbatched prefill-only blockwise PINE$^\dagger$ implementation runs $1.87$--$3.81\times$ slower at p50 ($1.06$--$2.31\times$ at p95) on the three supported dense Transformers and is inapplicable to the four hybrid/recurrent models.}
\label{tab:app_cora_latency}
\end{table*}

\paragraph{Audit duality on CONF.} Every exact coset arm satisfies Corollary~A to a maximum recorded residual of $3.33\times10^{-16}$ (Table~\ref{tab:app_cora_latency}). The logged \texttt{energy\_recall} is computed as $\mathrm{mean}(\text{AuditVar})/\mathrm{mean}(E_{\mathrm{tot}})$ over items. For online (sampled-anchor) arms this ratio of means is not numerically identical to $1-\apBX/E_{\mathrm{tot}}$; for example, Gemma-4 Williams gives $0.9135$ vs.\ $0.9335$. The identity is exact only for the exact arms.

\paragraph{Synchronized latency contract.} The contract is fixed at $N=128$ items, the first outer anchor and a warmed model. Every arm is rendered and prefilled independently with no KV reuse, and CUDA is synchronized before and after each wall-clock sample. Hardware and software are as in Table~\ref{tab:app_models}. Under this contract, six-pass Williams and six-pass Antithetic have p95 latency ratios between $0.99934$ and $1.00480$ across the seven backbones. This is a statement about this setup, not universal latency equivalence. The PINE$^\dagger$ / Williams ratios are $1.87$--$3.81\times$ at p50 and $1.06$--$2.31\times$ at p95. Our implementation does serial per-group work, extra score and softmax computation, and host-side conversions. These overheads plausibly make a one-pass implementation slower than batched six-pass Williams, so the ratios do not establish an inherent PINE latency penalty.

\paragraph{Scope of PINE$^\dagger$.} PINE \citep{wang2025eliminating} removes position bias inside attention. PINE$^\dagger$ denotes \emph{our} prefill-only, blockwise implementation. A pre-release audit of the PINE implementation (items M1/M5) established the facts below, which define exactly what our PINE$^\dagger$ rows can and cannot support.
\begin{enumerate}
\item \emph{Index and mask correctness (verified).} The implementation restores choice coordinates by inverse permutation. Its tie key hashes choice text, not the slot letter. Physical block IDs are sorted by those keys, and the ranking mask compares physical token IDs. In final attention the current group is placed last, and the sliding-window mask is $0\le q-k<w$. An exhaustive check over all \textbf{744} $K{=}4$/$K{=}6$ permutations with unequal option lengths (\textbf{19{,}080} group/query coordinate checks) found \textbf{zero} inverse-permutation, slice or window off-by-one errors. Swapping two immutable option blocks changes outputs by exactly $0$: the operator is \emph{fixed-block permutation invariant}.
\item \emph{Scope of the 16/16 CUDA parity check.} On each supported backbone, $16$ CAL fixtures compare native attention with reconstructed attention \emph{under native (identity) positions}. The recorded maximum absolute attention errors are $1.0$ (Gemma-4; covered sliding window $1024$), $0.0625$ (Llama-3.1; no sliding window) and $0.25$ (Olmo-3; window $4096$), accepted under $\mathrm{atol}=0.05$, $\mathrm{rtol}=0.025$. This check validates the native path. It does \emph{not} execute the dynamic option ranking, and it is not a comparison of the edited path against a reference PINE. The separate edited pass checks finiteness and label mass only. No adversarial or reference comparison of Gemma-4's edited shared-KV path was available.
\item \emph{Group-level suffix lookahead (demonstrated).} The implementation accumulates importance over \emph{every} query token in a group, sorts the options once, and uses that order for all queries in the group. The preliminary ranking mask is global even on sliding layers, and only the final SDPA applies the window. A later suffix token can therefore change the effective position of an option for an earlier suffix query. We probed this with the actual hook on CPU: one head, identity $Q/K/V/O$, scale $1$, RoPE angle equal to token position, prefix $(0,0)$, options $(1,0)$ and $(-1,0)$, and first suffix vector $(0.2,1)$. Changing \emph{only} the second suffix vector from $(1.5,0)$ to $(-1.5,0)$ changed the earlier suffix token's output by up to \textbf{0.053} (global attention; $0.053361$) and \textbf{0.625} (window $2$; $0.624832$). Swapping the two option blocks changed it by $0$. Fixed-block invariance therefore holds while prefix causality fails within a group. This is a property of the implemented operator, not leakage of gold labels or CONF data.
\item \emph{Non-invariance under standard slot relabeling.} The prompts render options with slot letters (\texttt{A.}, \texttt{B.}, \ldots; for judging, \texttt{Response A:}). Permuting $[x,y]$ to $[y,x]$ changes the input tokens themselves (``\texttt{A.} $x$ / \texttt{B.} $y$'' becomes ``\texttt{A.} $y$ / \texttt{B.} $x$''), so the result is not a permutation of immutable blocks. Inverse-permuting the output fixes coordinates but not the changed input embeddings. PINE$^\dagger$ can thus produce flips under relabeling even though it is fixed-block invariant. This mechanism is sufficient to explain non-zero PINE$^\dagger$ flip rates; out-of-distribution attention or RoPE geometry is a plausible additional factor, but not an identified cause. High allowed-label mass cannot tell these mechanisms apart.
\item \emph{Structural N/A.} PINE operates on softmax attention. It does not apply to the four hybrid or recurrent backbones (Qwen3.8-27B, Qwen3.5-9B, Olmo-Hybrid-7B, Nemotron-3-30B), which are marked n/a throughout.
\end{enumerate}
On E2 CONF $K{=}6$ ($N=500$), PINE$^\dagger$ lowered accuracy relative to Base on all three supported backbones: Gemma-4 $70.30\%\to63.35\%$, Llama-3.1 $38.90\%\to31.75\%$ and Olmo-3 $39.25\%\to35.80\%$. Flip rates were $32.37\%$, $59.83\%$ and $57.20\%$ (Table~\ref{tab:app_k6_full}). Given items 2--4, every PINE$^\dagger$ result is a result for this audited blockwise variant on letter-prefixed prompts. We do not claim exact invariance, equivalence to the reference PINE, or an inherent limitation of PINE as a method. Any corrected implementation will be archived separately and will not replace the preregistered comparator.

% ===========================================================================
\section{Verbatim Exploratory Pilot Ledger and S0.2 Estimator Validation}
\label{app:pilot}

\begin{table*}[t]
\centering
\scriptsize
\setlength{\tabcolsep}{4.0pt}
\begin{tabular}{@{}lcp{0.64\textwidth}@{}}
\toprule
Exploratory / Pre-GPU Gate & Status & Recorded Verbatim Summary \\
\midrule
Pilot G1 (FP32/BF16 drift) & PASS & Effect fraction 0.979; drift CI90 within $\pm 0.00035$. \\
Pilot G2 (task selection) & FAIL & Exploratory multi-task gate; no task pre-selected without Stage-2. \\
Pilot G3 (large-$K$ pilot) & FAIL & Exploratory; motivated Stage-2 Freeze-A.1 paired reversal probe. \\
Pilot G4 (judge pilot) & FAIL & Exploratory; motivated Stage-2 RewardBench-2 $N{=}500$ confirmation. \\
Pilot G5 (architecture) & PENDING & Exploratory cross-lab confound; replaced by Olmo-3/Olmo-Hybrid pair. \\
\midrule
Stage-2 S0.2 (pre-GPU) & PASS & 17 matched cells; exact max error $3.89\times 10^{-16} < 10^{-10}$. \\
\bottomrule
\end{tabular}
\caption{\textbf{Verbatim exploratory pilot gate ledger (manifest \texttt{432c04f3d0b0}\ldots) and Stage-2 S0.2 estimator validation.} Pilot runs are strictly exploratory and never cited as confirmatory evidence.}
\label{tab:app_pilot}
\end{table*}

\paragraph{Role of the pilot.} The pilot ran before Stage~2 on separate archived data. It is reproduced verbatim in Table~\ref{tab:app_pilot} and is \emph{never} cited as confirmatory evidence. It shaped the Stage-2 design as follows.
\begin{itemize}
\item \textbf{G1} (drift, PASS) justified \texttt{bfloat16} inference.
\item \textbf{G2} (task selection, FAIL) meant that no task was preselected from pilot outcomes. Stage~2 fixes MMLU-Pro and RewardBench~2 in Freeze-A.
\item \textbf{G3} (large-$K$, FAIL) was decisive for Freeze-A.1. At $K\ge6$ the pilot's i.i.d.\ probes produced unstable estimates of $E_{\mathrm{even}}$ and $R_C$. Theorem~3(h) explains this as coincidence degeneracy. On pilot $K{=}6$ cells, the reversal-paired $32{\times}2$ panel reduced the single-panel SD of $\hat E_{\mathrm{even}}$ by a median factor of $96.7$. In a separate variance diagnostic (Table~\ref{tab:app_estvar}), L2O centring also reduced the SD of $\hat E_{\mathrm{curl}}$ by a median factor of $11.4$ while remaining unbiased. Production Freeze-A.1 adopted the reversal-paired $32{\times}2$ panel with fixed-center ($c=\mathbf{1}/K$) between-origin U-statistics before CONF (L2O was not used in production CAL).
\item \textbf{G4} (judge, FAIL) motivated a fresh RewardBench~2 confirmation with $N=500$.
\item \textbf{G5} (architecture, PENDING) could not separate architecture from cross-lab training differences. Stage~2 replaced it with the matched Olmo-3/Olmo-Hybrid pair (\texttt{E4\_Olmo\_matched}), whose results are descriptive of that pair and not causal.
\end{itemize}

\paragraph{S0.2 estimator validation.} Before any GPU run, S0.2 compared the production estimators with exact enumeration on $17$ matched cells. The maximum absolute error was $3.89\times10^{-16}$, below the $10^{-10}$ gate (PASS). Two caveats apply. First, the S0.2 \emph{sign} gate is scored on the mean of $256$ replicate panels, roughly $150$ times the information in a single Stage-2 CAL panel. Stored single-panel Antithetic sign-match fractions are as low as $0.48$. Stage-2 sign calls are therefore issued only when the interval excludes zero; S1's abstentions (\S\ref{app:gates}) reflect this. Second, in a replay under the final estimator code (same random stream, $256$ panels), the 256-panel Monte Carlo relative mean deviation of $\hat E_{\mathrm{curl}}$ was $2.41\%$ for the plain estimator and $0.14\%$ with L2O. Whether the plain estimator passes a $5\%$ empirical threshold on 256 panels therefore depends on Monte Carlo variance, whereas the L2O estimator passes robustly.

\begin{table}[t]
\centering
\scriptsize
\setlength{\tabcolsep}{2.5pt}
\resizebox{\columnwidth}{!}{\begin{tabular}{@{}lcc@{}}
\toprule
\multicolumn{3}{@{}l}{\emph{Pilot $K{=}6$ MCQ, 6 backbones; single panel, $M{=}64$, 60 items, 200 MC panels}} \\
Estimator & Rel.\ SD (min--max) & Max $|z|$ bias \\
\midrule
$\hat E_{\mathrm{curl}}$, plain ($c=\mathbf{1}/K$) & 2.18--5.36 & $<2$ \\
$\hat E_{\mathrm{curl}}$, L2O & 0.21--1.49 & $<2$ \\
$\hat E_{\mathrm{even}}$, i.i.d. & 0.52--6.65 & $<2$ \\
$\hat E_{\mathrm{even}}$, reversal-paired $32{\times}2$ & 0.018--0.034 & $<2$ \\
\midrule
\multicolumn{3}{@{}l}{\emph{$K{=}8$ synthetic, exhaustive truth (40,320 orders), 2,000 panels}} \\
Estimator & Rel.\ SD / \% $<0$ & Median/truth \\
\midrule
$\hat E_{\mathrm{even}}$, i.i.d. & 6.54 / 95.6\% & $-0.28$ \\
$\hat E_{\mathrm{even}}$, reversal-paired & 0.11 / 0\% & 1.01 \\
$\hat R_C$, i.i.d.\ cyclic kernel & 20.9 / 67.7\% & $-10.1$ \\
$\hat R_C$, 8-origin orbit panel & 0.31 / 0\% & 0.98 \\
\bottomrule
\end{tabular}}
\caption{Estimator variance on archived pilot tensors and on a $K{=}8$ synthetic readout (exploratory). The median SD ratios are $11.4\times$ (L2O vs.\ plain $\hat E_{\mathrm{curl}}$) and $96.7\times$ (paired vs.\ i.i.d.\ $\hat E_{\mathrm{even}}$).}
\label{tab:app_estvar}
\end{table}

\paragraph{Exact pilot accounting.} Table~\ref{tab:app_pilot_exact} applies Corollary~B to the archived pilot tensors. These numbers are exploratory and are not recomputed on Stage-2 data. They match the S0.2 ``exact'' fields to $2\times10^{-17}$. Read off the table: $\Gamma_K>1$ in exactly the three cells where Cyclic is worse than Antithetic; $\Delta^{-\kappa}_{CA}<0$ in all $18$ cells; and Gemma-4 MCQ $K{=}6$ is \emph{not} a reversal ($\Gamma_6=0.92$). Stage~2 tested these patterns afresh on hash-frozen data (\S\ref{app:gates}). There the Qwen IUT was not confirmed.

\begin{table*}[t]
\centering
\scriptsize
\setlength{\tabcolsep}{2.4pt}
\begin{tabular}{@{}lcccccccccccc@{}}
\toprule
Cell & Curl share & $E_{\mathrm{curl}}$ & $\|P_Ch_-\|^2$ & $\|P_Cf_+\|^2$ & $R_C$ & $R_A$ & $\Gamma_K$ & $R_C{-}R_A$ & $R_C{-}E_{\mathrm{curl}}{-}R_A$ & $R_W$ & $\kappa_C$ & $\kappa_W$ \\
\midrule
Gemma-4 mcq K4 & .085 & 6.778 & 0 & 3.057 & 9.835 & 13.978 & 0.70 & $-4.143$ & $-10.921$ & 7.739 & .877 & .903 \\
Llama-3.1 mcq K4 & .079 & 5.907 & 0 & 1.065 & 6.972 & 11.125 & 0.63 & $-4.153$ & $-10.060$ & 2.412 & .907 & .968 \\
Nemotron-3 mcq K4 & .071 & 3.409 & 0 & 0.774 & 4.183 & 5.516 & 0.76 & $-1.333$ & $-4.742$ & 2.075 & .913 & .957 \\
Olmo-Hybrid mcq K4 & .097 & 11.845 & 0 & 2.228 & 14.073 & 19.932 & 0.71 & $-5.859$ & $-17.704$ & 7.047 & .885 & .942 \\
Qwen3.5-9B mcq K4 & .084 & 4.025 & 0 & 0.944 & 4.969 & 5.119 & 0.97 & $-0.149$ & $-4.174$ & 2.391 & .896 & .950 \\
Qwen3.8-27B mcq K4 & .064 & 2.258 & 0 & 0.497 & 2.756 & 4.575 & 0.60 & $-1.819$ & $-4.078$ & 1.503 & .922 & .957 \\
Gemma-4 judge K4 & .071 & 8.434 & 0 & 3.265 & 11.699 & 17.433 & 0.67 & $-5.734$ & $-14.168$ & 9.514 & .901 & .920 \\
Llama-3.1 judge K4 & .044 & 9.538 & 0 & 2.488 & 12.026 & 25.757 & 0.47 & $-13.731$ & $-23.269$ & 6.519 & .944 & .970 \\
Nemotron-3 judge K4 & .085 & 3.664 & 0 & 0.989 & 4.653 & 7.390 & 0.63 & $-2.737$ & $-6.401$ & 2.422 & .892 & .944 \\
Olmo-Hybrid judge K4 & .075 & 11.361 & 0 & 3.272 & 14.633 & 31.611 & 0.46 & $-16.978$ & $-28.339$ & 8.601 & .904 & .944 \\
\textbf{Qwen3.5-9B judge K4} & .122 & 6.473 & 0 & 1.580 & 8.054 & 6.945 & \textbf{1.16} & $\mathbf{+1.109}$ & $-5.364$ & 3.207 & .848 & .940 \\
Qwen3.8-27B judge K4 & .068 & 3.255 & 0 & 1.159 & 4.415 & 7.419 & 0.60 & $-3.004$ & $-6.260$ & 2.003 & .908 & .958 \\
Gemma-4 mcq K6 & .056 & 9.803 & 3.546 & 7.178 & 20.527 & 22.306 & 0.92 & $-1.779$ & $-11.583$ & 13.200 & .882 & .924 \\
Llama-3.1 mcq K6 & .018 & 3.802 & 0.621 & 1.701 & 6.124 & 21.778 & 0.28 & $-15.654$ & $-19.456$ & 3.027 & .971 & .986 \\
Nemotron-3 mcq K6 & .053 & 3.963 & 0.410 & 1.337 & 5.710 & 9.661 & 0.59 & $-3.951$ & $-7.914$ & 1.799 & .924 & .976 \\
Olmo-Hybrid mcq K6 & .055 & 9.574 & 1.116 & 3.345 & 14.034 & 22.808 & 0.62 & $-8.774$ & $-18.348$ & 4.989 & .919 & .971 \\
\textbf{Qwen3.5-9B mcq K6} & .092 & 5.374 & 0.547 & 2.066 & 7.987 & 5.960 & \textbf{1.34} & $\mathbf{+2.027}$ & $-3.347$ & 2.487 & .864 & .958 \\
\textbf{Qwen3.8-27B mcq K6} & .104 & 5.524 & 0.602 & 2.088 & 8.214 & 6.225 & \textbf{1.32} & $\mathbf{+1.989}$ & $-3.535$ & 2.689 & .846 & .950 \\
\bottomrule
\end{tabular}
\caption{Exact Corollary~B accounting on archived \emph{exploratory} pilot tensors (units $10^{-3}$, item-averaged; $\Gamma_K=R_C/(\rho_KE_{\mathrm{even}})$, $\kappa_D=1-R_D/E_{\mathrm{tot}}$). $\|P_Ch_-\|^2\equiv0$ at $K{=}4$ by Corollary~B(i). Alt\_Terrace at $K{=}6$: Qwen3.8 $2.669$ vs.\ Williams $2.689$, and Gemma-4 $13.104$ vs.\ $13.200$, i.e.\ close but not equal (Theorem~2(h)). Not confirmatory.}
\label{tab:app_pilot_exact}
\end{table*}

\end{document}